\documentclass[mnsc,noblindrev]{template/informs3}
\OneAndAHalfSpacedXI

\usepackage[english]{babel}

\usepackage[letterpaper,top=2cm,bottom=2cm,left=3cm,right=3cm,marginparwidth=1.75cm]{geometry}

\usepackage{amsmath}
\usepackage{amssymb}
\usepackage{graphicx}
\usepackage{subcaption}
\usepackage[colorlinks=true, allcolors=blue]{hyperref}
\usepackage{multirow}
\usepackage{siunitx}
\usepackage{diagbox}
\usepackage[utf8]{inputenc}
\usepackage{tabularx}
\usepackage{tabulary}
\usepackage{array} 
\usepackage{xcolor}
\usepackage{colortbl}
\usepackage{booktabs}
\usepackage{amssymb}
\usepackage{xcolor}
\usepackage{blindtext}

\definecolor{efc}{rgb}{0.87, 0.19, 0.3}

\newcommand{\cmark}{\textcolor{green}{\checkmark}} % Green tick
\newcommand{\xmark}{\textcolor{red}{$\times$}} % Red cross
\newcommand{\cmnt}[1]{\ignorespaces}  % To make sections invisible

\renewcommand{\theARTICLETOP}{}
\usepackage{geometry}
\usepackage{fancyhdr} 
\makeatletter
\let\c@table\c@figure %
\let\ftype@table\ftype@figure %
\makeatother

\fancypagestyle{firstpage}
{
    \fancyhead[L]{}    
    \fancyhead[R]{}
}

\pdfoutput=1

\usepackage{calc}
\usepackage{floatrow}

\begin{document}

\pagestyle{plain} 

\RUNTITLE{Synthetic Data Applications in Finance}

\TITLE{Synthetic Data Applications in Finance}

\ARTICLEAUTHORS{%

\AUTHOR{Vamsi K. Potluru, Daniel Borrajo, Andrea Coletta\thanks{work done while at AI Research}, Niccol\`o Dalmasso, Yousef El-Laham, Elizabeth Fons, Mohsen Ghassemi, Sriram Gopalakrishnan, Vikesh Gosai, Eleonora Krea{\v{c}}i{\'c},  Ganapathy Mani, Saheed Obitayo, Deepak Paramanand, Natraj Raman, Mikhail Solonin, \\Srijan Sood, Svitlana Vyetrenko, Haibei Zhu, Manuela Veloso, Tucker Balch
%\thanks{}
}
\centerline{ J.P. Morgan AI Research}
\centerline{\{vamsi.k.potluru, first.last\}@jpmchase.com
}
}
\date{today}
%\centerline{July 2023. This Version: July 2023}}
%\ABSTRACT{
%
%\noindent Synthetic data is making enormous strides in various commercial settings including virtual reality, healthcare, and finance. 
%We present a broad overview of typical applications of synthetic data in the finance space and in particular elucidate some of them in greater detail. These cover a wide variety of modalities including tabular, time-series, event-series, and unstructured arising from both from markets and retail applications. Additionally, since banking is a highly regulated industry, we also provide risk mitigation in the context of synthetic data while accounting for privacy, fairness, and explainability. Various metrics are shown in evaluating the quality and effectiveness of our approaches in these rich applications. 
%Rewrite after we have content (placeholder)...
%}%
%

\ABSTRACT{
Synthetic data has made tremendous strides in various commercial settings including finance, healthcare, and virtual reality. 
We present a broad overview of prototypical applications of synthetic data in the financial sector and in particular provide richer details for a few select ones. These cover a wide variety of data modalities including tabular, time-series, event-series, and unstructured arising from both markets and retail financial applications. Since finance is a highly regulated industry, synthetic data is a potential approach for dealing with issues related to privacy, fairness, and explainability. Various metrics are utilized in evaluating the quality and effectiveness of our approaches in these applications. We conclude with open directions in synthetic data in the context of the financial domain. 
%\vp{Rewrite after we have content (placeholder)...}
}
%\end{abstract}
\maketitle

\section{Introduction}

As the name suggests, synthetic data is artificially
generated rather than produced by real world events. 
Synthetic data is created via two primary methods, namely: 1) By 
{\it transforming}
real data, or 2) By {\it simulation} of real processes.
We refer the reader to the  rich literature
on synthetic data and the many mechanisms for creating it~\cite{assefa2020generatingb}. 
Synthetic data has been receiving increased attention in the research community and beyond 
with the wide-spread popularity of generative models such as DALL-E~\cite{ramesh2021zero}  and GPT4~\cite{openai2023gpt4} for the domains of image and text generation, respectively.  In this work, we will primarily focus on tabular and time-series synthetic data which have wide applicability in the retail and investment banking applications such as marketing, trading, and anti-money laundering. Also, we will briefly touch upon modalities involving images and text as can be seen in applications such as check fraud and document understanding. The latter has seen tremendous uptick in financial use-cases with the introduction of chatGPT. %\vp{Talk about some more successes here and prior works on synthetic data in particular}
%\vp{summarize and contrast lukasz paper and also the survey on tabular paper by alvaro}. 

%\subsection{Motivation}
Synthetic data applications in finance can be primarily tagged into the following use-cases. 

\paragraph{\textbf{Data Liberation}:} Data use and sharing within and outside the financial institutions is highly restrictive due to internal policies designed to protect the important relationship of trust between consumers and financial institutions and to ensure compliance with the various regulatory regimes across the globe. These are mostly concerned about the privacy and legal aspects of the customer data and they in turn lead to limits or bureaucracy for data use. Other risks include leakage of institutional knowledge which could pose a competitive risk for the bank~\cite{lin2023summary,tillman2023privacy}. Certain types of synthetic data may potentially alleviate this issue and thereby speed up the adoption of AI and model development process in the firm~\cite{assefa2020generating}. There are many criteria for evaluating the quality of synthetic data and one such notion is that of epistemic parity~\cite{rosenblatt2022epistemic} where the findings on the original dataset match those that are found on the synthetic dataset. Other applications include explanations where instead of using samples from the real datasets, we can generate private synthetic data~\cite{patel2022model}. %\vp{reword so we don't use synthetic data too often}   

\paragraph{\textbf{Augmentation}:}  We can also utilize synthetic data to augment our training data for improving the performance of downstream classifiers. Intuitively, synthetic data can help robustify our training samples when the generated samples are sufficiently diverse from the original dataset. The benefits of synthetic data in image domains have been clearly established~\cite{chang2022style}. It is an open question as to in which regimes does synthetic data provide a lift for training machine learning (ML) models in tabular regimes~\cite{xing2022artificially,manousakas2023usefulness}. Other applications include fairness, where synthetic data can be utilized to the datasets in such a manner that the downstream fairness metrics are improved while preserving classification performance~\cite{li2022fairgan}. %Synthetic data for fairness

\paragraph{\textbf{Counterfactual Scenarios and Testing}:} 

%we utilize a multi-agent market simulator
%to build a synthetic LOB dataset, named DSLOB, with and without market stress
%scenarios, which allows for the design of controlled distributional shift benchmarking. Using the proposed synthetic dataset, we provide a holistic analysis on the
%forecasting performance of three different state-of-the-art forecasting methods.
Learning machine learning (ML) models which are robust to distributional shifts is a challenging problem and synthetic data offers a way of modeling counterfactual scenarios to benchmark these models~\cite{cao2022dslob}. 
Testing robustness of ML systems is a big challenge in operations due to the wider variability of real-world data than the training data that was used to build these production systems (Section~\ref{sec:stress}). 

 %UAT testing use-cases (level 6).

\begin{table}
 \begin{tabular}{ |p{3.9cm}||p{5.3cm}|p{6.2cm}|  }
 \hline
 %\multicolumn{3}{|c||c| c| }{} \\
Modalities & Models & Applications   \\
 \hline
 Tabular& CTGAN~\cite{xu2019modeling}& Fraud, AML (section~\ref{sec:tabular} )\\
 \hline
 Event series & Hawkes~\cite{Zuo2020transformer} & Multi-touch attribution (section~\ref{sec:event}) \\
 & Automated Planning~\cite{planning-book}& Customer Journeys (section~\ref{sec:event})  \\
 \hline
 Time series& TimeGAN~\cite{Yoon2019TimeseriesGA}  & Market counterfactuals (section~\ref{sec:time})\\
 \hline
  Discrete time-series & Bayes Net~\cite{sanner2010relational}~\cite{ taitler2022pyrddlgym}& Asset allocation (section~\ref{sec:related work})  \\
 \hline
 Images &   ScrabbleGAN~\cite{fogel2020scrabblegan} & Check OCR (section~\ref{sec:image} )\\
 \hline
Documents & Bayes Network~\cite{raman2022synthetic} & Layout generation (section~\ref{sec:document})\\ 
 \hline
\end{tabular}
\caption{List of modalities and illustrative applications}
\end{table}

The paper is structured as follows:  we provide a brief review to the various synthetic data generation techniques in the literature (Section~\ref{sec:related work}). This includes simulation based techniques (Section~\ref{sec:sim_methods}), various metrics for measuring the quality of the synthetic data as well as publicly available libraries for the generation. Privacy is a huge requirement in many applications and we provide a novel framework for categorizing all synthetic data into six levels (Section~\ref{sec:privacy}). We then consider data arising from various modalities and tackle them one by one. We start with various applications of synthetic data in the tabular settings (Section~\ref{sec:tabular}) and follow it up by providing succinct applications for event-series in the case of customer journeys and multi-touch attribution (Section~\ref{sec:event}). Time series is another modality which is widely prevalent in many financial applications and we provide compelling synthetic data use-cases in generation, imputation, constraint satisfaction among others (Section~\ref{sec:time}). Finally, we consider applications in unstructured data such as images and text (Section~\ref{sec:document}) for check processing and document understanding. We conclude with some of the open questions in the field.

\section{Background and Related Work}
\label{sec:related work}
%\subsection{Modalities}

%\subsection{Risks}
%\vp{Give general intro to privacy and fairness}
Availability of data is a crucial factor for decision making in the finance domain. However, the sensitive nature of information in this domain makes access to shared data difficult. Although synthetic data offers a route to mitigate this limitation, synthetic data per se is not automatically private~\cite{jordon2022synthetic}. We devote Section~\ref{sec:privacy} to discuss this important issue. In this section, we will provide a high-level overview of the various generation techniques that are available in the literature and the metrics to evaluate the quality of the synthetic data that is generated. Additional modality-specific generation models and metrics will be discussed in greater detail in the corresponding Sections~\ref{sec:tabular}~\ref{sec:event}~\ref{sec:time},~\ref{sec:unstructured}. Finally, we briefly review some popular packages for generating synthetic data. 
%\vp{Can we add in the table of attacks and risks? Also, let's add in the 6 levels of privacy.}

%\subsection{Packages}

%Fill in available software tools that are available for
%generation, privacy, fairness metrics etc. 
%Synthcity, sdv, tidysynthesis ...

\subsection{Generation techniques}
The classical method (SMOTE) for generating synthetic data for imbalanced datasets is by interpolating between the samples~\cite{chawla2002smote}.
With the recent success of deep learning techniques in various ML tasks, it is no surprise that they work exceedingly well even for generation. In particular, models such as generative adversarial networks (GANs)~\cite{goodfellow2014generative}, diffusion models~\cite{sohl2015deep}, and energy based models (EBMs)~\cite{teh2003energy,du2019implicit} have been widely successful in  a wide variety of synthetic data generation tasks. They gain their flexibility from the fact that they are universal function approximators~\cite{hornik1989multilayer}, as well as their high capacity arising from over-parameterization. GANs have been quite successful for tabular~\cite{patki2016synthetic}, time~\cite{Yoon2019TimeseriesGA}, and image data~\cite{fogel2020scrabblegan}. Recently, diffusion based models have been  performing increasingly well in a variety of generation tasks, avoiding some of the training pitfalls associated with GANs~\cite{kotelnikov2023tabddpm}.
 
Financial data can also be generated using simulators, which can be powerful for a multitude of reasons including helping with data privacy concerns, generating rare data, and their ability to incorporate expert knowledge. Simulators can be built by: (1) encoding domain knowledge into models~\cite{chorafas1995financial_models_and_sim};  (2) learning the model directly from data (eg: deep generative models~\cite{oussidi2018deep_gen_models_survey}); or (3) a hybrid method that includes expert knowledge into the model and then fine-tunes the model with data (e.g. with Bayesian Networks~\cite{abdulkareem2019bayesian_data_human_mix_agentModels}). In this work, when we refer to simulators, we mean a program that rolls out the state of a system (its variables) over time or steps. So, simulation-based methods are mostly pertinent for time series or event-series data in finance.

%\vp{Effectiveness, realism, private. - comment on them 6 levels of privacy - where does each method stand.}
%this is done in each subsection, at least discussed in model-based sim subsec preamble.

%\subsubsection{}
%
%We go over ML, DL methods like smote, vaes, gans, diffusion models, factored markov models based simulators, planning, etc..
%
%GANs: DPGANs, PATEGAN, DECAF, TimeGAN, ScrabbleGAN
%VAEs: TVAE, CTGAN, TimeVAE, 
%Diffusion models: TabDDPM

\subsubsection{Model-Based Simulation methods}\label{sec:sim_methods}

 % \vp{Agent based modeling at the top of the document and leave the modalities at the later section.}
 % \vp{Keep the separate on simulation different from the types of modalities.}
Model-based simulation is a powerful approach for generating synthetic time series data which can leverage domain knowledge of predefined rules, and helpful assumptions for effective simulation ~\cite{fox1988knowledgeBasedSim}. One advantage of using such simulators to generate data is the ability to explicitly represent the underlying mechanism of data generation by incorporating domain knowledge %such as the mechanisms driving agent behavior~ 
 \cite{abar2017agentbasedsim_survey}. This allows simulators to generate data for rare or extreme scenarios ~\cite{rubino2009rare_event_sim_book} that may not be present in the data. Rare events play an important role in financial applications~\cite{embrechts2013modelling_rare_events_finance}. Such simulators also have the advantage of control and reproducibility by design, allowing researchers to replicate data generation, analyze the effect of small parameter changes, and ensure consistency.

Model-based simulators are also helpful for generating realistic datasets for downstream ML models in cases where private data cannot be released; this can be due to privacy and or regulatory constraints in finance~\cite{FTC_CFPB_van2018technology}. With respect to privacy levels, using a simulator can offer us the highest level of data privacy, % (level 6), 
as the model can be programmed by a domain expert without referencing the data. However, if the model used in a simulator was built referencing (for example) statistical properties or cases from a dataset, then information leakage can occur and the privacy level is reduced.% to level 5. 
%\vp{Is this calibration? If not, how is this different from it?}
% The data generated by the simulator would be analogous to a synthetic dataset generated using statistical methods --which limits privacy to a measure of differential privacy-- albeit generated by a different mechanism. 

%----@ram continue from here----

As stated, simulators can be helpful in simulating potential (unseen) scenarios in markets, and also useful for testing various strategies or policies in finance. The effectiveness and realism of simulators will depend on how accurate (fidelity) the model used in the simulator captures the dynamics of the target features. This often involves a trade off between model accuracy and simplicity (affects speed and comprehensibility). A simulator needs not be highly accurate (lower fidelity) to be useful~\cite{shahrooei2023falsification_multifidelity_sim_safety_testing}; as the statistician George Box put it, \textit{"Essentially, all models are wrong, but some are useful"}~\cite{wrongModels_useful_box1919essentially}.

%https://www.jstor.org/stable/pdf/24587003.pdf?casa_token=q8633dbjtAUAAAAA:i5IMup3gBE2BRY6SuSKLV1HLTdLWx0AQgxNGWGP8yAq1JoH5R381W4ENvIfsQNDbwKpdNHVoUxlBBrD8_rPps9_F23rc1oH6qx8mT0ZmunzAQ88ZMUGCXw

% A simulator may also be intentionally biased to generate data for possible best or worst case scenarios from any state. 

For generating such synthetic data in finance using model-based simulators, we will discuss a set of approaches that include Markov models, automated planning, and agent-based simulators~\cite{abar2017agentbasedsim_survey}.

\paragraph{\textbf{Using Markov models}:}
% RDDL and pyRDDL
% FinRDDL
The Markov assumption~\cite{puterman2014markov_MDPbook,cinlar2013introduction_stochMarkovprocess_book} is often used in modeling systems. By this assumption, the successor state of a system is only dependent on the current state of a system. Markov models have been extensively used to model real-world systems~\cite{white1985real_mdpUses,boucherie2017markov_mdpuses_2}, including finance~\cite{bauerle2011markov_financeUses}. 

A Markov process is one in which the state evolves over time based on the current state alone, based on dynamics defined by the modeler. These processes can be used directly to simulate financial scenarios and generate data~\cite{MonteCarloSim_finance,iacus2008simulation_markov_process_finance}. Data can also be generated by including one or more agents and their behavior, or a policy of actions, in the Markov process. Agents would act in the environment (described by the model) to optimize their objectives. The data generated would then be reflective of system dynamics as well as of dynamics that emerge from the policies (actions) of agents in the environment. How these agents behave need not be explicitly modeled, and can be learned after specifying a reward function; this is what defines Markov Decision Processes (MDPs)~\cite{MDP_book} and their variants. Such models are used in reinforcement learning. In addition to data generation, one can also test (investment) policies on such models and simulate unseen situations by appropriately modifying the model.

\begin{figure}[!ht]
    \includegraphics[width=0.4\textwidth]{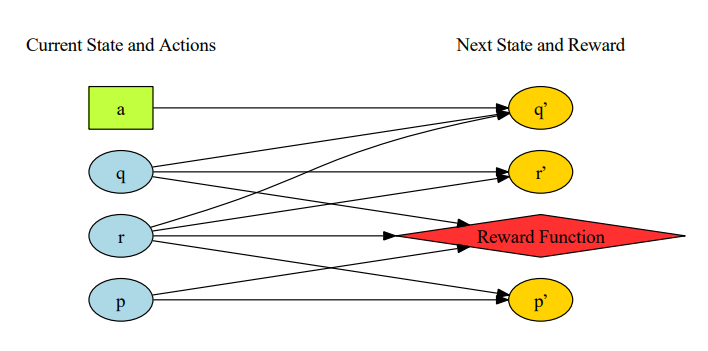}
    \hfill
    \includegraphics[width=0.57\textwidth]{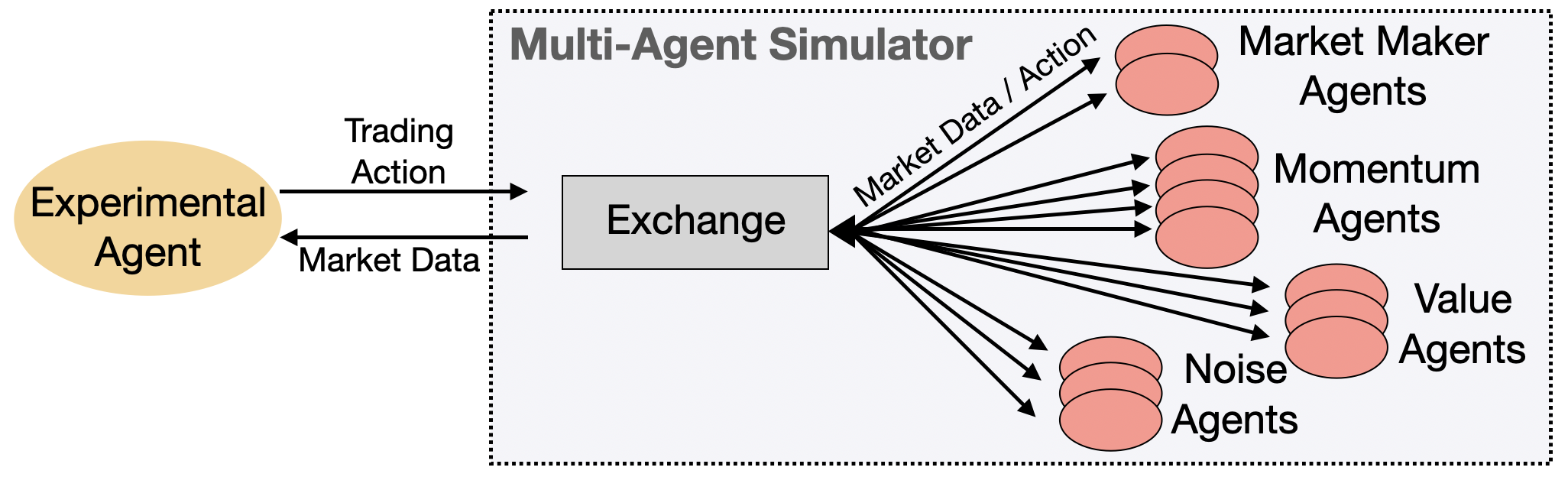}
    \caption{ (Left) A Markov model in RDDL~\cite{sanner2010relational}, (Right)  
    A Multi-Agent Market Simulator~\cite{byrd2019abides}. }
    \label{fig:RDDL}
    %\label{fig:abides_sim}
\end{figure}

In order to define Markov models for generating synthetic data, one popular language is the Resource Domain Definition Language (RDDL)~\cite{sanner2010relational, taitler2022pyrddlgym} which allows one to define MDPs as well as Partially Observable Markov Decision Processes (POMDPs)~\cite{krishnamurthy2016_pomdp}. POMDPs are a type of MDP that includes the ability to reason about state information that is not fully specified. Using RDDL to define a Markov model involves defining how the dynamics of the environment evolves over time, actions an agent may take, and a reward function. However, if one is interested in generating data purely by simulation using a particular model (a Markov process), and not by including or wanting agent behavior data, then the actions and reward function can be made irrelevant. The RDDL simulator\footnote{RDDL simulator code: https://github.com/ssanner/rddlsim}\footnote{RDDL simulator in python: https://github.com/ataitler/pyRDDLGym} (which runs an RDDL model) can be run to generate data. If one ignores actions and rewards, an RDDL model is essentially a dynamic Bayes net~\cite{murphy2002dynamic_DBN} model. In~\cite{finRDDL_patra} the authors give an example of using RDDL to model the problem of asset allocation and optimal trade execution.

\paragraph{\textbf{Agent-based models}:} 
%(Andrea/Svitlana):
%\vp{Is there a more general intro or citation for agent-based models? For instance, can these models be use for modeling financial crime like AML?}
It is often necessary to test trading algorithms against the hypothetical market stress scenarios in order to ensure their robustness before deploying them in real settings. For that purpose, financial price time series are commonly simulated by stochastic processes (for instance, by an Ornstein-Uhlenbeck process as in~\cite{chakraboty}). However, such approaches suffer from an inability to explicitly model interactions among market agents. The knowledge of those interactions is often required to understand scenario nuances. In contrast, agent-based simulation presents a natural bottom-up approach to modeling agent interaction in financial markets~\cite{byrd2019abides,ardon2022phantom,mizuta2016brief}. A schematic example of multi-agent marked simulator, called ABIDES (which stands for Agent-Based Interactive Discrete Event Simulator)~\cite{byrd2019abides}, is shown in Figure~\ref{fig:RDDL}. To emulate the real market, ABIDES defines a pool of heterogeneous agents (i.e., traders) with different strategies to mimic the real market traders. Multi-agent simulators, such as ABIDES, can be used in a {\it forward simulation} mode - i.e., to test how a trading strategy interacts with the simulated market~\cite{vyetrenko2019real}; as well as to generate synthetic series data that contains agent identity/market regime labels. For example, in~\cite{cao2022dslob} the synthetic time series benchmark dataset with volatility regime labels, that was generated using ABIDES, was provided to test the robustness of forecasting algorithms to distributional shifts.

\subsection{Metrics}
\label{sec:metrics}

In order to evaluate the quality of our synthetic data generators, we consider various metrics utilizing both the real and synthetic samples. They can be mainly divided into three main categories:
%\SV{Metric description here seems to be specific to packages, is it the intention??? Should metric description be more accurate (with formulas, references, etc) and detached from the packages?? additionally, packages are described later in section 2.5...}
%\vp{if we add definitions for all it will blow up, might add references for now and see how it looks}

\begin{paragraph}{Fidelity:} These metrics capture the distributional aspects of the synthetic data with respect to the real samples. Typical metrics include the Kolmogorov-Smirnov (KS) test statistic and Chi-Squared (CS) test which measure for the similarity for continuous and categorical variables (columns) respectively. Other distributional divergence measures include
 Jensen-Shannon distance, MMD, and Wasserstein distance. Specialized metrics of distributional similarity such as stylized fact comparison (e.g., similarity of price returns between real and synthetic time series) are used for financial time series~\cite{vyetrenko2019real}. Additionally, T-SNE plots provide a common way to visually check the distributional similarity between real and  synthetic data  in a two dimensional projection~\cite{JMLR:v9:vandermaaten08a}.
\end{paragraph}

\begin{paragraph}{Utility:}
These metrics evaluate the quality of the synthetic data as an effective proxy for real data in classification and regression tasks among others. The various downstream tasks are typically solved
by using XGBoost~\cite{chen2016xgboost}, neural nets, or logistic regression among others. The Training on Synthetic data and Testing on Real data (TSTR) approach allows us to evaluate the usefulness of synthetic data by training a prediction or classification model on synthetic data and testing its performance on a real downstream task~\cite{esteban2017realvalued}.
\end{paragraph}

\begin{paragraph}{Privacy:}
These metrics provide a measure of risk mitigation across the various types of potential attacks on the released synthetic data such as membership inference~\cite{DBLP:journals/corr/ShokriSS16}, attribute inference~\cite{narayanan2007break}, and property inference\cite{lin2023summary}. 
Privacy metrics such as $k$-anonymity and $\ell$-diversity, in addition to reidentification scores such as delta-presence and identifiability score attempt to quantify privacy risk. 
\end{paragraph}

Recently, metrics have also been introduced to account for diversity and authenticity of the generated samples~\cite{alaa2020generative}. We will review additional domain-specific metrics in the corresponding data modality section.
%\SV{
%\begin{itemize}
%    \item Fidelity (aka realism) - various distributional similarity metrics (such as stylized facts in time series modeling), TSTR, t-SNE, discriminator score, likelihood, precision/recall
%    \item Authenticity - Mihaela van der Schaar paper
%    \item Usefulness/effectiveness - performance on downstream tasks (e.g., generalization to unseen scenarios), TSTR
%\end{itemize}
%}

\subsection{Synthetic Data Generation with Python Libraries}

There are several Python libraries available for synthetic data generation, each with its distinct capabilities and features. This section provides a detailed overview of five of these libraries\textemdash SynthCity, SDV, DataSynthesizer, Faker, and Metadata to Data\textemdash and compares them on various metrics. Each library has its own strengths, weaknesses, and unique features. Here, we provide a detailed description of each major Python library for synthetic data generation, including the aforementioned five libraries using common Python libraries like \texttt{pandas, NumPy}, and \texttt{scikit-learn}. A comparison of these libraries can be found in Exhibit~\ref{table:comparison}.

\begin{itemize}
    \item \textbf{SynthCity}: It is tailored for the generation and evaluation of synthetic tabular data. The package has a pluginable architecture, and it encompasses a wide array of reference models, ranging from GAN-based methods to Bayesian Networks, with specialized tools for time series, survival analysis, and privacy-centric synthesis.   
    \item \textbf{SDV (Synthetic Data Vault)}: A Python framework focused on the generation and evaluation of synthetic tabular, multi-table, and time series data. Harnessing a combination of state-of-the-art machine learning models, SDV provides capabilities and data synthesis across diverse use-cases while ensuring that generated datasets closely resemble original data in structure and statistical properties. 
    
    \item \textbf{DataSynthesizer}: This Python-based tool designed for the creation of synthetic datasets, with an emphasis on preserving data structure while ensuring privacy. Leveraging differential privacy and other techniques, it offers a balance between data utility and privacy, making it an optimal choice for researchers and practitioners concerned with data anonymization. 
    
    \item \textbf{TGAN (TableGAN)}:  A specialized generative adversarial network (GAN) tailored for generating synthetic tabular data. Merging the capabilities of deep learning with the nuances of structured data, TableGAN enables the creation of high-quality synthetic datasets, preserving intricate data patterns and relationships while offering an alternative to traditional data augmentation methods. 
    
    \item \textbf{Faker}: A lightweight library primarily used for creating fake data for testing purposes. It can rapidly generate large volumes of data in a variety of formats. While Faker does support complex data types, it does not provide advanced features like data anonymization, or correlation modeling.
    
    \item \textbf{Metadata to Data (using Python libraries like pandas, NumPy, scikit-learn)}: This approach involves generating synthetic data based on the metadata of a given dataset. It uses common Python libraries, making it highly flexible and customizable to specific needs. The functionality of this approach heavily depends on the specific implementation, but it can potentially support all functionalities, including complex data types, correlation modeling, and data anonymization. It also provides the freedom to optimize performance based on the specific requirements and computational resources available. 
\end{itemize}

\begin{table}[ht]
\centering
\renewcommand{\arraystretch}{1.5} % Adjusts the row height
\begin{tabulary}{\linewidth}{|m{2.8cm}|m{2.2cm}|m{1.5cm}|m{2.2cm}|m{1.5cm}|m{1.2cm}|m{2.2cm}|}
\hline
\diagbox[width=3.1cm]{Criteria}{Library} & \textbf{SynthCity} & \textbf{SDV} & \textbf{Data Synthesizer} & \textbf{TGAN} & \textbf{Faker} & \textbf{Metadata to Data} \\
\hline
\textbf{Spatially Aware Data} & \Large\cmark & \Large\xmark & \Large\xmark & \Large\xmark & \Large\xmark & Variable \\
\hline
\textbf{Data Anonymization} & \Large\xmark & \Large\cmark & \Large\cmark & \Large\xmark & \Large\xmark & Variable \\
\hline
\textbf{Supports Complex Data Types} & \Large\xmark & \Large\cmark & \Large\xmark & \Large\cmark & \Large\cmark & Variable \\
\hline
\textbf{Statistical Similarity} & \Large\cmark & \Large\cmark & \Large\cmark & \Large\cmark & \Large\xmark & Variable \\
\hline
\textbf{Advanced ML Models} & \Large\xmark & \Large\cmark & \Large\xmark & \Large\cmark & \Large\xmark & Variable \\
\hline
\textbf{Correlation Modeling} & \Large\xmark & \Large\cmark & \Large\cmark & \Large\cmark & Manual & \Large\cmark \\
\hline
\textbf{Performance} & High & Variable & Variable & Variable & High & Variable \\
\hline
\end{tabulary}
\caption{Comparison of Python Libraries for Synthetic Data Generation}
\label{table:comparison}
\end{table}

%\vp{Let's ensure that the contributions are clear in each and every section. What's ours vs literature.}
\section{Privacy}
\label{sec:privacy}

We begin this section by discussing risks posed by data sharing that are specific to the financial domain. We next review {privacy attacks} studied in machine learning literature. We then discuss the relationship between the privacy risks in the financial domain and the privacy attacks. Finally, we discuss various levels of (privacy) defense that can be applied to the original data or embedded into the synthetic data generation process, and the protection these provide in the context of privacy attacks. We refer to these levels as {privacy levels}. %or \vp{PETS connection?}.

\subsection{Privacy risks in the finance domain}

In financial institutions, data sharing between various lines of businesses within the institution as well as externally is governed by various regulations and internal guidelines that are put in place to protect clients’ sensitive information and protect firms from MNPI (Material Non-Public Information), litigation, reputation, and competitive risks. In this section, we review some prominent risks and relevant regulations in this space. 

\paragraph{Fair Credit Reporting Act (FCRA)}

This act requires that information collected by consumer reporting agencies (e.g. credit bureaus) cannot be provided to anyone who does not have a purpose specified by the FCRA. In particular, it is not enough to only remove from the data fields that identify an individual. In addition, one needs to ensure that the identity cannot be revealed using other data fields, outcome of an algorithm used on the data, and/or publicly available information. 

\paragraph{Regulation on Unfair, Deceptive or Abusive Acts or Practices (UDAAP)} Sharing and certain uses of identifiable data may be sensitive to consumer or a client, and it thus represents potential UDAAP risks if used or shared in a manner contrary to elections made by, or representations made to, consumers or clients. In particular, in many settings sharing identifiable data is subject to privacy elections made by consumers.

\paragraph{Litigation risks}  Inappropriate release of data or functions of data (e.g., models trained on data, insights from data, or synthetic data resembling these datasets) that reveal personally identifying information or statistics (a.k.a. global characteristics) of the data, may pose litigation risks. This is particularly prominent in the context of data sourced from external vendors: use of such data is typically bounded by contracts that precisely define the scope of the use.

\paragraph{Competitive risks} Publishing data that resembles characteristics of a firm's client base or industries and publicly traded companies the firm has interest in, may pose competitive, antitrust and increased insider trading risks. This holds even if the published data is synthetic.

%\paragraph{Third-party/partner/client contracts}

\subsection{Privacy attacks in machine learning literature}

ML models and their outputs are vulnerable to various privacy attacks~\cite{Sun_2023}. The basic assumption of a privacy attack is the existence of a maliciously intentioned adversary who aims to elicit some private information based on the model output. In this paper, we focus on the model output in the form of synthetic data. Privacy attacks come in various flavors. Each distinct attack is characterized by a set of assumptions; e.g. what information is available to the adversary, what information needs to be protected, what is the goal of the attack, etc. Here, we briefly overview some of the most relevant attacks, in the context of privacy risks in the domain of finance (see Table~\ref{table: privacy attacks vs reg risks}).

\paragraph{Membership inference attacks (MIAs)}
%The idea behind the MIA \cite{DBLP:journals/corr/ShokriSS16} is that the fact that 
In many cases, the presence of an individual’s data in a dataset by itself can reveal sensitive information. The adversary's task in MIA~\cite{DBLP:journals/corr/ShokriSS16} is to successfully infer whether an individual was present in the training dataset or not, based on the output of a data processing procedure (e.g. an ML classifier or a synthetic data generator). Moreover, an adversary with the knowledge of an individual’s presence in the dataset can further use linkage attacks (reconstruction attacks) to identify sensitive attributes of that individual. For example, if all data columns matching public information of an individual correspond to an entry in the dataset with a given private attribute, the presence of the individual in the dataset reveals that private attribute for that individual. Thus, MIA can be used as a stepping stone to launch other types of attack.

\paragraph{Reconstruction attacks (attribute inference attacks)}
Reconstruction attacks are characterized by an adversary who is in possession of partial knowledge of a set of features with the aim to recover \textit{sensitive} features or the full data sample. A notable example of an attribute inference attack is the one where an adversary relies on a public set of (non-sensitive) attributes in order to infer values of a sensitive attribute~\cite{narayanan2007break}.

\paragraph{Property inference attacks}
Property inference represents the ability to extract properties of the original dataset based on the corresponding synthetic data. %For example, if the synthetic data has the 40:60 ratio between men and women, one can infer that women are more numerous in the original data too. 
In general, property inference covers learning of any summary statistic of the original data (e.g. mean value, quantiles, histograms, etc.) under the assumption of the access to synthetic data. Preventing property inference attack necessarily degrades fidelity of the synthetic data~\cite{lin2023summary}.

\begin{table}[]
\begin{tabular}{c|c|c|c|c|}
\cline{2-5}
& FCRA       & UDAAP      & Litigation Risk & Competitive Risk \\ \hline
\multicolumn{1}{|c|}{Membership Inference Attack} & Applicable & Applicable & Applicable      & N/A              \\ \hline
\multicolumn{1}{|c|}{Attribute Inference Attack}  & Applicable & Applicable & Applicable      & N/A              \\ \hline
\multicolumn{1}{|c|}{Property Inference Attack}   & N/A        & N/A        & Applicable      & Applicable       \\ \hline
\multicolumn{1}{|c|}{Model Inference Attack}      & Applicable        & Applicable          & Applicable               & Applicable                \\ \hline
\end{tabular}
\caption{Privacy attacks on synthetic data can lead to breach of various regulations in the financial domain}\label{table: privacy attacks vs reg risks}
\end{table}

\cmnt{
\begin{table}
\begin{center}
\begin{tabular}{ c | c | c | c | c }
  & FCRA & UDAAP & Litigation Risk & Competitive Risk \\ 
 MIA & Applicable & Applicable &  Applicable & NA \\  
Attribute Inference Attack & Applicable & Applicable &  Applicable & NA\\  
Property Inference Attack & NA & NA & Applicable & Applicable \
\end{tabular}
\end{center}
\caption{Privacy attacks on synthetic data can lead to breach of various regulations in the financial domain.}
\end{table}
}

\subsection{Defences against privacy attacks}

In this section, we review some defenses against privacy attacks. Note that the list is non exhaustive. Moreover, we propose a novel hierarchy of defenses against privacy attacks we refer to as \textit{privacy levels}.

\paragraph{Anonymization/PII obscuration} There is a wide range of techniques that rely on personal identifiable information (PII) obscuration or anonymization of sensitive fields (e.g. full or partial masking of characters, mapping of categories into codes, etc.). These are however generally prone to linkage attacks, and thus they do not provide formal guarantees against privacy attacks~\cite{narayanan2007break}.

\paragraph{Randomization}
Randomization is a data swapping technique that aims to provide plausible deniability by making it impossible for an adversary to infer any information regarding the data with absolute certainty. It involves swapping values of certain (or all) data points between unique individuals in the dataset. Randomization often aims to provide privacy while preserving the utility of the downstream task (or the accuracy of a query from the dataset) to an acceptable degree.

\paragraph{Differential privacy}

Differential privacy~\cite{dwork2014algorithmic,arora2023faster} as defence belongs to the family of randomization techniques. It provides theoretical guarantees that a potential adversary with the knowledge of an algorithm’s output (e.g. synthetic data) is not able to distinguish with certainty whether a particular individual was present in the input dataset (e.g. original data).  More precisely, let $\mathcal{X}^n$ represent the universe of datasets consisting of $n$ entries. We say that two datasets $D, D' \in \mathcal{X}^n$ are \textit{neighbouring} if they differ in exactly one data entry, i.e. individual. Let $\mathcal{M}$ represent a mechanism that takes as an input a dataset from $\mathcal{X}^n$ and provides an output, e.g. a synthetic dataset. We say that $\mathcal{M}$ is $(\epsilon, \delta)$-differentially private if for any two neighbouring datasets $D$ and $D'$ and any set $\mathcal{C}$ of outcomes of mechanism $\mathcal{M}$ we have
\begin{equation}
    \mathbb{P} \left( \mathcal{M}(D)\in\mathcal{C}\right) \leq e^{\epsilon}  \mathbb{P} \left( \mathcal{M}(D')\in\mathcal{C}\right) + \delta.
\end{equation}

For $\delta = 0$ and small values of $\epsilon$, $\epsilon$-differential privacy guranatees that for any run of the mechanism, any output is almost equally likely to be observed on any neighbouring database. On the other hand, $(\epsilon, \delta)$-differential privacy guarantees that for all neighbouring datasets the absolute value of privacy loss is bounded by $\epsilon$, with probability at least $1-\delta$. We are now ready to introduce our privacy framework.

\subsection{Privacy levels}
Now we introduce a six-level
privacy defense hierarchy and discuss the privacy attacks, utility implications, and potential privacy guarantees for each level. Each level corresponds to a group of defense mechanisms with increasingly stronger privacy protections.

These levels can provide guidance to businesses regarding
the security and utility of their Synthetic Data.
For instance, they might choose to allow internal sharing
of Level 2 data if it arises from a non-critical source,
but require Level 4 protections for more sensitive data. 
The relevant privacy level should be determined according to
the use case to balance multiple objectives
such as the business goal, security, speed of generation, and utility.

In the first 4 levels, we consider methods where the data is {\it transformed} from
the original dataset to the Synthetic Data.  In the figures, the original data appears
on the left, and the arrows indicate how the data is transformed.  We focus on tabular
data in these examples, but the principles can apply to other types of
data.

\begin{figure}[hbt!]
\center{
\includegraphics[width=0.9\linewidth]{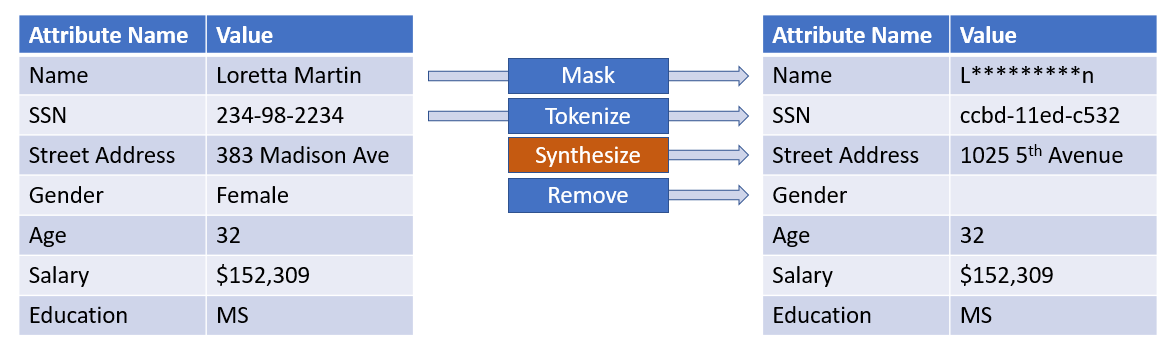}
}
\caption{Privacy Level 1: Obscure PII}
\end{figure}

\subsubsection{Privacy Level 1: Obscure PII}  
Examples of mechanisms at this level include dropping, replacing, masking, or anonymizing the PII attributes. Since this approach does not modify non-PII attributes in any way, it dones not reduce the utility of downstream tasks and accordingly
there is no utility degradation. This however represents weak privacy protection as data remains vulnerable to reconstruction attacks \cite{narayanan2007break}.

\newpage

\begin{figure}[hbt!]
\center{
\includegraphics[width=0.9\linewidth]{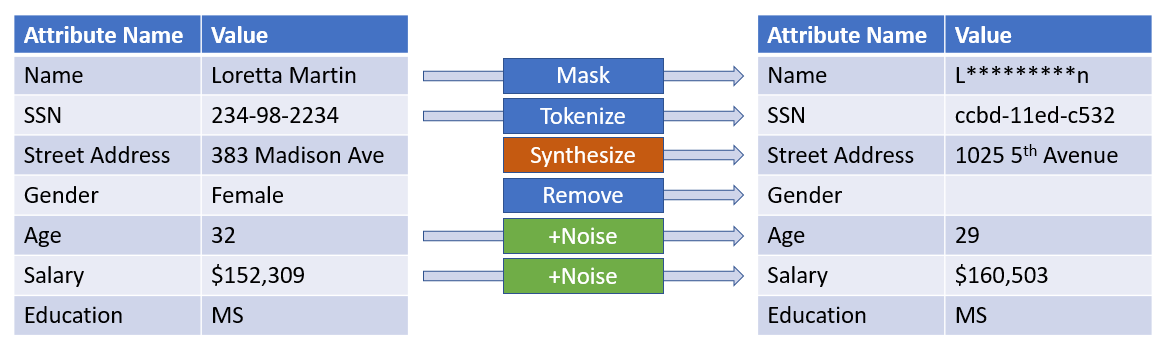}
}
\caption{Privacy Level 2: Obscure PII + noise}
\end{figure}

\subsubsection{Privacy Level 2: Obscure PII + noise} 
In addition to obscuring PII columns, we can deliberately add
noise to other attributes to reduce the effectiveness of potential attacks. 
Differential privacy techniques, for instance, can provide formal guarantees against MIA.

Another approach involves randomly ``swapping''
data between entries.  So for instance, in a demographic dataset, the
ages of the included individuals might be reordered randomly in the
records.  This technique aims to provide plausible but randomized data
by making it more difficult for an adversary to infer any information regarding 
any particular individual. These techniques
aim to elevate privacy while preserving the utility of the data to a downstream 
task. 

Depending on the amount of noise and the downstream task, some degree of utility degradation is expected. 

\newpage
\begin{figure}[hbt!]
\center{
\includegraphics[width=0.9\linewidth]{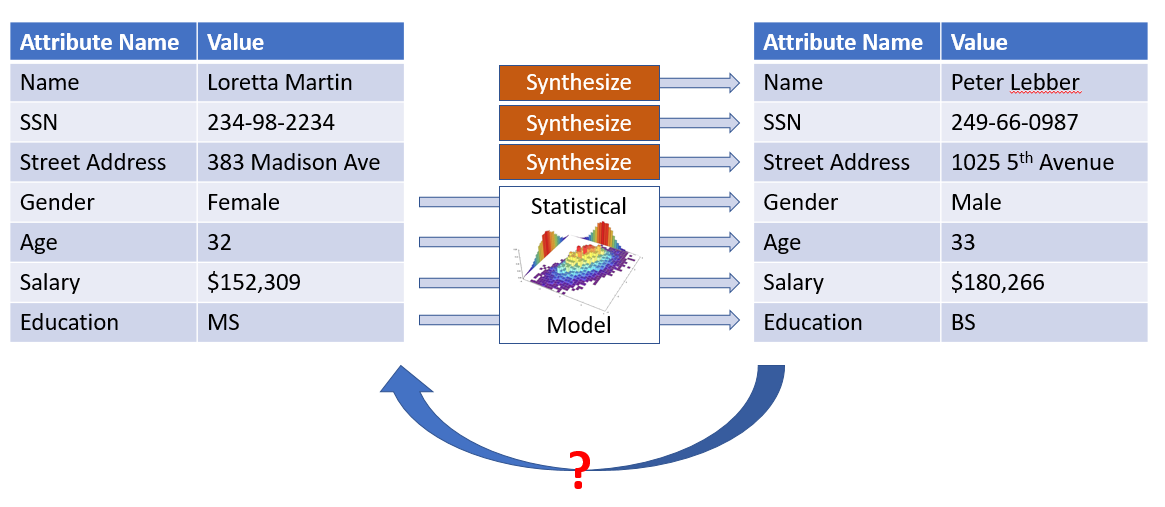}
}
\caption{Privacy Level 3: Generative modeling. The question mark suggests the possibility of
reverse-engineering the data.}
\end{figure}
%\paragraph{Level 3: Synthetic data generation trained on real data}

\subsubsection{Privacy Level 3: Generative modeling} 
Note that Privacy Levels 1 and 2 involve row-by-row transcription of the original data
(with obfuscation or noise as appropriate).
Accordingly, such datasets cannot be larger than the original.

With Level 3 we move to {\it generative} techniques where we analyze the original
data to build a model that can create new data.
Example approaches include Gaussian copula, and Generative-Adversarial-Networks (GAN)
\cite{7796926, goodfellow2014generative, Park_2018}. Other
methods use differential privacy techniques to offer additional guarantees
\cite{asghar2019differentially, xie2018differentially, yoon2018pategan}.
In our own work, we have
introduced a KD-tree-based formulation
to model the data that offers additional protections as well \cite{KDTree2023differentially}.

All these methods enable the creation of new data elements
distinct from the original data. They offer stronger protection
than in Level 1 or Level 2, but are still potentially
subject to attack.  The risk is
increased when the relative size of the generated data to 
the original data is large: For example, if we generate one million 
samples using an original dataset
of only 1,000 we would expect to see generated samples clustering around the
samples in the original data.

%\cn%, with the utility degradation depending on the fidelity of the data generation process and the downstream task. Privacy guarantees in this space generally represent an open question, however there are results for some specific models \cite{DBLP:journals/corr/abs-2206-01349}.

\newpage
\begin{figure}[hbt!]
\center{
\includegraphics[width=0.9\linewidth]{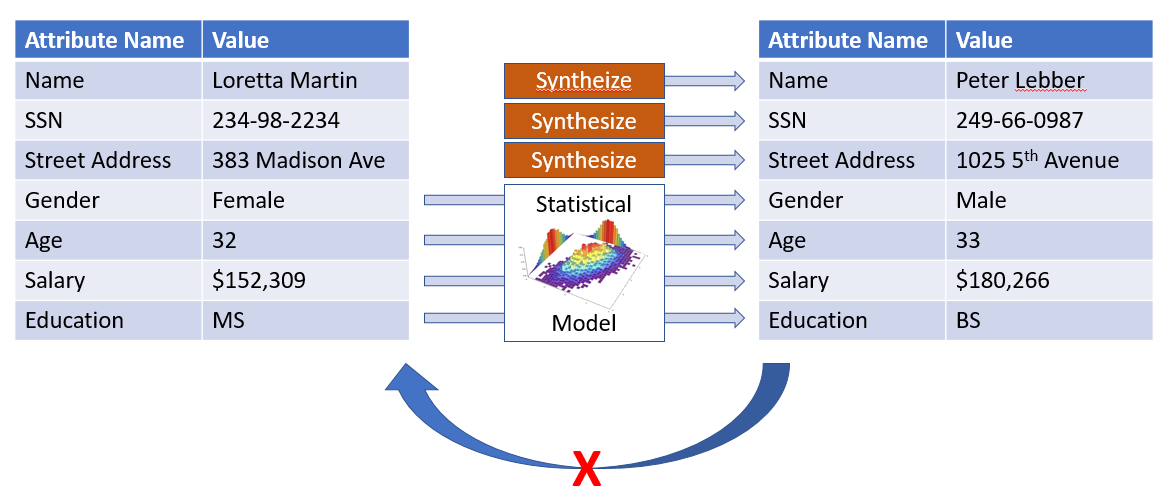}
}
\caption{Privacy Level 4: Generative modeling + testing}
\end{figure}

\subsubsection{Level 4: Generative modeling + testing} 

For Level 4 we add explicit testing of each generated dataset to validate its
resistance to specific attacks.
The particular tests  and the corresponding scores
required to ``pass'' depend on the data and the application.  For instance,
it may be acceptable for certain properties of the data to ``leak'' while others should not.
To operationalize this, we leverage published attack algorithms, then score the 
data depending on the success of the attack.

While it is hard to specify which test and which score would be necessary
to achieve Level 4 privacy in all cases, the important and
critical difference above Level 3, is the fact that the data is explicitly tested.
The test and the scoring criteria must be determined by the individual business for the
use case. Example scoring criteria measure resistance to membership inference, attribute 
reconstruction, and property attacks. among others~\cite{anonymeter,houssiau2022framework,houssiau2022tapas,belgodere2023auditing,du2024towards}. 

\newpage
\begin{figure}
\center{
\includegraphics[width=0.9\linewidth]{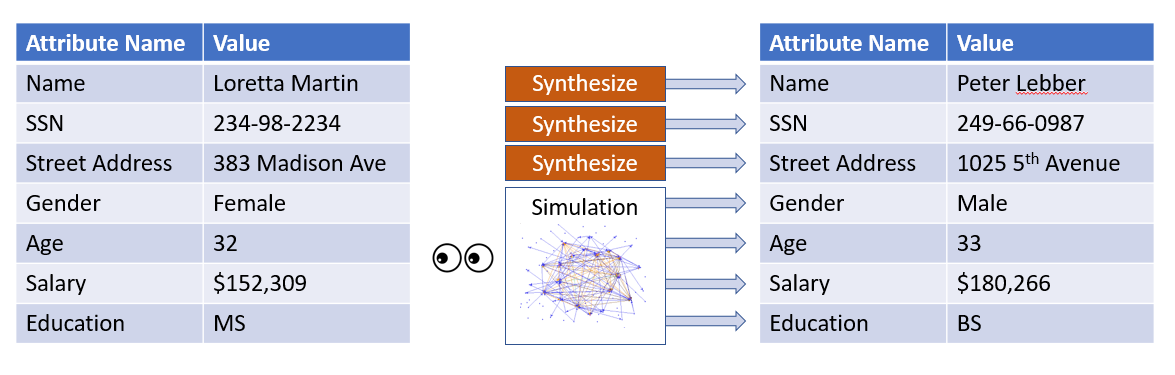}
}
\caption{Privacy Level 5: Calibrated simulation}
\end{figure}
\subsubsection{Level 5: Calibrated simulation}
In this approach the generation method 
is not trained on real data. In fact, there 
is (usually)
no learning in this approach. Instead, we rely on
simulations governed by rules or knowledge
of the process
that would otherwise generate real data. 
These rules, however, are calibrated with reference to the real process
such that the generated data follows 
some statistical properties of the original, real system. 
As an example, we might use a simulation of the stock market to generate
stock price data.
In our own work, we have developed calibrated simulations of equity markets that correspond to 
Level 5 privacy \cite{vyetrenko2019real}.

Utility degradation depends on the downstream task and the simulation framework. This approach generally represents a strong defense against adversarial attacks.
However, they may be exposed to Property Inference Attacks, because
the simulator is calibrated with respect to statistical properties of 
the real system.

\newpage
\begin{figure}
\center{
\includegraphics[width=0.9\linewidth]{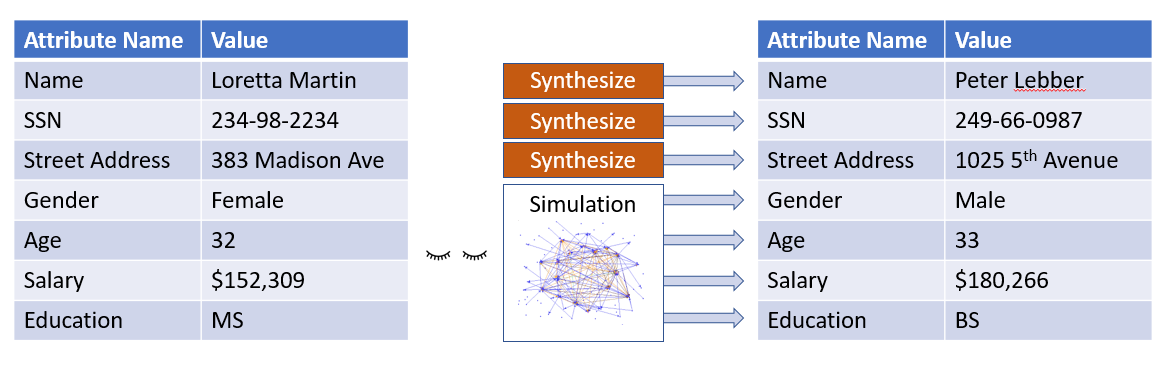}
}
\caption{Privacy Level 6: Uncalibrated simulation}
\end{figure}
\subsubsection{Level 6: Uncalibrated simulation} 
In this case we may not be aware of the statistical properties
of the modeled system, or we might deliberately avoid adjusting the simulation
to correspond to the properties of the original system.
Even though such a simulation may not provide high fidelity data, it can
still prove quite useful.

An important
use is testing, in which we might use a simulation to
explore all the potential
values of data fields to see if they ``break'' our downstream processes.
Additionally, we might choose to embed known examples
of situations we want to be sure our systems detect (e.g., fraudlent transactions).
Another use is to create what-if scenarios
where we hypothesize the impact of one factor on another, to see if
visualization techniques might enable us to discover those relationships in practice.

In general, this method yields a strong privacy guarantee.  
It remediates one of the consequences of level 5 generation of defence against PIA attacks, 
given that the statistical properties of the data is uncalibrated to the real dataset.

\subsubsection{Privacy Levels to Guide Synthetic Data Use Cases}

The creation of these privacy levels has the benefit of being transposed across business use-cases. Levels one and two lend themselves to use cases around sharing data after removing the confidential elements of the original dataset. Levels three and four translate well to improving AI models, be that through augmenting the original data, or testing AI models. Finally, levels five and six can support software engineers in testing applications among others. We will see additional level five synthetic data applications in the time-series Section~\ref{sec:time} using ABIDES. Level 6 data in particular can be utilized for software testing, proof of concepts, hackathons, stress testing business rules in applications, and data migrations among others.

%\vp{There are two examples here, maybe cut it down to one but add how it was done? If it is not calibrated how is it useful?}
%\subsubsection{Synthetic Data Generation - Stress Testing Software Applications}

\paragraph{Case Study: Stress Testing Software Applications}\label{sec:stress}

%\paragraph{Problem Statement}
A common challenge in the software engineering space is stress testing applications to better understand performance. For financial systems, market volatility can cause unexpected increases in the volumes traded, when unexpected events occur, such as Brexit or the outbreak of Covid-19, which saw abnormal increases in the volume of trades. To better understand how the supporting technology will function when such events occur, it is important to test these types of scenarios before they occur to ensure stability of these systems. Using production data for testing is not feasible due to privacy concerns.
%\paragraph{Solution}

We created level $5$ calibrated simulation-based synthetic data to contain similar statistical properties, while removing all records of confidential data in the generated dataset. Millions of rows of data were generated in order to test the application if a sudden spike in trades were to occur. This enabled a streamlined approach to stress testing that had previously not existed.

%\section{Applications by modality}

\section{Tabular Data}
\label{sec:tabular}

In this section, we focus on one of the most ubiquitous type of financial data,  namely tabular data, and in particular synthetic data generation, privacy, fairness and robustness of downstream classifiers.

\subsection{Generation}
\begin{figure}[!ht]
    \centering
    \includegraphics[width=0.4\textwidth]{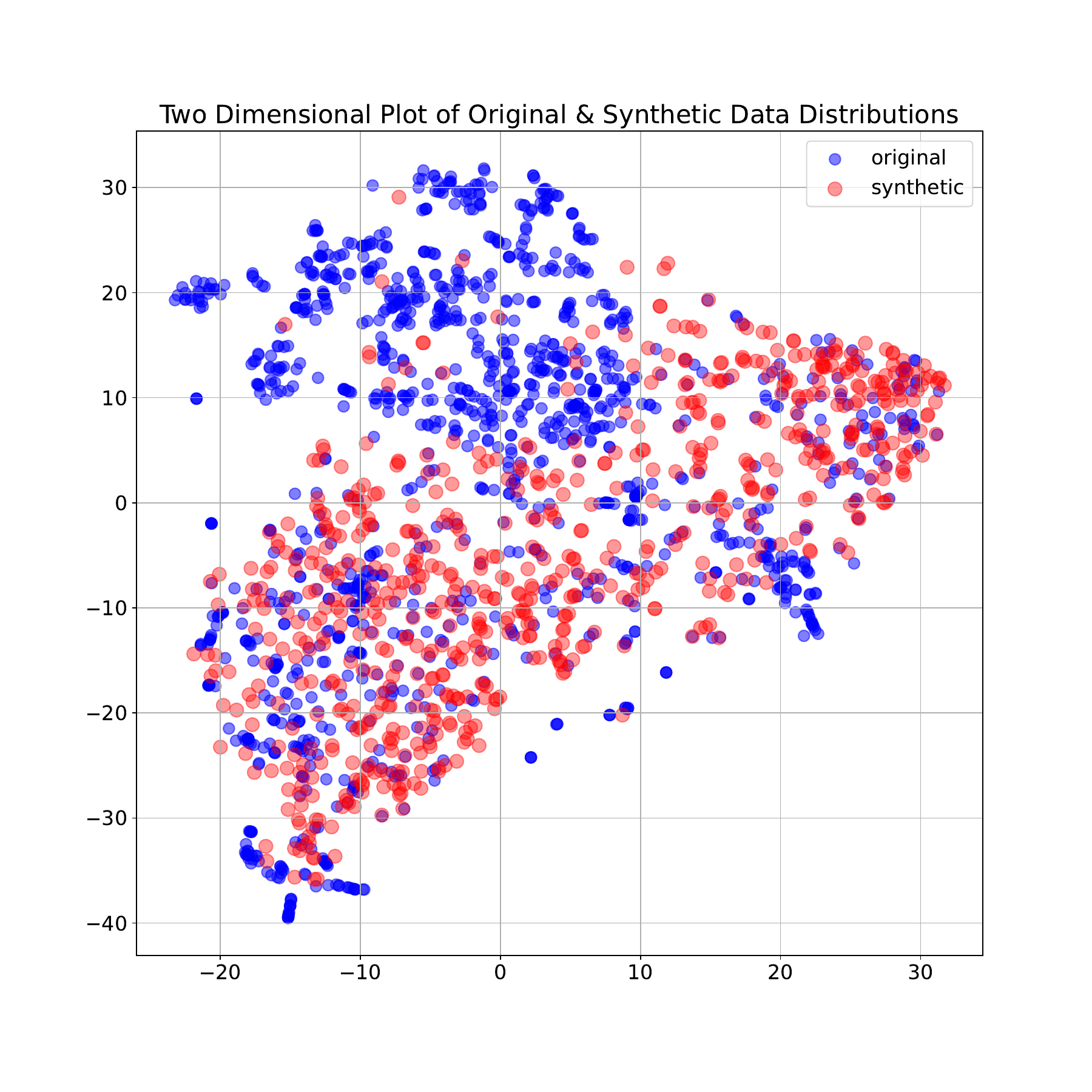}
    \caption{TSNE plot showing similarity between original data and synthetic data.}
    \label{fig:tsne-dt-figure}
\end{figure}

%\vp{let's give visual example of sample data and synthetic generated table and maybe we can do a tsne-type plot for similarity}
%\vp{Can you prune this section? Augmentation is not the only use-case of synthetic data: liberation/sharing, testing are the other ones}
Real-world domains are often described by what we call tabular data, i.e., data that can be structured and organized in a table-like format. Synthetic data generation of these types of datasets is vital as it resolves data scarcity and quality concerns by providing synthetic data that preserves the overall statistical properties and relationships among the attributes of the original dataset (Figure~\ref{fig:tsne-dt-figure}). Synthetic data enables testing new ideas without compromising real data, blending multiple sources, and protecting individual privacy when sharing~\cite{voigt2017eu,tantipongpipat2021differentially}. However, major questions arise from the use of synthetic data. Particularly, is there a major cost that comes with replacing the original training data with the synthetic data we generated? To address this question, the development of a framework to mitigate performance degradation is indispensable.
% We often meet situations where the original data we have is not enough for training models. This is a major problem as the training dataset is small and lacks diversity leading to inaccurate models. To overcome this issue, we can use synthetic data to increase the our dataset creating a more effective model. To provide an analogous purpose to that of the original dataset,  the synthetic data generated needs to preserve overall statistical properties and relations between the attributes of the original data (Figure~\ref{fig:tsne-dt-figure}). Major questions arise from this process. Particularly, is there a major cost that comes with replacing the original training data with the synthetic data we generated?To address this question, the development of a framework to mitigate performance degradation is indispensable.

%\vp{lets prune the whole section to be more compact}
Recent work~\cite{eno2008generating} shows that there are various approaches to model tabular data distribution: statistical-based~\cite{li2020sync}, machine learning-based~\cite{caiola2010random},  Bayesian network-based~\cite{gogoshin2021synthetic}, neural network-based~\cite{Park_2018}. Each of these methods for synthetic data generation possesses unique capabilities and features. The appropriate method depends on many factors such as the distribution of the observed data or the objective of generating the synthetic data. For instance, statistical-based approaches prove advantageous when dealing with known marginal distributions. Due to this, reaching consensus on which method we should use for a specific dataset and use-case remains challenging. Recently, a potential solution for model selection was proposed by using Bayesian optimization~\cite{hamad2023supervised}. 

%\vp{can you be specific on the changes to the architecture to make it work for tabular data?}
We review three popular neural network-based models namely, TVAE, conditional tabular GAN (CTGAN) and CopulaGAN. Since the original GAN formulation~\cite{goodfellow2014generative}, ongoing research has proposed new optimization strategies and modifications to address limitations found on GANs. One of these GAN models, that builds on the success attained by previous architectures, is CTGAN. It uses mode-specific normalization to capture non-Gaussian and multimodal distributions~\cite{xu2019modeling}. This model also introduces a conditional generator and training by sampling to deal with challenges imposed by highly imbalanced categorical columns and sparsity of one-hot-encoded vectors, which is a limitation of previous GAN architectures. CopulaGAN is a variation of CTGAN which takes advantage of the cumulative distribution function (CDF)-based transformation. The other neural network approach called TVAE~\cite{xu2019modeling} is based on variational autoencoders (VAE). 
%Other than neural network generative models discussed, s
Synthetic data generation can also be achieved by treating each table column as a random variable, modeling a multivariate probability distribution, and sampling from it using statistical-based methods. This approach, known as GaussianCopula~\cite{masarotto2012gaussian}, is based on copula functions which are mathematical functions that allow us to describe the joint distribution of multiple random variables by analyzing the dependencies among their marginal distributions~\cite{patki2016synthetic}. This is used to model the covariances among features in addition to the distributions~\cite{llugiqi2022empirical}. Beyond these approaches, synthetic data can also be synthesized by inferring the domain of each attribute, deriving a description of the distribution of attributes in the dataset and sampling from the probabilistic model in the dataset description. This method is called DataSynthesizer (DS)~\cite{ping2017datasynthesizer}. 

%\vp{Can you call this out as our contribution? Add a subsection with problem statement.}
\subsubsection{Optimization Method}

Most of synthetic generation approaches mentioned earlier are ``unsupervised'' in the sense that they do not take into account the downstream task. Most of the approaches discussed earlier treat the label variable  like other covariates. The primary focus of these approaches is to create models that are ``similar'' to original datasets. This creates a conflict in some use cases where the primary objective is optimizing for downstream predictions as opposed to achieving similarity to the original data. We propose a novel synthetic data generation framework, Supervised and Composed Generative Optimization Approach for Tabular data (SC-GOAT)~\cite{hamad2023supervised}, which incorporates a supervised component and optimizes directly on the downstream loss function to address the aforementioned issues. This approach comprises of two steps. In the first step, we incorporate a supervised component customized for the specific downstream task leveraging a Bayesian optimization approach to fine-tune the hyperparameters related to the neural networks. For the second step, we adopt a meta-learning approach to identify the optimal mixture distribution of the existing synthetic data generation approaches. Therefore, the SC-GOAT approach generates synthetic data based on the mixture of multiple synthetic data generation approaches we mentioned earlier.

\begin{table*}[ht!]
  \caption{Description of data sets.}
  \label{table:description-data sets}
  \centering
  \begin{tabular}{llllllll}
    \toprule
    Data set  & Label & Observation & Continuous & Binary & Multi-class & Label = 0 & Label = 1 \\
    \midrule

    Credit Balanced  & 'Class' & 50,000  & 30  &  1 & 0 & 66.70\%\% & 33.3\%\\

  \end{tabular}
\end{table*}

\subsection{Application: Credit Card Fraud}

In this section, we investigate our generative models in terms of the utility of using machine learning for fraud detection.  
%We answer whether synthetic data can help with downstream tasks in the fraud management process. 
%The utility from models trained on fraud dataset allows us to measure the effectiveness of detecting and predicting potential fraudulent operations. 
%This provides guidance to fraud practitioners interested in utility using synthetic data to train fraud detection models. %Compared to using similarity and distance between real and synthetic samples, we will adopt a downstream approach to evaluate the generative models mentioned earlier in the previous section. We will use our generative models to generate synthetic data from the original dataset, train and test our datasets on fraud  detection models and evaluate this via AUROC metric. From this, we can decipher the best synthetic data generative models for different metric combinations and select the best fraud detection models.

\paragraph{Data:}
To showcase the usefulness of synthetic tabular data, we use the credit card fraud dataset\footnote{Data available on the Kaggle platform at \url{https://www.kaggle.com/datasets/mlg-ulb/creditcardfraud}}. The dataset contains transactions collected in the span of two days made by European cardholders for the month of September 2013. The dataset is highly imbalanced, containing 492 frauds out of 284807 total transactions, a 0.172\% of all transactions. The credit card fraud dataset contains only numerical input variables with a total of 31 features. With respect to confidentiality and privacy, 28 of the features - V1 to V28 are principal components obtained by using PCA. 'Time', 'Amount', and 'Class' are the only features not to be transformed with PCA. 'Time' contains the seconds elapsed between each transaction and the first transaction in the dataset. 'Class' is the target variable which takes the value of 0 for cases of no fraud and 1 for cases of fraud. And 'Amount' is the transaction amount. Given the class imbalance ratio of the credit fraud dataset, we processed the dataset by oversampling the minority class with random undersampling of the majority class, leading to a more balanced dataset. This involved duplicating examples in the minority class in order to reach an equal balance between the minority and majority class. We used Synthetic Minority Oversampling Technique (SMOTE)~\cite{chawla2002smote}. The dataset descriptions are summarized in Table~\ref{table:description-data sets}.

\begin{table}[]
\centering
\begin{tabular}{@{}ccc@{}}
\hline
Synthesizer     & Class Frauds (1) & Class No Frauds (0) \\ 
\hline
Original        & 33.33\%           & 66.67\%             \\
GaussianCopula  & 41.9\%           & 58.1\%             \\
CopulaGAN       & 74.76\%           & 25.24\%             \\
CTGAN           & 74.76\%           & 25.24\%             \\
TVAE            & 40.07\%          & 59.93\%             \\
EmpiricalCopula & 33.81\%           & 66.19\%             \\ 
DS 0            & 0.173\%           & 99.83\%             \\
DS 0.1          & 45.47\%           & 54.53\%             \\
SC-GOAT         & 36.10\%           & 63.90\%             \\ 
\hline
\end{tabular}
\caption{Percentage class of synthetic datasets using each model.}
\label{tabular:tabular-class-table}
\end{table}

\paragraph{Evaluation:}

%A very common concern when generating synthetic data is evaluating the quality of the data generated. 
To evaluate the synthetic data generated, various benchmarking approaches are available allowing flexibility in adapting the loss function to suit the specific objectives of synthetic data generation. In this paper, we will adopt a downstream approach to evaluate the generative models. We will use our generative models to generate synthetic data from the original dataset, train and test our datasets on fraud  detection models, and evaluate this via the AUROC metric. This serves as a  well-rounded metric to assess the models' overall performance.  From this analysis, we can decide on the best synthetic data generative models, and select the best fraud detection models.

\begin{figure}[!ht]
    \centering
    \includegraphics[width=0.77\textwidth]{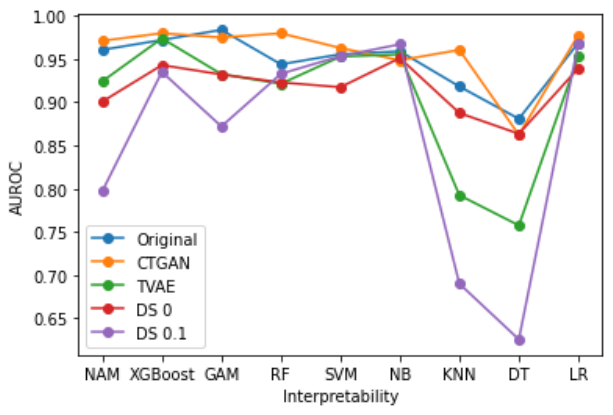}
    \caption{Utility metrics for fraud detection classifiers.  See text for details.}
    \label{fig:AUC-dt-figure}
\end{figure}

\begin{table}[]
\caption{Average, standard deviation, and one-sided paired t-test for the downstream test AUC score, using XGBoost fitted on the generated data by each method on $10$ experiments~\cite{hamad2023supervised}.}
\label{table:results-all-models}
\centering
\begin{tabular}{llllllllll}
\toprule
      & \multicolumn{4}{c}{Untuned}  & \multicolumn{4}{c}{Tuned}  \\
Method  & average & std & test statistic & p-value & average & std & test statistic & p-value \\
   \midrule
   Gaussian Copula & 94.45\% & 0.01    & 14.10 & 0   & 94.45\% & 0.01    & 14.31 & 0   \\
   CTGAN  & 95.34\% & 0.01    & 16.21 & 0   & 95.93\% & 0.01    & 13.15 & 0   \\
   CopulaGAN  & 95.50\% & 0.01    & 14.18 & 0   & 96.41\% & 0.01    & 7.80  & 0   \\
   TVAE   & 98.52\% & 0.00    & 0.00  & 0.5 & 98.48\% & 0.00    & 0.00  & 0.5 \\
   SC-GOAT & \textbf{98.52}\% & 0.00    & - & -   & \textbf{98.48}\% & 0.00    & - & -   \\
  
   \bottomrule
\end{tabular}
\end{table}

Table~\ref{tabular:tabular-class-table} reports the level of balance of various synthetic datasets generated from different approaches without conditional sampling for one experiment. The datasets generated by CTGAN and CopulaGAN  synthesize a  more imbalanced dataset that is dissimilar from the original dataset. The other approaches, TVAE, GaussianCopula and EmpiricalCopula synthesize datasets that are more similar to the original dataset in terms of class balance.  Figure~\ref{fig:AUC-dt-figure} shows a comparison among the performance of the original data and the synthetic data. This chart shows the AUROC results of different models for the data generated with different approaches. The original dataset performs the best for GAM and decision tree models. CTGAN and TVAE, which happen to be neural-network based generative models, improve the results for XGBoost. Therefore, in the fraud detection scenario, when training data synthesized by neural network approaches, XGBoost performs better at distinguishing between fraudulent and non-fraudulent cases. 

Given the positive results of using the XGBoost model, another experiment was performed solely  by training
an XGBoost classifier~\cite{chen2016xgboost} on the training data set and subsequently evaluating its performance on a separate validation data set. Table~\ref{table:results-all-models} shows the AUROC results between the various approaches. The results are reported based on 10 runs of the experiment with 70\% of the real data used for training, 20\% for validation, and the remaining 10\% for testing. The SC-GOAT~\cite{hamad2023supervised} approach outperforms all other approaches by identifying the optimal mixture distribution of existing synthetic data generation methods.

\subsection{Privacy}
     %Privacy-preservation (KD Trees, DUC requirements?)

As previously discussed, privacy preservation of data in the financial domain is crucial in order to assure compliance with the relevant privacy  regulations. A vast body of financial data is presented in tabular format, and thus privacy-preserving generation of synthetic tabular data is a particularly relevant topic in ML for finance.

Privacy considerations in the tabular setting typically assume that the data consists of a number of independent individuals represented by rows of the table, each of which is characterized by a number of attributes contained within the columns of the table. There are various angles of privacy concerns in this setting, and we review the most prominent ones. Often one requires protection of \textit{sensitive attributes} represented by values within a subset of columns~\cite{SUN2011526}. This is relevant not only in the context of publishing the original dataset, but also any output based on the original dataset input (in particular its synthetic data counterpart) that would enable learning of sensitive attributes~\cite{narayanan2007break}. 
Another usual requirement is to protect global statistics of the datasets, such as quantiles or correlations~\cite{lin2023summary}. Finally,  most literature on privacy in tabular data quantifies it with respect to its ability to successfully perform MIA, i.e. correctly identify which individuals were present in the original dataset~\cite{DBLP:journals/corr/ShokriSS16}. As previously discussed, a de facto standard differential privacy represents a defense against MIA. In the remainder of this section, we review existing approaches for differentially private generation of tabular synthetic data.

In the space of statistical-based methods for tabular data, differential privacy is typically achieved by employing perturbation with the use of e.g. the Laplace mechanism~\cite{Li2014DifferentiallyPS, asghar2019differentially}. Marginal-based methods~\cite{tao2022benchmarking} rely on low order marginals in order to fit a graphical model, e.g. PrivBayes~\cite{PrivBayes} and PrivSyn~\cite{PrivSyn}.
Recent work in the area of private synthetic data generation mainly focused on deep generative models which utilize the DP-SGD framework in order to achieve privacy~\cite{Abadi_2016}. Some 
prominent examples of this line of work are DPGAN~\cite{xie2018differentially}, PATE~\cite{papernot2017}, PATEGAN~\cite{yoon2018pategan}, DPCTGAN and PATECTGAN~\cite{rosenblatt2020differentially}. 
%Other deep learning based approaches rely on noise perturbation of mean embedding of data distribution in order to achieve privacy, which is fed into MMD objective that can be optimized in order to train a data generator without further compromising privacy~\cite{pmlr-v80-balog18a, harder2021dpmerf}. 
These approaches however lack interpretability which is of crucial importance in the finance domain. Some recent works that focus on interpretability rely on space partitioning techniques together with noise perturbation in order to achieve differentially-private synthetic data~\cite{KDTree2023differentially}.
We have also explored privacy in diffusion models for tabular data~\cite{wei2023inherent}

\subsubsection{Privacy in Credit Card Fraud Use case}
Our goal is to demonstrate the trade-off between the extent of privacy protection (measured by the scale of noisy perturbation) and the utility of synthetic data in the context of a downstream task. We consider two algorithms that output DP synthetic data based on the original data input: the data-dependent algorithm from~\cite{KDTree2023differentially} which relies on space partitioning and noisy perturbation, and deep learning based DP-MERF~\cite{harder2021dpmerf}. %We present results from~\cite{KDTree2023differentially} which evaluate performance of the two algorithms on downstream binary classification task.

\paragraph{Dataset:} We use a credit card fraud dataset (see discussion above), and use all features except time. We use $80\%$ of input data for synthetic data generation, for various privacy budgets $\epsilon$. Synthetic data is then used to train $12$ classifiers (see~\cite{KDTree2023differentially} for a detailed setup), which are tested on the remaining $20\%$ of the original input data.

\begin{figure}[!ht]
    \centering
    \includegraphics[width=0.6\textwidth]{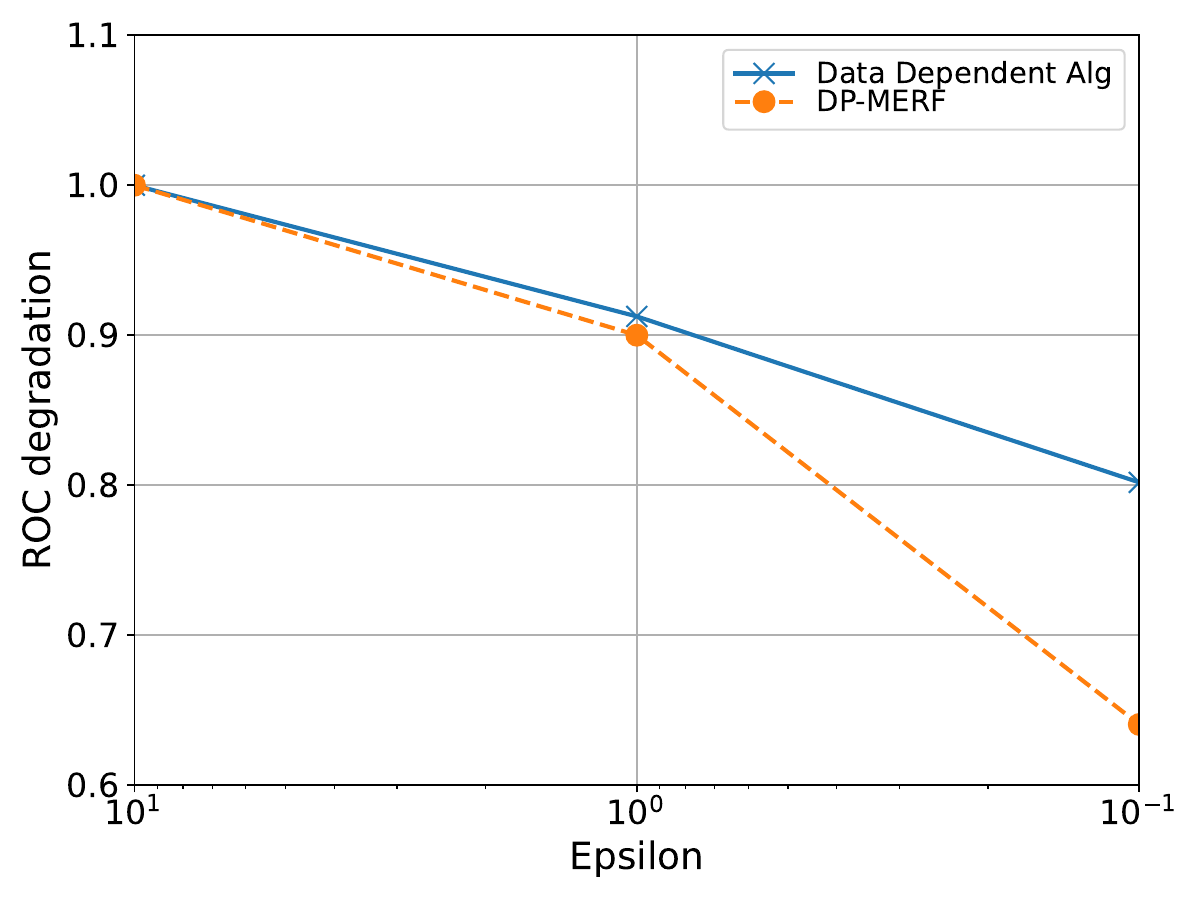}
    \caption{Comparison of the data- algorithm of~\cite{KDTree2023differentially} and DP-MERF~\cite{harder2021dpmerf} on a downstream classification task. ROC degradation is represented as a ratio of the ROC corresponding to a specified $\epsilon$ budget and the ROC for $\epsilon=10$. }
    \label{fig:ROC_degradation}
\end{figure}

\paragraph{Evaluation:} We present degradation of average ROC (area under the receiver operating curve) over $20$ repetitions. The algorithm introduced in~\cite{KDTree2023differentially} does not outperform DP-MERF in terms of ROC values. This is not surprising as it does not rely on deep generative models. However, their performance degrades slower as privacy increases compared to the DP-MERF (Figure~\ref{fig:ROC_degradation}).

%Table \ref{tab:classification_data_dep} shows average ROC for our data dependent algorithm over $20$ repetitions for each classifier, as well as average ROC over the classifiers. Table \ref{tab:classification_dp_merf} shows average ROC over classifiers for DP-MERF with $5$ repetitions for each classifier. Although our data dependent algorithm does not outperform DP-MERF, its performance degrades slower for increasing privacy.

%We follow their experimental setup to train $12$ classifiers on synthetic data and evaluate their performance on original data. 
%%Data dependent algorithm of~\cite{KDTree2023differentially} progresses in two stages: firstly, it achieves differentially private space partitioning, such that the more populated areas are characterized by a finer grid in comparison to sparesly populated ares; once space is partitioned into the set of bins, the algorithm outputs the centers of these bins together with the noisy counts that represent their weights. The simplicity of the approach enables theoretical guarantees on the utility for some special cases.

%For training DP-MERF synthesizers, we set parameters as in~\cite{DBLP:conf/aistats/HarderAP21}, i.e. number of epochs $4000$, number of Furier features $5000$, mini-batch side $0.5$, undersampling rate $0.005$. For our data dependent algorithm, we use undersampling rate of $0.005$, and set the number of data independent levels to be equal to $30$ and maximal number of levels to $60$.

%\begin{figure}[h]
%    \centering
%%\includegraphics[width=0.3\textwidth]{figures/privacy/KD_tree_binning_example.pdf}
%    \caption{Data dependent algorithm of~\cite{KDTree2023differentially} achieves more refined partitioning in densely populated areas. This yields better utility of synthetic data.} 
%    \label{fig: KD_binning}
%\end{figure}
%\textcolor{red}{We need to cite the work of mufei and rongzhe %~\cite{li2023graphmaker,wei2023inherent}} %added them in

\subsection{Fairness}

As synthetic data generation is a powerful tool to both augment and correct automatic decision-making tools, more recent developments have explored and analysed the biases inherited in synthetic data and their downstream effects. This is particularly relevant to commercial banking, where machine learning algorithms are directly used for tasks such as auto loans or credit card applications decisions. Most of recent work revolves around diagnosing and improving generative models~\cite{xu2018fairgan, sattigeri2019fairness, xu2019achieving, xu2019fairganplus, van2021decaf, li2022fairgan}, while other approaches tackle the problem from a transfer learning~\cite{teo2023fair} or optimization~\cite{xiong2023fwc} perspective. We refer the reader to~\cite{mehrabi2021survey} for a  high-level overview.

\subsection{Robustness}

Recent years saw the surge of interest in utilizing synthetic data in order to improve model robustness.~\cite{carmon2022unlabeled} show that training a classifier with additional unlabelled data from the same distribution improves adversarial robustness, i.e. robustness against adversarial attacks that are launched by manipulating features of the test instances.~\cite{deng2021improving} explore how data from a different domain/distribution affects adversarial robustness in the original domain.~\cite{sehwag2022robust} investigate how adversarial robustness of a classifier trained on synthetic data from a proxy distribution translates to the robustness on the real data. In comparison to data arising from a related domain/proxy distribution, the advantage of relying on a synthetic data generator trained on real data is that control over the distance between real and synthetic distribution often comes for free as a consequence of theoretical guarantees on the fidelity of the chosen generator~\cite{Goodfellow_GAN, WGAN, DBLP:conf/nips/LiCCYP17}.

We have reviewed various aspects of synthetic tabular data and there are many open questions. Some of them include: can we generate high quality synthetic data which match the performance of real datasets, differentially private explanations for customer decisions as opposed to utilizing real data, distinguishing synthetic data from real data. We continue our discussion on tabular data with event series which has an additional temporal component that needs to be taken into account.

\section{Event series data}
\label{sec:event}

Event series data record sequences of events, including associated time and related features. In general, a sequence of event series data is composed of multiple events of different types, in which the occurrence of any event at time $t$ might depend on the history (i.e., the sequence of past events) up to time $t$. Unlike time series data, in which a continuously occurring phenomenon is sampled at different resolutions, time interval between two events is usually not predetermined, so event series data are asynchronous. Event series data are ubiquitous in many financial domains. In commercial banking, “customer journeys" log the interactions of customers with the bank. In trading, limit order books record buy or sell orders on a security at a specific price or better. In marketing applications, customer impressions record when a specific ad was served to the customer and whether the customer eventually purchased the advertised product. Outside of financial applications, infectious diseases patterns~\cite{chiang2022hawkes}, earthquake logs~\cite{ogata1998space}, social media interactions~\cite{rizoiu2017hawkes}, and crime modeling~\cite{reinhart2018self}  are among other examples of event series data. Graphs are typically used to model the rich interaction structures between entities such as customers~\cite{longa2023graph} but this makes the generation process much more involved. Diffusion models have been recently explored for the generation of these large-scale graphs~\cite{li2023graphmaker}. 

The literature on modeling sequential data encompasses a broad spectrum of techniques. Traditional approaches include Hidden Markov Models~\cite{baum1966statistical, rabiner1986introduction}, Kalman filters~\cite{kalman1960new, welch1995introduction}, dynamic Bayesian networks~\cite{ghahramani1997learning} and sequential mining patterns algorithms~\cite{mabroukeh2010taxonomy, fournier2017survey}. More recently, deep learning approaches such as recurrent neural networks~\cite{hochreiter1997long, yu2019review}, deep autoregressive models~\cite{oord2016wavenet, van2016conditional} and transformer-based architectures~\cite{vaswani2017attention, lin2022survey} have proven highly effective in sequential data prediction. Within this diverse landscape of sequential data models, a powerful set of tools to model the occurrence of events in event series data are temporal point processes~\cite{cox1980point, daley2003introduction}. Temporal point processes model the probability of occurrence of an event of a given type at any time $t$ (possibly as a function of the history up to time t). Hawkes processes (self-exciting point processes) are a common class of temporal point processes~\cite{hawkes1971spectra, isham1979self, reinhart2018review} that model sequential event series in which the occurrence of an event may increase the probability of occurrence of future events. Temporal point processes are particularly useful as, once trained, one can directly generate synthetic event series data with the same dynamics of the training data; one of the most popular algorithms is the thinning algorithm~\cite{ogata1981lewis}, (see~\cite{dassios2013hawkes} and~\cite{xiao2017wasserstein} for more recent simulation-based approaches). Recent works have focused on improving the Hawkes model capacity by estimating the event dynamics more flexibly beyond self-excitation or for different data types~\cite{Luo2015multi, zhang2019self, Zuo2020transformer, passino2022mutually, ghassemi2022online, zhao2022fast}, while other approaches have focused on the privacy properties of Hawkes processes~\cite{ghassemi2022hawkesprivacy, zuo2022differentially}. Additionally, point processes have been extended to incorporate a stochastic process as intensity function (Cox processes~\cite{cox1955some}) or to assume a probability distribution which can be expressed as a functional determinant (determinantal point processes~\cite{macchi1975coincidence}).
Finally, goodness-of-fit evaluation is usually conducted by looking at the integral of the estimated intensity between out-of-sample events, which is known to follow an exponential distribution with rate $1$ under the ground truth intensity~\cite{daley2003introduction}. Other goodness-of-fit approaches include the difference between the empirical and estimated intensity~\cite{xiao2017wasserstein} as well as non-parametric two sample tests~\cite{wei2021goodness}.

\subsection{Using Automated Planning}
A way of generating synthetic event series data assumes there is a model of the underlying environment and perform a simulation using the model on different scenarios. The different events that appear in the simulation are converted into data points of the output dataset. An example of such technique~\cite{arxiv-finplan-simulator,icaif20} uses classical automated planning~\cite{planning-book}. The assumption is that most data that financial institutions keep on clients, come from client's interaction with the bank. Each interaction can be seen as an action that the client executes and that is observable by the bank. Examples are opening accounts, making wire transfers, or withdrawing money from ATMs. At each time step, a clients state can be described in terms of some facts, such as the accounts they opened, the balance on their accounts, or their regular payments (e.g. monthly rent, yearly taxes, utility bills). Also, at each time step, clients have short-term financial goals, such as paying the rent, buying a product, or receiving the payroll from their employer. Given that we can define clients' interactions in terms of actions, states and goals, we can use an automated planning framework to randomly generate goals for clients, generate plans that achieve those goals from the current state of clients, and execute the actions in those plans. Each action execution leaves a trace that can be logged into a record of a dataset that defines that interaction. Data generated using this method has been used in several papers for purposes ranging from reasoning on entanglements~\cite{KDD23-SD} to goal recognition~\cite{FinPlan23-goal}. In particular, this approach generated datasets that described money laundering activities, fraudulent transactions, or customer journeys~\cite{arxiv-finplan-simulator}.

Figure~\ref{fig:example_journeys} shows an excerpt of a customer journey dataset generated from a simulated trace using planning. Events are usually described by a date-time tag, a label for the executed event, and a client id. The event description includes the channel used by the client, mobile in this case and the corresponding action.

\begin{figure}[hbt]
    \centering
    \begin{tabular}{lll}
    \multicolumn{1}{c}{Date and time} &
        \multicolumn{1}{c}{Event} &
            \multicolumn{1}{c}{Customer ID}\\ \hline
     2021-05-24 21:22:14&mobile : logon &ID-22522\\
2021-05-24 21:25:14&mobile : transaction summary business&ID-22522\\
2021-05-24 21:26:14&mobile : transaction history prepaid account&ID-22522\\
2021-05-24 21:28:14&mobile : ultimate rewards info&ID-22522\\
2021-05-24 21:31:14&mobile : ultimate rewards activity&ID-22522\\
2021-05-24 21:36:14&mobile : logoff &ID-22522\\ \hline
    \end{tabular}
    \caption{Small section of a generated trace of events using AI planning.}
    \label{fig:example_journeys}
\end{figure}

\subsection{Application: Marketing spending insight via synthetic customer journeys}

In this section, we show how data augmentation with synthetic customer journeys can help downstream tasks related to marketing campaigns. In marketing, customer journeys are composed of the customer interactions with different advertising channels, such as search, social media or television ads. Post mortem spending insight is usually conducted by identifying the importance of different channels in terms of customer conversions, i.e., in terms of the number of customers who purchased the advertised product after seeing the ad. Multi-Touch Attribution (MTA) models measure this importance by assigning credit to each channel to drive future budget allocation. Unlike more traditional approaches, where all the credit was assigned to the last customer touch point, MTA models take into account the entirety of the customer journeys; see~\cite{yadagiri2015non, abhishek2017multi, arava2018deep} for examples of MTA models. Temporal point processes can be used to learn the customer journey dynamics, enrich the available training data, and improve the downstream MTA model insights.

% In this section we will show how simulated events data can be useful when running marketing campaigns, more specifically in multi-touch attribution (MTA) tasks. MTA is a marketing measurement model that assigns credit to the different advertising media, or channels, used in a marketing campaign. The input of an MTA model is usually the list of customer interactions with different advertising channels, also called touchpoints. The traditional approach to attribution was to assign 100\% of the credit to the last touchpoint that directly led to the conversion. However, this approach fails to account for the impact of all the other touchpoints that influenced the customer's decision. MTA, on the other hand, takes into account all the touchpoints that a customer encounters on their journey. It assigns credit to each touchpoint based on its level of influence on the customer's decision to purchase. Several different models have been developed in the literature~\cite{yadagiri2015non, abhishek2017multi, arava2018deep}.

\begin{figure}[!ht]
    \centering
    \includegraphics[width=0.57\textwidth]{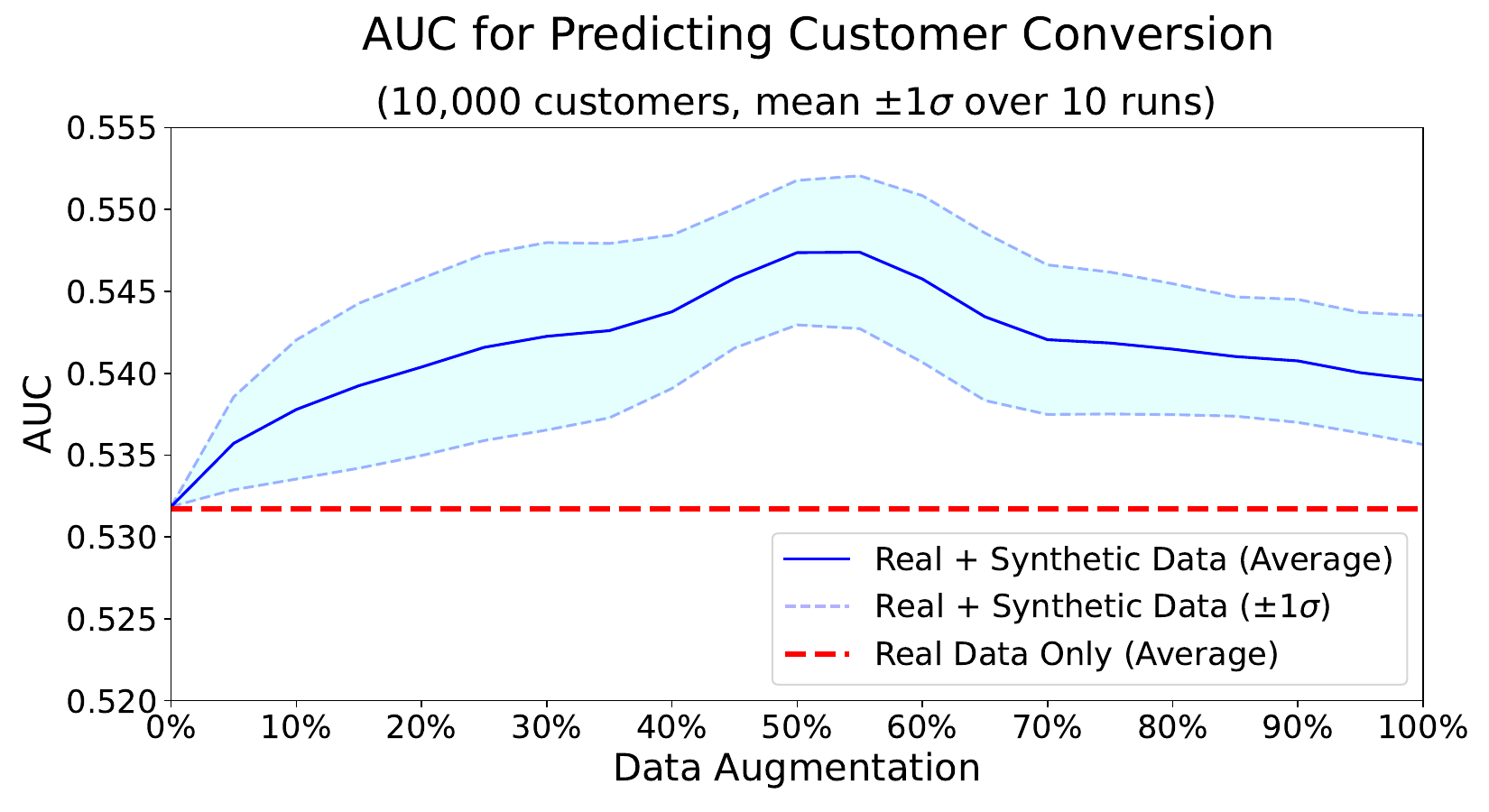}
    \includegraphics[width=0.42\textwidth]{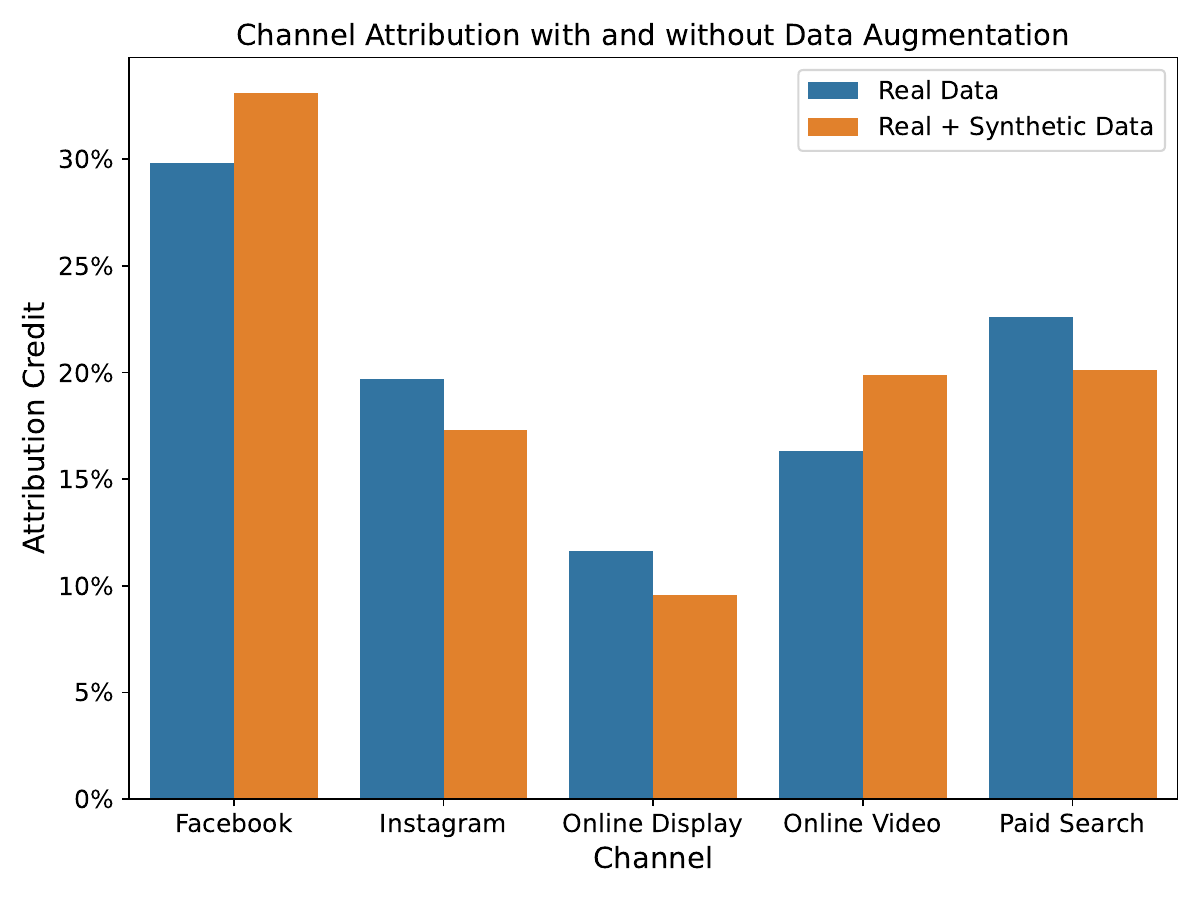}
    \caption{Data augmentation with synthetic event series data (left) and change in MTA credit assignments when using a combination of both real and synthetic data (right).}
    \label{fig:mta-results-figure}
\end{figure}

To showcase the usefulness of synthetic event series data, we use a public MTA dataset\footnote{Data available on the Kaggle platform at \url{https://www.kaggle.com/code/hughhuyton/multitouch-attribution-modelling/notebook}}. The data cover a month-long campaign with the ads channels available in the data being Facebook, Instagram, Paid Search, Online display and Paid Search. We extract $10,000$ customer journeys with a $\sim 7\%$ conversion rate from the campaign, and use a Markov chain MTA model~\cite{kakalejvcik2018multichannel}. We train an XGBoost~\cite{chen2016xgboost} model trained on features extracted from customer journeys data to predict the customer conversion. An improvement in prediction performance would indicate that synthetic event series data do indeed capture the dynamics within the real data. We use a mixture of 4 Hawkes processes for augmentation, conducting a separate hyper-parameter optimization across the number of mixtures and the decay parameters. Figure~\ref{fig:mta-results-figure} (left) reports the XGBoost model AUC as a function of the percentage of data augmentation used in training (shaded bands indicate one standard deviation computed over 10 runs). We see that the addition of synthetic customer journeys improves the classifier performance, although the increase saturates after 50\% augmentation. Figure~\ref{fig:mta-results-figure} (right) shows the MTA credit assignments when using 50\% of the synthetic customer journeys. The synthetic journeys do not radically change the credit assignment, but increase the importance of online video advertisement, which gains credit and ranking.

% \subsubsection{privacy}
% We can have the planning stuff here (AML/Fraud).
%  Also add in the DP hawkes work here for instance. Also, the planning/simulation would be perfect here

% Not sure if we have anything here yet
% \begin{itemize}
%     \item Ab-initio synthetic data (from simulation, HR Data)
% \end{itemize}

%\vp{Lets have the following structure: Introduction, statement of the problem, Application, Experiments and link to published/arxiv paper}

\section{Time series data}
\label{sec:time}
%Structure for time series

\newcommand{\SubItem}[1]{
    {\setlength\itemindent{45pt} \item[-] #1}
}

%\begin{itemize}
%\item	Intro to time series (1 page)
%\SubItem{ Related work (all)}  
%\SubItem{ Describe the various approaches to generation (parametric vs non-parametric (implicit, explicit, agent-based)}
%\SubItem{ Specific Metrics}
%\item	Generation
%\SubItem{	INR   (hypertime common for both imputation/aug) (1 page) (Elizabeth)}
%\SubItem{	Imputation (workshop IJCAI)  0.5 page (Elizabeth)}
%\SubItem{	Augmentation (ICASSP + submission, Dat intern) 0.5 page (Elizabeth PhD work, Yousef ICAIF)}
%\SubItem{	constrained time-series scenarios (hard/soft constraints submitted) (0.5 page, andrea)}
%\SubItem{	Style transfer (Stylized facts) (ICAIF) (supplementary) 0.5 page, yousef}
%\item	Simulation
%\SubItem{   Intro to ABIDES (Haibei)}
%\SubItem{	World agents (1 page, andrea)}
%\SubItem{	Distributional shifts  (neurips workshop 22) (1 page, yousef)}
%\end{itemize}

%%Time-series Synthetic Data (IPO Volumes, Rate Curves, and more..)
%\vp{folks, please figure out a application for each of the subtopics and maybe give an experimental result. We can point to the corresponding arxiv/published paper for further details.}

% {Introduction}
Synthetic time series (TS) data refers to artificially generated time series data that emulate the statistical properties and patterns observed in the real world. %It is generated using models or simulations rather than being derived from observations or measurements.
Synthetic time series are useful when only limited number of historical observations are available. They are also needed to simulate specific scenarios (for instance, market crashes). Such data finds utility in the finance domain in multiple ways, including back-testing investment strategies, evaluating trading algorithm performance for robustness, and simulating market stress scenarios for risk management. Additionally, synthetic time series data can be combined with historically observed data to make model training datasets larger as well as to enhance the representation of minority classes in imbalanced datasets -- both of which typically lead to better generalization capabilities of machine learning models.

\paragraph{\textbf{Approaches to synthetic time-series}} 

We can divide the existing methods for synthetic time-series generation and simulation in three major classes~\cite{goodfellow2016nips}:
\begin{itemize}
    \item \textbf{Parametric models} assume a specific parametric form for the underlying distribution of the data, and they have a fixed number of parameters that are usually estimated from the training data. Once their parameters are estimated, we can generate synthetic samples by sampling directly from the learned distribution. A Gaussian distribution is an example of parametric model, where the mean and variance are the parameters. For time series data, suitable models include probabilistic variants of vector autoregression~\cite{newbold1983arima}, or state-space models. % \textcolor{red}{\cite{}}. 
    In terms of financial time series, stochastic differential equations (SDEs)~\cite{paine2021quantum,jia2019neural} are among the most popular class of models due to their ability to explicitly model time series data as a continuous time stochastic process with an underlying drift (i.e., trend)  and diffusion (i.e., noise). Common SDEs employed to model price time series include geometric Brownian motion and the Ornstein–Uhlenbeck (OU)~\cite{maller2009ornstein} process, where the OU process captures the mean-reversion property 
    %\vp{maybe define the mean-reversion property?} 
    typically observed in financial time series~\cite{bouchaud2018trades}. 
    \item \textbf{Non-Parametric models} do not take any parametric assumptions, i.e., they do not make strong assumptions about the form of the underlying distribution. Instead, these models aim at estimating the data distribution directly from the training samples, often without specifying a set of fixed parameters. Example of non-parametric models are GANs~\cite{goodfellow2016nips} and VAEs~\cite{doersch2016tutorial}. Among non-parametric models we can distinguish between: \textbf{implicit models} which do not need to make any assumption on the density function form, but they rather train and sample directly the samples (e.g., GANs); and \textbf{explicit models} which make an explicit assumption on the form of density function from which they sample the synthetic samples (e.g,. VAEs). A recent work using an implicit model is shown in Section~\ref{sec:simulation_ts}, where a conditional generative network (cGAN) is used to generated synthetic order books~\cite{coletta2022learning}.     
    %\AC{Elizabeth, Yousef, do we cite some work using an explicit model like VAE?}
    
    \item \textbf{Agent-based models} provide a natural bottom-up approach to model the underlying system dynamics (e.g., traders) and to simulate and generate synthetic markets. Agent-based models have been widely used in the economic literature to replicate complex processes, and perform "what-if" studies~\cite{lebaron2006agent}. In Section~\ref{sec:simulation_ts} we discuss ABIDES~\cite{byrd2019abides} a state-of-art Multi-Agent Simulator for high-fidelity market data. 
\end{itemize}

\paragraph{\textbf{Metrics}:}
Fidelity and utility metrics described in Section~\ref{sec:metrics} are widely used for assessing the  quality of generic (i.e., not necessarily financial) synthetic time series data. Properties of financial time series that are repeated across a wide range of instruments, venues, and time periods are referred to as stylized facts~\cite{vyetrenko2019real}. For instance, it is a well known fact that the distribution of daily stock price returns shows fat tails and that distribution is time-invariant. Evaluating the stylized facts of generated financial time series such as distributions of asset returns, order volumes, order arrival times, order cancellations, etc., and comparing them to those derived from real historical data allows us to infer the level of fidelity of a time series generation process.

\paragraph{\textbf{Related Work}:} 
Realistic time series generation has been previously studied in the literature by using the generative adversarial networks (GANs). In the TimeGAN architecture~\cite{Yoon2019TimeseriesGA}, realistic generation of temporal patterns was achieved by jointly optimizing with both supervised and adversarial objectives to learn an embedding space. QuantGAN~\cite{Wiese_2020} consists of a generator and discriminator functions represented by temporal convolutional networks, which allows it to synthesize long-range dependencies such as the presence of volatility clusters that are characteristic of financial time series. TimeVAE~\cite{desai2021timevae} was recently proposed as a variational autoencoder alternative to GAN-based time-series generation. GANs and VAEs are typically used for creating statistical replicas of the training data, and not the distributionally new scenarios needed for data augmentation. Recently, neural SDEs have also been proposed for realistic time series generation. Neural SDEs assume that the time series data follow some underlying latent SDE, where the drift and diffusion of the SDE are modelled via a deep neural network.  
Data augmentation is well established in computer vision tasks due to the simplicity of label-preserving geometric image transformation techniques, but it is still not widely used for time series with some early work being discussed in the literature~\cite{timeseries_augmentation, Wen2020TimeSD}. For example, simple augmentation techniques applied to financial price time series, such as adding noise or time warping, were shown to improve the quality of next day price prediction model~\cite{fons2020evaluating, fons2021transfer}. However, such transformations were not required to produce realistic synthetic time series.
Furthermore, some augmentation methods might work in certain domain-specific time series, but not in others. For example, if noise is added to a sample, what scale of noise should be used? Given a certain dataset, what set of transformations would work best? Therefore, a major challenge in data augmentation is how to search over the space of possible transformations, which can be prohibitive given the large number of possible transformations and their associated hyperparameters. This issue motivated the investigation of several automated data augmentation algorithms~\cite{autoaugment}. Previous methods perform the search by using proxy tasks with small models and training subsets~\cite{fastautoaugment, adversarialautoaugment, PBA2019}, which might not give optimal results in the final task. More recently, RandAugment~\cite{randaugment} proposes to sample augmentations uniformly with the same shared magnitude, where the number of augmentations and magnitude can be tuned with a grid search. A similar approach is proposed for time series in~\cite{fons2021adaptive} where all possible augmentations are weighted with a learnable parameter, while in~\cite{Iwana2022gating} a neural network that dynamically selects the best combination of data augmentation methods is proposed, using a feature consistency loss.

\subsection{Generation}\label{sec:gen_ts}
    \paragraph{\textbf{Application: Realistic Generation of Financial Time Series}}
    \begin{figure}[!h]
    \noindent
    \begin{minipage}{6in}
      \centering
      \raisebox{-0.5\height}{\includegraphics[height=0.7in]{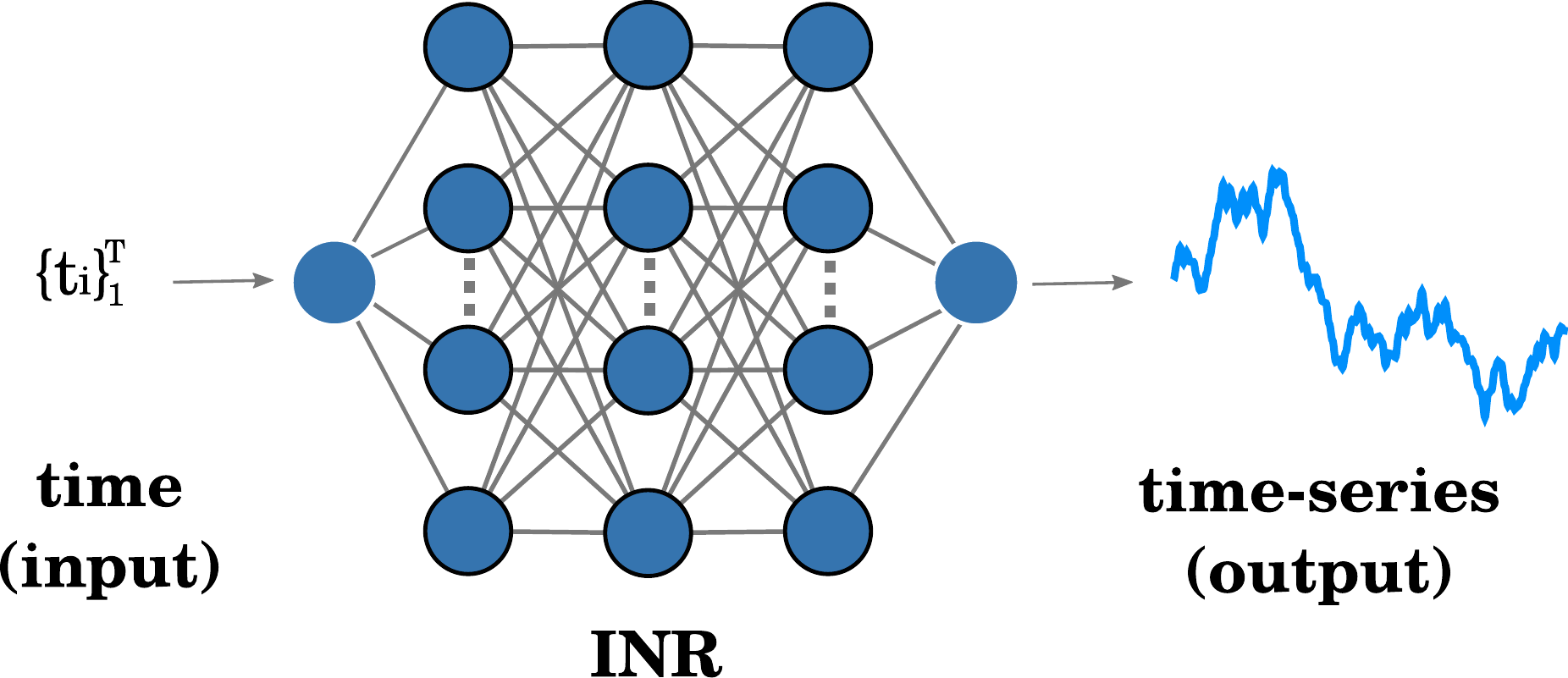}}
      \hspace*{.3in}
      \raisebox{-0.5\height}{\includegraphics[height=1.5in]{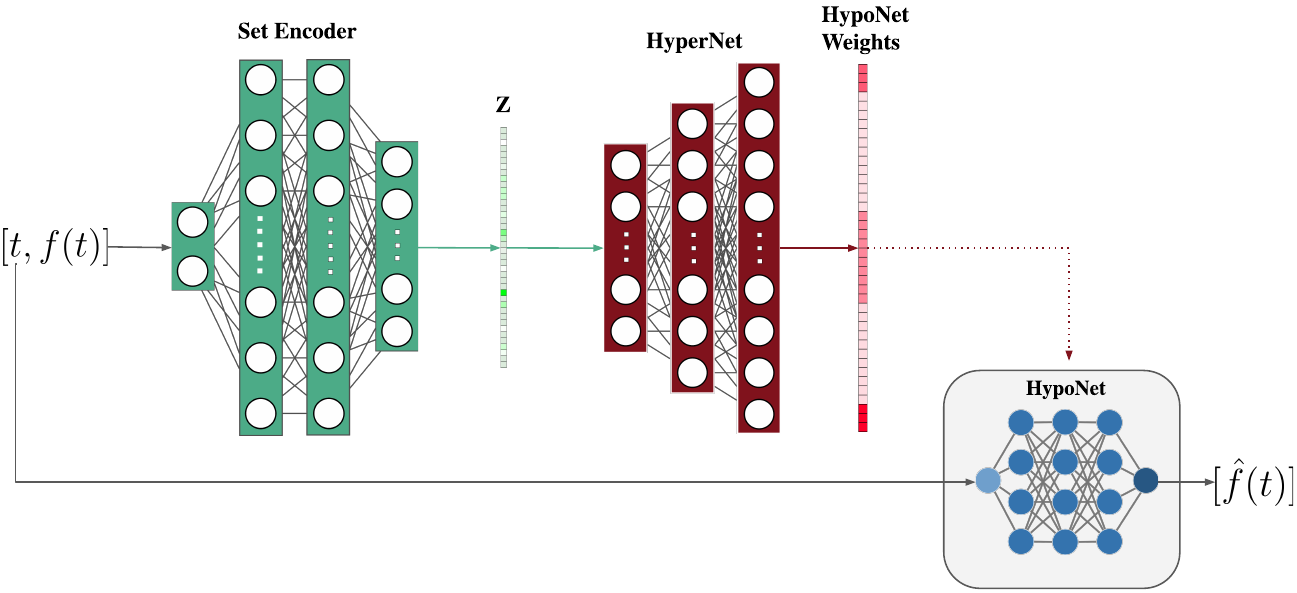}}
    \end{minipage}
    \caption{Diagram of the implicit neural representation (INR) for univariate time series (left). Diagram of HyperTime (right) where each pair of time-coordinate $t$ and time series $f(t)$ is encoded as an embedding $Z$ by the Set Encoder. The HyperNet decoder learns to predict INR weights from the embeddings.}
    % During training, the output of the HyperNet is used to build a HypoNet and evaluate it on in the input time-coordinates. The loss is computed as a difference between $f(t)$ and the output of the HypoNet $\hat f(t)$.}
    \label{fig:inrs}
    \end{figure}
    
    Time series data often presents unique challenges due to its potentially irregular sampling or presence of missing values. These issues can significantly affect the performance of certain deep learning models, leading to inaccurate predictions or flawed insights. Given the crucial role time series data plays across various scientific and economic domains, researchers have focused on the development and deployment of robust deep learning models capable of accommodating these inconsistencies. In recent years, Implicit Neural Representations (INRs) have gained popularity as an accurate and flexible method to parameterize signals from diverse sources, including images, videos, audio, and 3D scenes~\cite{shaham2021, chen2021, sitzmann2020siren}. Traditional methods  primarily use discrete representations like data grids that often grapple with limited spatial resolution and inherent discretization issues. Instead, INRs encode data in terms of continuous functional relationships. Thus, they are are resolution-independent, offering a novel framework for data representation. Furthermore, modifications to INRs, such as SIREN's periodic activations~\cite{sitzmann2020siren} and NeRFs' positional encodings~\cite{mildenhall2020nerf}, have successfully mitigated the spectral bias commonly faced by traditional neural networks. Therefore, the resolution free nature of INRs, in combination with their capacity for accurate spectral reconstruction makes them particularly useful in time series applications, where data irregularities and missing values are prevalent.
    
    In particular,~\cite{Fons2022HyperTimeIN} proposes HyperTime that leverages INRs with hypernetworks for time series generation. Given that each time series is represented by an INR, the hypernetwork allows it to learn a prior over the time series dataset, which generates a compressed latent representation of the entire time series dataset. The embeddings can then be interpolated to generate novel time series. The method was evaluated using two quantitative metrics: 1) the predictive score, which measures the \textit{usefulness} of the generated data by using a \textit{train on synthetic, test on real} (TSRT) approach where a model is trained using the synthetic data to predict the next step in a sequence, and then it is evaluated using the real data; and 2)  the \textbf{discriminative score} which serves as a measurement of \textit{fidelity} of the generated data, where the aim is to assess if the synthetic data is indistinguishable from real data. A discriminative model is trained to classify real and fake samples, and then used to test whether the original and generated data are correctly classified. The discriminative score is computed as $|\text{Accuracy}-0.5|$, where a low value means that the classification is challenging, and therefore, the model cannot tell which samples are real and which are generated. Results for multiple lengths of Google stock data are shown in Table~\ref{tab:inr_reg_results} where HyperTime outperforms all other benchmarks with regards to the predictive score and shows competitive performance with regards to the discriminative score.

    \begin{table}[]
        \centering
                \caption{Performance of regular time series generation in terms of the predictive and discriminative scores.}
        \label{tab:inr_reg_results}

    \def\arraystretch{1.1}
    % \resizebox{\linewidth}{!}{
      \begin{tabular}{clccc} %{lp{4em}p{9em}p{8em}p{3em}p{3em}}
\toprule
 & {Method}       &  \bf Stock24 & \bf Stocks72   & \bf Stock360   \\
\midrule\vspace{-1em}
\parbox[t]{1mm}{\multirow{8}{*}{\rotatebox[origin=c]{90}{\footnotesize Predictive Score}}} & & & & \\
& iHT &  \bf .037 $\pm$ .000 & \bf .188 $\pm$ .000 & \bf .168 $\pm$ .000  \\
% & iHT (full) & .254 $\pm$ .000 & .048 $\pm$ .001 &  .189 $\pm$ .000 &  .172 $\pm$ .001\\
\cmidrule{2-5}
& GT-GAN & .040 $\pm$ .000  & .207 $\pm$ .000 & .188 $\pm$ .000 \\
& TimeGAN &  {.038 $\pm$ .001} & .226 $\pm$ .002 & .206 $\pm$ .000 \\
& RCGAN &  .040 $\pm$ .001 & .192 $\pm$ .001 & .189 $\pm$ .000 \\
& DiffTime &  {.038 $\pm$ .001} & .213 $\pm$ .000 & .215 $\pm$ .000 \\
& LS4 & .103 $\pm$ .001 & .194 $\pm$ .000 & \bf .168 $\pm$ .000 \\
& FF & .076 $\pm$ .001 &  {.191 $\pm$ .000} &  {.169 $\pm$ .000} \\
\cline{2-5}
\rowcolor{gray!10} & Original & .036 $\pm$ .001 & .186 $\pm$ .001 & .167 $\pm$ .001\\
\midrule\vspace{-1em}
\parbox[t]{1mm}{\multirow{7}{*}{\rotatebox[origin=r]{90}{\footnotesize Discriminative Score}}} &  & & & \\
& iHT & \bf  .044 $\pm$ .011 & {.014 $\pm$ .009} & .018 $\pm$ .015 \\
% & iHT (full) & .321 $\pm$ .009 & .214 $\pm$ .001 & .027 $\pm$ .006 &    .017 $\pm$ .019 \\
% \midrule
\cmidrule{2-5}
& GT-GAN & {.077 $\pm$ .031} & .058 $\pm$ .017 & .085 $\pm$ .064 \\
& TimeGAN & .102 $\pm$ .021 & .073 $\pm$ .047 & .042 $\pm$ .074 \\
& RCGAN & .196 $\pm$ .027 & \bf .012 $\pm$ .09 & \bf .014 $\pm$ .007 \\
& DiffTime & .097 $\pm$ .016 & .097 $\pm$ .012 & .101 $\pm$ .018 \\
& LS4 & .363 $\pm$ .027 & .089 $\pm$ .081 & .088 $\pm$ .081 \\
& FF & .349 $\pm$ .113 & .016 $\pm$ .018 & {.015 $\pm$ .014} \\
\bottomrule
      \end{tabular}
    % }
    \end{table}

    \paragraph{\textbf{Application: Imputation of Financial Time Series}}
    
    Addressing time series imputation is crucial across different domains including finance, climate modelling, and healthcare, given the potentially vast differences in the type of data under study. 
    Classic strategies, such as averaging and regression, are generally too simplistic and fail to sufficiently encapsulate the underlying behavior. Although modern methods, like iterative imputation and maximum likelihood procedures, allow for a higher degree of algorithmic complexity and performance improvement, the assumptions they often make can introduce biases that are detrimental to intricate cases (refer to~\cite{Jarrett_HyperImpute} or~\cite{Cao_BRITS} for examples). Inspired by the success of~\cite{Fons2022HyperTimeIN} to generate time series that can be irregularly sampled,~\cite{Bamford2023MADSMA} proposes Modulated Auto-Decoding SIREN (MADS) for multivariate time series imputation. MADS utilises the capabilities of SIRENs for high fidelity reconstruction of signals and irregular data handling and combines the SIREN parameterizations with hypernetworks in order to learn a prior over the space of time series. Experimental results across three real-world time series datasets from different domains are shown in Table~\ref{tab:real_datasets_result_mads}, where MADS shows best performance in the majority of the datasets. Interestingly, in all cases, the best performance is achieved by INR-based methods. For details of the performance metrics and additional results, we refer to the original paper~\cite{Bamford2023MADSMA}.

\begin{table*}
\caption{Experimental results for time series imputation. Results are averaged over five runs, with the standard deviation shown in parenthesis (bold indicates best performance).}
\label{tab:real_datasets_result_mads}
\resizebox{\textwidth}{!}{%

% \multicolumn{2}{c}{Method}
\begin{tabular}{l|l|ccc|ccc|ccc|ccc}
    \toprule
\multicolumn{2}{c}{} & \multicolumn{3}{c}{\textit{Air Quality}} & \multicolumn{3}{c}{\textit{HAR ($30\%$)}} & \multicolumn{3}{c}{\textit{HAR ($70\%$)}} & \multicolumn{3}{c}{\textit{PhysioNet}} \\
Family & Method  & {MSE} & {Max MSE} & {W2} & {MSE} & {Max MSE} & {W2} & {MSE} & {Max MSE} & {W2} & {MSE} & {Max MSE} & {W2} \\
\midrule
\multirow{4}{*}{Classic}& Mean & 0.091 & 0.470 & 0.260 & 0.027 & 0.332 & 0.078 & 0.028 & 0.358 & 0.182 & 0.024 & 0.386 & \textbf{0.029} \\
 &     & (0.000) & (0.000) & (0.000) & (0.001) & (0.004) & (0.001) & (0.000) & (0.004) & (0.004) & (0.001) & (0.018) & (0.001) \\
& Median & 0.091 & 0.480 & 0.260 & 0.029 & 0.332 & 0.215 & 0.029 & 0.362 & 0.180 & 0.026 & 0.428 & 0.086 \\
&  & (0.000) & (0.000) & (0.000) & (0.000) & (0.004) & (0.305) & (0.000) & (0.004) & (0.000) & (0.001) & (0.022) & (0.125) \\
 \midrule
\multirow{6}{*}{Deep learning} & BRITS & 0.224 & 0.880 & 0.388 & 0.308 & 0.952 & 0.284 & 0.312 & 0.988 & 0.666 & 0.522 & 0.986 & 0.144 \\
       &    &      (0.015) &          (0.016) &  (0.022) &         (0.004) &          (0.004) &  (0.005) &         (0.004) &          (0.004) &  (0.005) &      (0.011) &          (0.011) &  (0.005) \\
& CSDI & 0.075 & 0.844 & 1.220 & 0.010 & 0.280 & 0.122 & 0.020 & 0.436 & 0.168 & 0.037 & 0.506 & 0.422 \\
  &          &      (0.012) &          (0.015) &  (0.045) &         (0.002) &          (0.021) &  (0.017) &         (0.006) &          (0.056) &  (0.031) &      (0.005) &          (0.035) &  (0.029) \\
& GP-VAE & 0.083 & 0.772 & 0.764 & 0.035 & 0.374 & 0.280 & 0.038 & 0.416 & 0.302 & 0.024 & 0.386 & 0.352 \\
  &        &      (0.003) &          (0.013) &  (0.015) &         (0.007) &          (0.027) &  (0.020) &         (0.007) &          (0.025) &  (0.026) &      (0.001) &          (0.019) &  (0.013) \\
\midrule
\multirow{7}{*}{INR} & SIREN+ & 0.085 & 0.504 & 0.226 & 0.007 & \textbf{0.178} & 0.106 & 0.008 & 0.270 & 0.123 & \textbf{0.017} & \textbf{0.368} & 0.270 \\
 &    &      (0.009) &          (0.034) &  (0.030) &         (0.002) &          (0.090) &  (0.018) &         (0.003) &          (0.023) &  (0.035) &      (0.001) &          (0.019) &  (0.014) \\
& HN+SIREN & 0.077 & 0.516 & 0.210 & 0.010 & 0.266 & 0.119 & 0.013 & 0.354 & 0.138 & 0.024 & 0.418 & 0.332 \\
  &     &      (0.009) &          (0.011) &  (0.029) &         (0.005) &          (0.077) &  (0.035) &         (0.006) &          (0.070) &  (0.043) &      (0.002) &          (0.026) &  (0.023) \\
& Mod-SIREN & \textbf{0.070} & 0.480 & 0.214 & 0.007 & 0.210 & 0.100 & 0.010 & 0.276 & 0.121 & 0.018 & 0.378 & 0.286 \\
&  &      (0.007) &          (0.016) &  (0.036) &         (0.002) &          (0.010) &  (0.016) &         (0.010) &          (0.047) &  (0.072) &      (0.002) &          (0.029) &  (0.023) \\
 & MADS & 0.072 & \textbf{0.458} & \textbf{0.202} & \textbf{0.005} & 0.186 & \textbf{0.072} & \textbf{0.006} & \textbf{0.252} & \textbf{0.082} & 0.018 & 0.372 & 0.276 \\
       &  &      (0.012) &          (0.016) &  (0.028) &         (0.000) &          (0.005) &  (0.003) &         (0.001) &          (0.011) &  (0.005) &      (0.001) &          (0.018) &  (0.013) \\
\bottomrule
\end{tabular}
}
\end{table*}

    %\vp{is this complete or will you add some results here?}
%%%\EF{I added results in this section, I can add more detail if you want)}
%%%
%%%
%%%    \paragraph{\textbf{Application; Improving performance of forecasting algorithms by time synthetic time series augmentation (Yousef)}} {\color{red} add also Elizabeth phd work and Yousef ICAIF paper}
%%%
%%%    {\color{green}SV: Yousef - could you please populate}
%%%
%%%    
%%%    
%%%    {\color{green}SV: techniques used for augmentation (might not necessarily be driven by realism) - reference to Elizabeth's prior papers!!!!}
%%%
    \paragraph{\textbf{Application: Constrained Market Scenarios Generation }} 
    % Application / Problem
    Synthetic time-series are extremely useful to test hypotheses and algorithms before employing them in real settings, especially for unseen and counterfactual scenarios. For instance, the US Federal Reserve publishes synthetic market stress scenarios given by the constrained time series for financial institutions to assess their performance in hypothetical recessions~\cite{federalreserve}. We refer to the constrained time-series generation problem as the problem of generating synthetic time-series that are statistically similar to historical times series while matching some input constraints~\cite{di2020efficient,xu2018semantic,guimaraes2017objective}. 
    %\vp{this seems like a very important problem. Can we add more citations or scenarios where it was considered in the literature?}
    
    % Existing Approach
    Existing work that employs deep generative models  attempts to capture the statistical data properties and temporal dynamics of time-series, neglecting additional requirements, such as input constraints. These constraints are often introduced by re-training the existing generative models, and penalizing them proportionally to the mass they allocate to invalid data~\cite{takeishi2021knowledge,di2020efficient,guimaraes2017objective,xu2018semantic}. Other works consider unconstrained generative models to generate synthetic markets while rejecting and re-sampling time-series that do not match the constraints~\cite{7796926}.

    \begin{figure}[t]
    \centering
    \subcaptionbox{Unconstrained}{\includegraphics[width=0.20\textwidth]{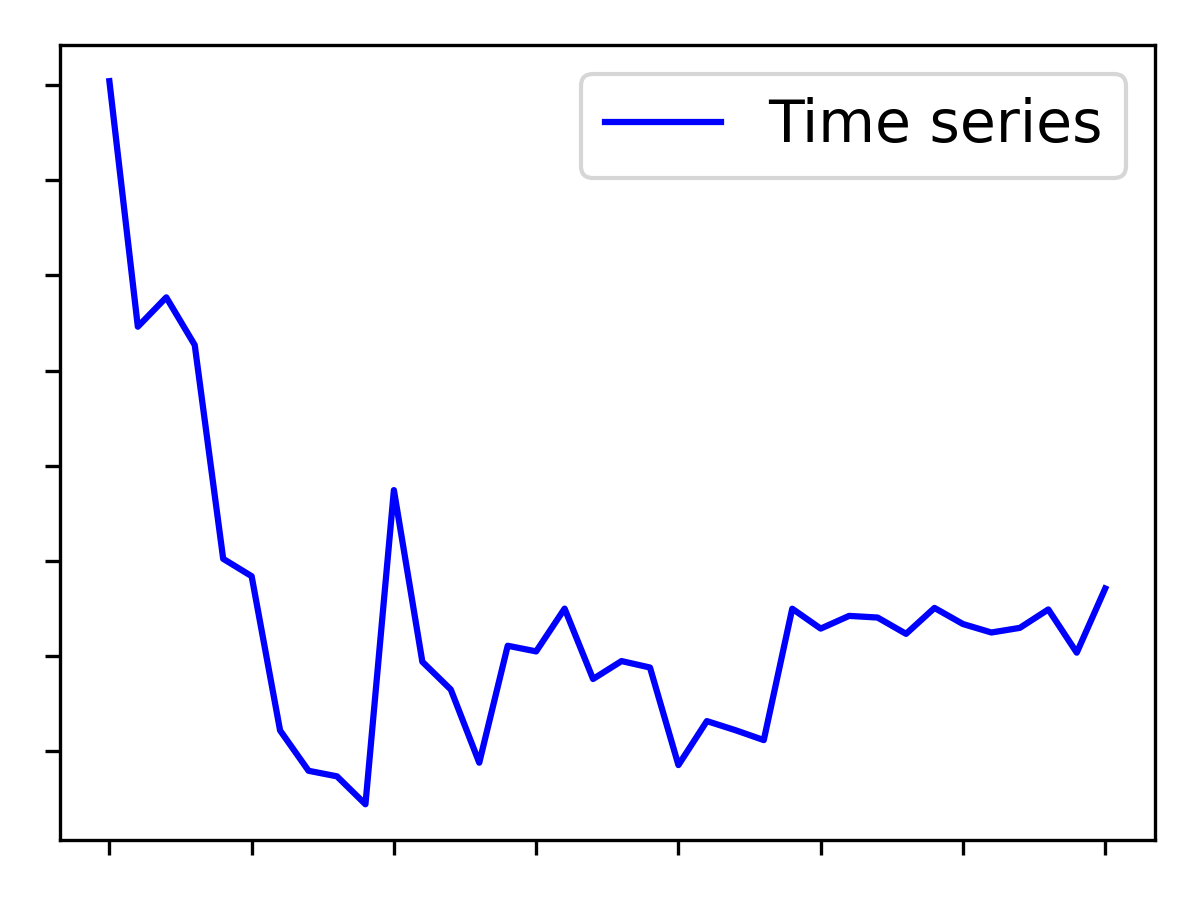}}%\label{fig:unc_example}%
    \hfill
    \subcaptionbox{Trend}{\includegraphics[width=0.20\textwidth]{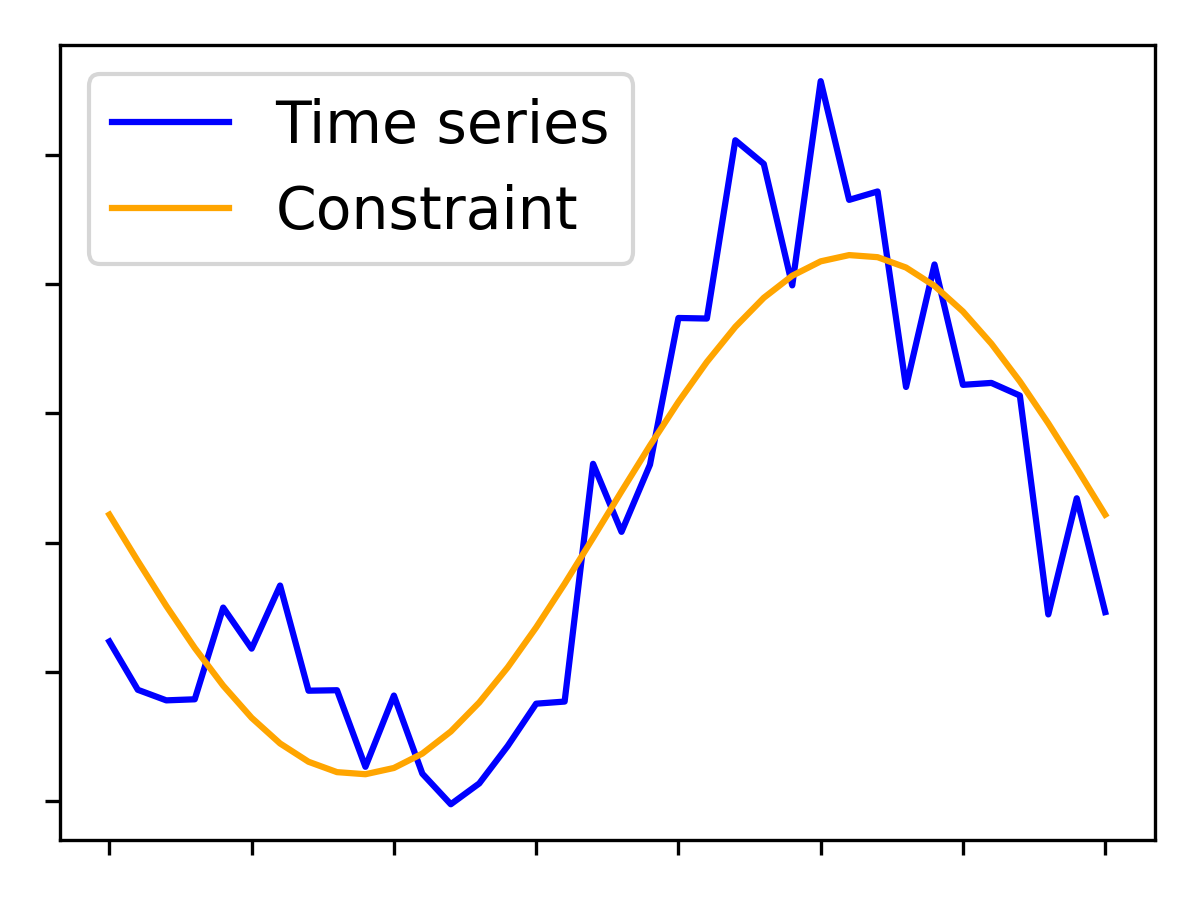}}%\label{fig:trend_example}%
    \hfill
    \subcaptionbox{Fixed value}{\includegraphics[width=0.20\textwidth]{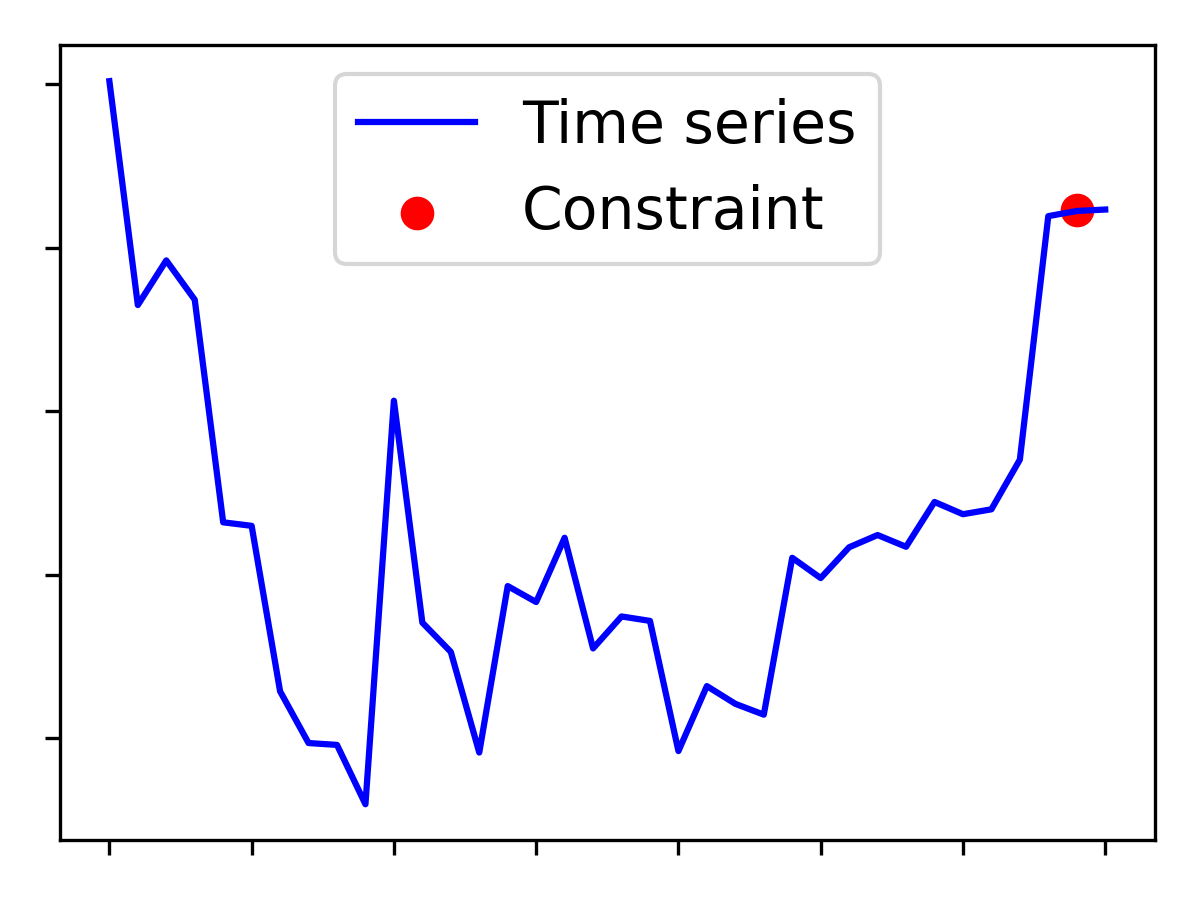}}%\label{fig:point_example}%
    \hfill
    \subcaptionbox{Global min}{\includegraphics[width=0.20\textwidth]{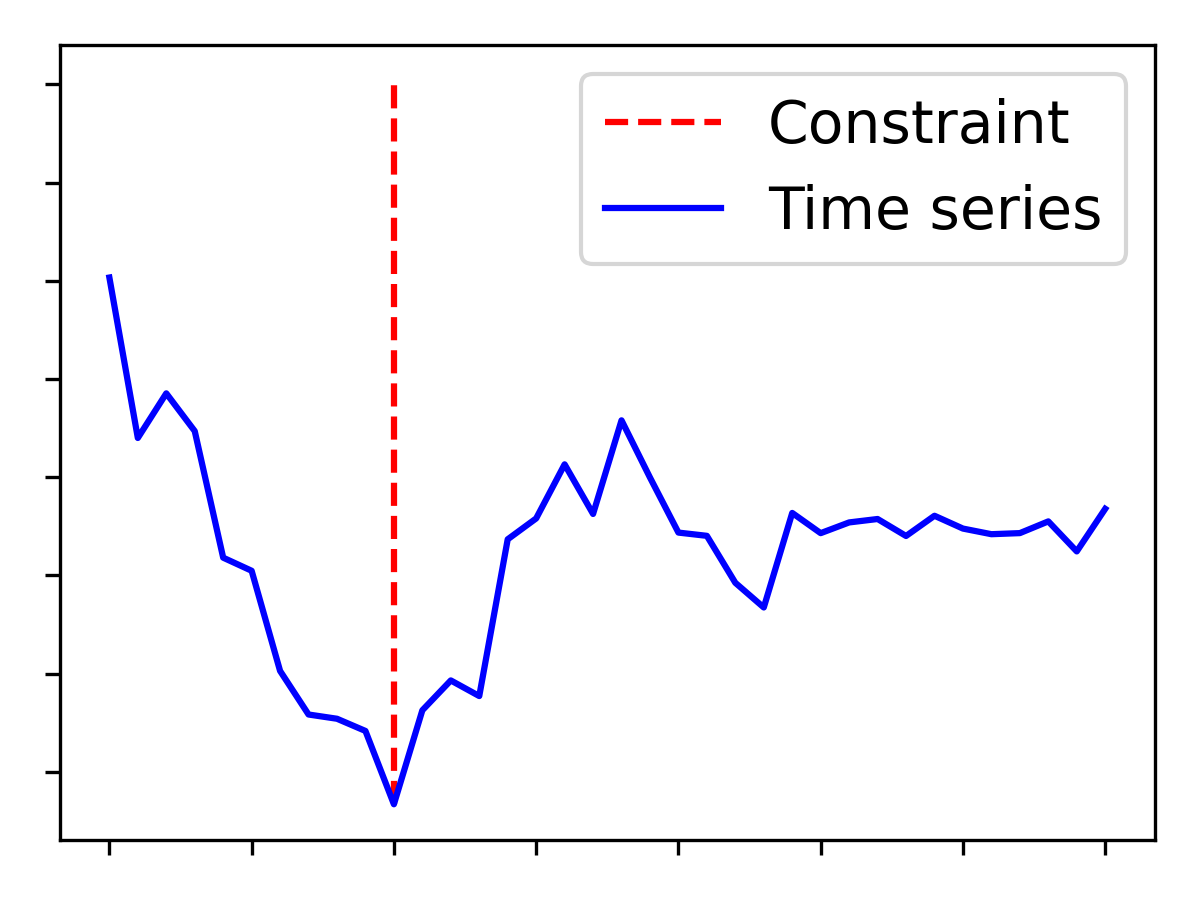}}%\label{fig:global_example}%
    \hfill
    \subcaptionbox{Multivariate}{\includegraphics[width=0.20\textwidth]{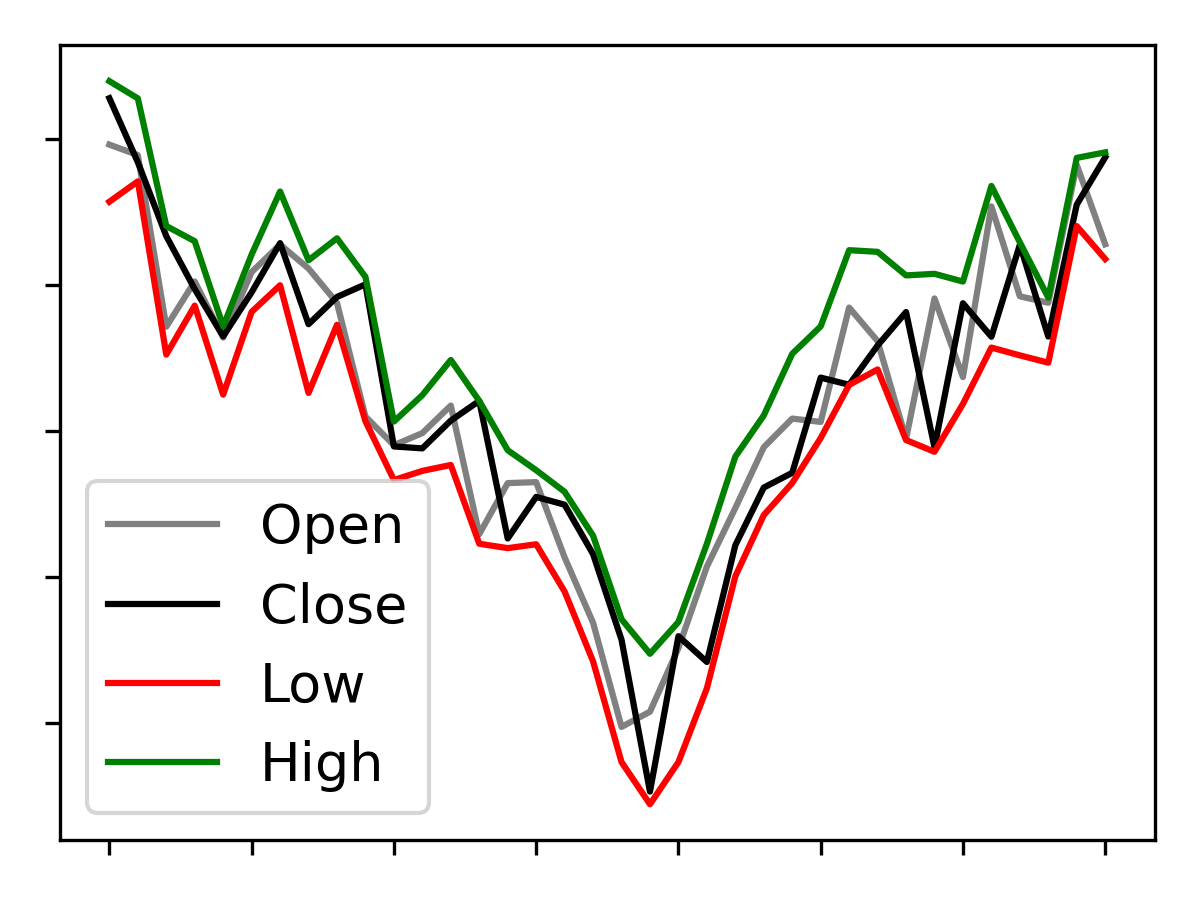}}%\label{fig:multivariate_example}%
    \caption{An example of synthetic stock market time-series under different constraints: (a) unconstrained generation; (b) a time-series following a trend constraint; (c) the final value of the TS has to hold a specific value; (d) the global minimum must be at a given time point; (e) multi-variate TS where the \textit{High} and \textit{Low} dimensions have the maximum and minimum values, respectively.
    }\label{fig:example_constraints}
    \end{figure}
    
    % Approach
    A formal definition of constrained time-series generation is introduced in~\cite{coletta2023constrained}. This work divides the constraints in \textit{Soft} and \textit{Hard} constraints, and additionally it divides them between \textit{Global} and \textit{Local} constraints. \textit{Soft} constraints optimize the time-series to minimize a constraint score (i.e., minimize the distance between the time-series and the constraint), and do not require sample rejection. \textit{Hard} constraints require the time-series to exactly match the input-constraints, and a time-series can be rejected if it does not respect a constraint. 
    Additional, \textit{Global} constraints compare across all the points in the time-series, while \textit{Local} constraints focus on specific sub-sections of the input time-series. 
    This work proposes four different approaches to tackle the constrained generative problem. It starts from a constrained-optimization (COP) framework, and then it proposes three different variant of denoising diffusion models. 
    The COP framework provides to the user the possibility to specify any constraints, and generate synthetic time-series starting from random noise or existing time-series. This method explicitly includes also all the statistical data properties as constraints, e.g., auto-correlation and returns for financial time-series. 
    Thus, it enables the users to constrain the time-series and also to control their statistical properties. 
    Instead, the diffusion models learn the statistical data properties directly from input data, and they can integrate additional constraints during training or inference. In particular, one promising diffusion model, called \textit{Guided-DiffTime}, shows the ability to incorporate any differentiable constraints into the sampling procedure. This model does not require to be re-trained when incorporating new constraints. The proposed approach outperform existing state-of-art benchmarks under different quantitative and qualitative metrics.

    %Results 
    Figure~\ref{fig:example_constraints} shows an example of generated time series from the work in~\cite{coletta2023constrained}. The figure shows the model ability to generate multiple different scenarios which can be used to further study and test investment strategies, and perform stress tests. To evaluate the model performance upon different constraints, the authors introduce several new metrics: the \textbf{L2 distance} which measures how much the synthetic data follows a trend \textit{soft}-constraint by evaluating the distance between the TS and the trend constraint using L2 norm; the \textbf{satisfaction rate} which measures the percentage of time a synthetic TS meets the input \textit{hard}-constraints; and the \textbf{inference time} and \textbf{fine-tuning time}, which measure the average time required to generate a new valid sample, and the time required to enforce constraints over invalid samples by fine-tuning them. The proposed approaches establish a new state-of-art performance on constrained time-series generation. We report the performance for the trend soft-constraint in Table~\ref{table:soft_constr}, while additional results can be found in the original paper~\cite{coletta2023constrained}. %and the hard-constraints performance in Table~\ref{table:hard_constr}. The tables  report also the TSTR (i.e., Discriminative and Predictive score) evaluation introduced in Section~\ref{sec:metrics} .  

\begin{table}[hbt]
\centering
\caption{Soft Constraints (Trend) Time-Series Generation (Bold indicates best performance).}\label{table:soft_constr}
\resizebox{0.6\linewidth}{!}{
\begin{tabular}{|l|c|c|c|c|}
\hline
Algo & Discr-Score & Pred-Score & Inference-Time & L2 Distance \\ \hline
COP (Ours) & \textbf{0.01±0.01} & \textbf{0.20±0.00} & 0.73±0.05 & \textbf{0.015±0} \\
DiffTime (Ours) & \textbf{0.01±0.01} & \textbf{0.20±0.00} & 0.02±0.00 & 0.018±0 \\
GT-GAN & 0.04±0.03 & 0.22±0.00 & \textbf{0.00±0.00} & 1.378±2 \\
TimeGAN & 0.02±0.02 & \textbf{0.20±0.00} & \textbf{0.00±0.00} & 0.073±0 \\ 
RCGAN & 0.02±0.01 &\textbf{0.20±0.00}& \textbf{0.00±0.00} & 0.071±0 \\ \hline
\end{tabular}}
\vspace{-0.1in}
\end{table}

\paragraph{\textbf{Application: Stylized Generation of Financial Time Series}}
    
    When it comes to generating realistic financial time series data, traditional generative models and simulation approaches often struggle to produce accurate representations of real-world market dynamics. Mainly, they fail to satisfy specific statistical properties, known as stylized facts, which are essential for accurately modeling financial phenomena. Stylized facts encompass various characteristics observed in financial time series, such as volatility clustering, fat-tailed distributions, and long-range dependencies. Ensuring that generative models and simulations satisfy all stylized facts simultaneously can be a challenging and computationally intensive task. 

    To overcome the limitations of traditional generative models and simulations, style transfer techniques have been proposed to enhance the realism of financial time series data~\cite{el2022styletime}. Style transfer, a concept originally developed in the field of computer vision, involves modifying the style or appearance of an image while preserving its content. In the context of financial time series, style transfer techniques can be employed to enhance synthetic data by imbuing it with specific stylized facts observed in real financial markets. By transferring the statistical characteristics and patterns from real financial time series to synthetic data, style transfer can bridge the gap between simulated and real-world data, resulting in more realistic representations of financial dynamics. This approach offers a promising avenue for creating reliable financial simulations that not only capture the overall structure and trends but also satisfy critical stylized facts. 

    Specifically, in~\cite{el2022styletime}, a style transfer technique called \emph{StyleTime} was introduced so to stylize synthetic time series data with statistical properties of real time series data. The technique was evaluated along three different dimensions:
    \begin{enumerate}
        \item \emph {Fidelity}: Synthetic data fidelity is quantified using precision/recall metrics from~\cite{sajjadi2018assessing} and by using the TSTR framework~\cite{esteban2017real} to assess the predictive utility of the synthetic time series.
        \item \emph {Predictive utility}: Stylized synthetic data is used to augment a historical time series (training) dataset and the predictive utility is measured by evaluating the percentage of improvement in time series forecasting in terms of MAE. 
        \item \emph {Authenticity}: An authenticity score, proposed in~\cite{alaa2021faithful}, is used as a means to assess similarity of the generated synthetic data from the original training data. A low authenticity score ($A=0$) means that the synthetic data is an exact replica of the training data, while a high authenticity score $(A=1)$ means that the synthetic data differs a lot from the original training samples.  
    \end{enumerate}
Results in~\cite{el2022styletime} indicate that by stylizing synthetic time series data with the statistical properties of historical data, we can see improvement across all of these metrics. We repeat the results for Google stock data here for completeness in Table~\ref{tab: style_transfer_results}.

\begin{table*}[t]
\centering
\resizebox{\textwidth}{!}{
\begin{tabular}{c|cc|c|c|}
\cline{2-5}
\textbf{}                                                    & \multicolumn{2}{c|}{\textbf{Fidelity}}                                                  & \textbf{Predictive Utility}     & \textbf{Authenticity}           \\ \cline{2-5} 
\textit{}                                                    & \multicolumn{1}{c|}{\textit{F-Score}}                 & \textit{TSTR MAE}               & \textit{Augmentation MAE}       & \textit{Authenticity Score}     \\ \hline
\multicolumn{1}{|c|}{\textbf{Fourier Flows}}                 & \multicolumn{1}{c|}{$0.9813 \pm 0.0010$}              & $0.0079 \pm 0.0022$             & $0.0057\pm0.0011$               & $0.9760\pm0.0025$               \\ \hline
\multicolumn{1}{|c|}{\textbf{Stylized Data (Perturbed)}}     & \multicolumn{1}{c|}{${\bf 0.9971} \pm {\bf 0.0002}$} & ${\bf 0.0057} \pm {\bf 0.0010}$ & ${\bf 0.0054} \pm {\bf 0.0004}$ & ${\bf 0.9943} \pm {\bf 0.0017}$ \\ \hline
\multicolumn{1}{|c|}{\textbf{Stylized Data (Fourier Flows)}} & \multicolumn{1}{c|}{$0.9827\pm 0.0099$}               & $0.0089 \pm 0.0022$             & $0.0058 \pm 0.0013$             & $0.9879 \pm 0.0013$             \\ \hline
\end{tabular}}
\caption{Experimental results across fidelity, predictive utility, and authenticity metrics for stylized synthetic time series data for the Google stock. These results are extracted from~\cite{el2022styletime}; for more insight on how the experiments were conducted and comparison to other benchmarks please reference the original paper.   }
\label{tab: style_transfer_results}
\end{table*}

    \paragraph{\textbf{Discussion: Uncertainty Quantification in Time Series Augmentation Methods}}

    Machine learning models that quantify uncertainty are of paramount importance for financial time series. In particular, uncertainty quantification in time series forecasting applications allows one to reason about the volatility of time series, the probability of extreme events (heavy-tailed phenomena), and the occurrence of a distributional shift. As previously mentioned, a major use-case for synthetic time series data is data augmentation for improving the generalization of supervised learning models, particularly for deep learning models. In most data augmentation formulations, the focus is on devising a data augmentation technique to improve the average predictive performance of a model (e.g., augmenting samples to the training dataset that would reduce the mean-squared error of the model on the test dataset).  There are many candidate probabilistic deep learning techniques that can account for predictive uncertainty in supervised learning settings (e.g., variational inference~\cite{graves2011practical}, Monte Carlo dropout~\cite{gal2016dropout}, deep ensembles~\cite{lakshminarayanan2017de}, deep Gaussian mixture ensembles~\cite{el2023deep}). However, there has been little effort on studying the impact of data augmentation on uncertainty estimates produced by these techniques. For example, does augmentation of synthetic data impact the model's ability to reliably account for rare events (i.e., change the tail behavior of predictive model) or reliably detect out-of-distribution inputs (i.e., underestimation of epistemic uncertainty)? Recent works for large-language models have revealed that repetitive training on synthetic examples (based on the distribution of training data) can cause the model to effectively lose its ability to account for heavy-tailed phenomena. An important line of research to consider is making synthetic data augmentation robust, so that important statistical properties of an uncertainty quantifying model's predictive distribution remain intact after data augmentation.

\subsection{Simulation}\label{sec:simulation_ts}
An extremely useful application of synthetic time series concerns the simulation of financial markets. Simulated markets can be used to train and test investment strategies, representing an invaluable alternative to historical market data. 
% a brief introduction to what we are going to discuss
Different from historical data, simulation can provide more variability of the market scenarios reducing possible "time-period bias" (i.e., overfitting to a particular history that was encountered). Synthetic data can be used to simulate counterfactual market scenarios, where test strategies and algorithms before approaching the real market. Similarly, simulated markets can test the robustness of various deep learning approaches upon temporal covariate shifts. Finally, simulated markets can also react to the presence of any investment strategies, providing a realistic price impact for exogenous orders~\cite{coletta2023conditional,coletta2022learning}. 

\paragraph{\textbf{Introduction to Multi-Agent Simulation: ABIDES}}
Agent-Based Interactive Discrete Event Simulation environment (ABIDES) serves as an advanced multi-agent system (MAS) specifically for simulating complex financial markets~\cite{byrd2019abides,byrd2019explaining,balch2019evaluate}. ABIDES offers a highly configurable messaging system as its core engine, enabling individual agents, each representing a market participant, to communicate via routed messages with controlled latency and nanosecond resolution. The system replicates the dynamics of continuous double-auction trading, resembling real-world markets like NASDAQ. A distinguishing feature of ABIDES is its capability to assign a unique trading strategy to each agent, ranging from standard rule-based strategies to sophisticated reinforcement learning-based approaches.

\begin{figure*}[h]
    \centering
    \subfloat[IBM: September 30, 2008]{
        \includegraphics[width=2.5in]{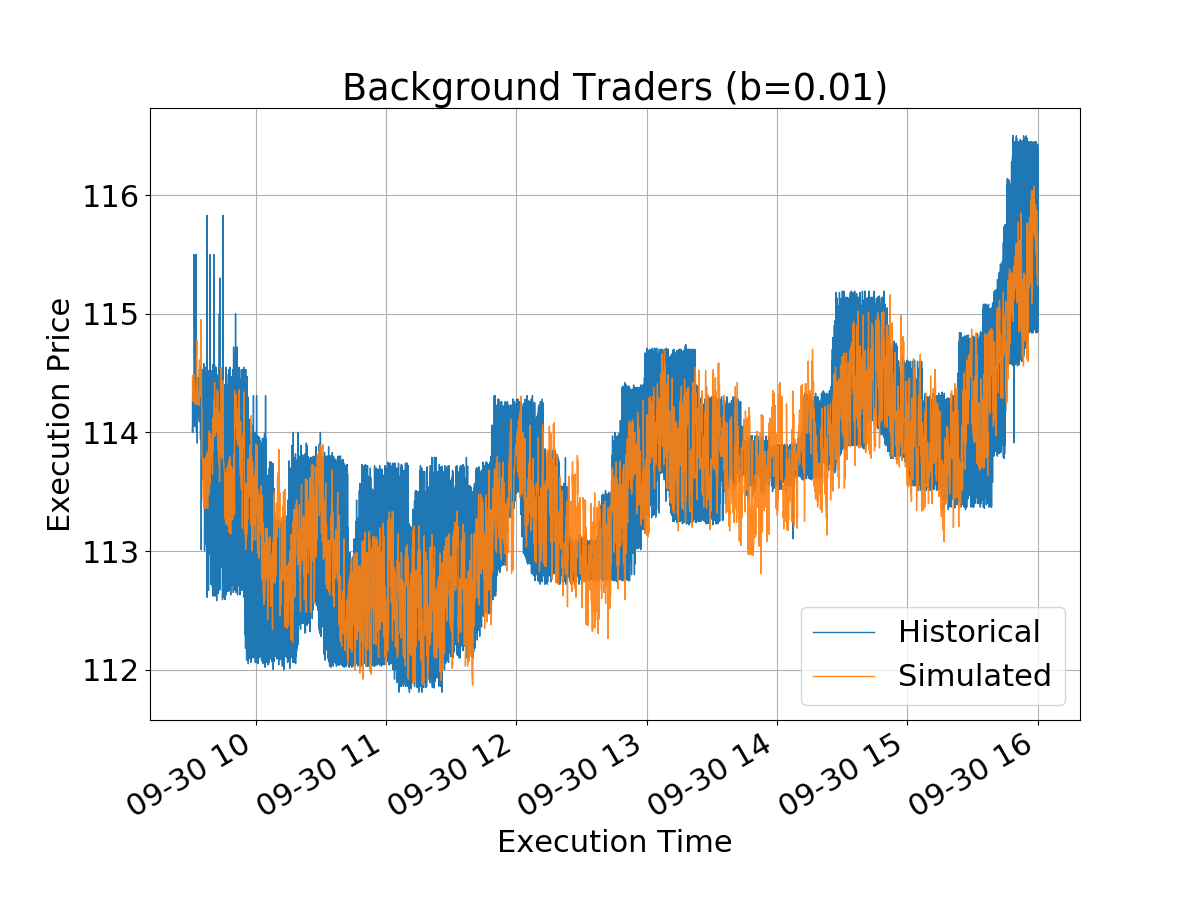}
    }
    \subfloat[MSFT: June 24, 2016]{
        \includegraphics[width=2.5in]{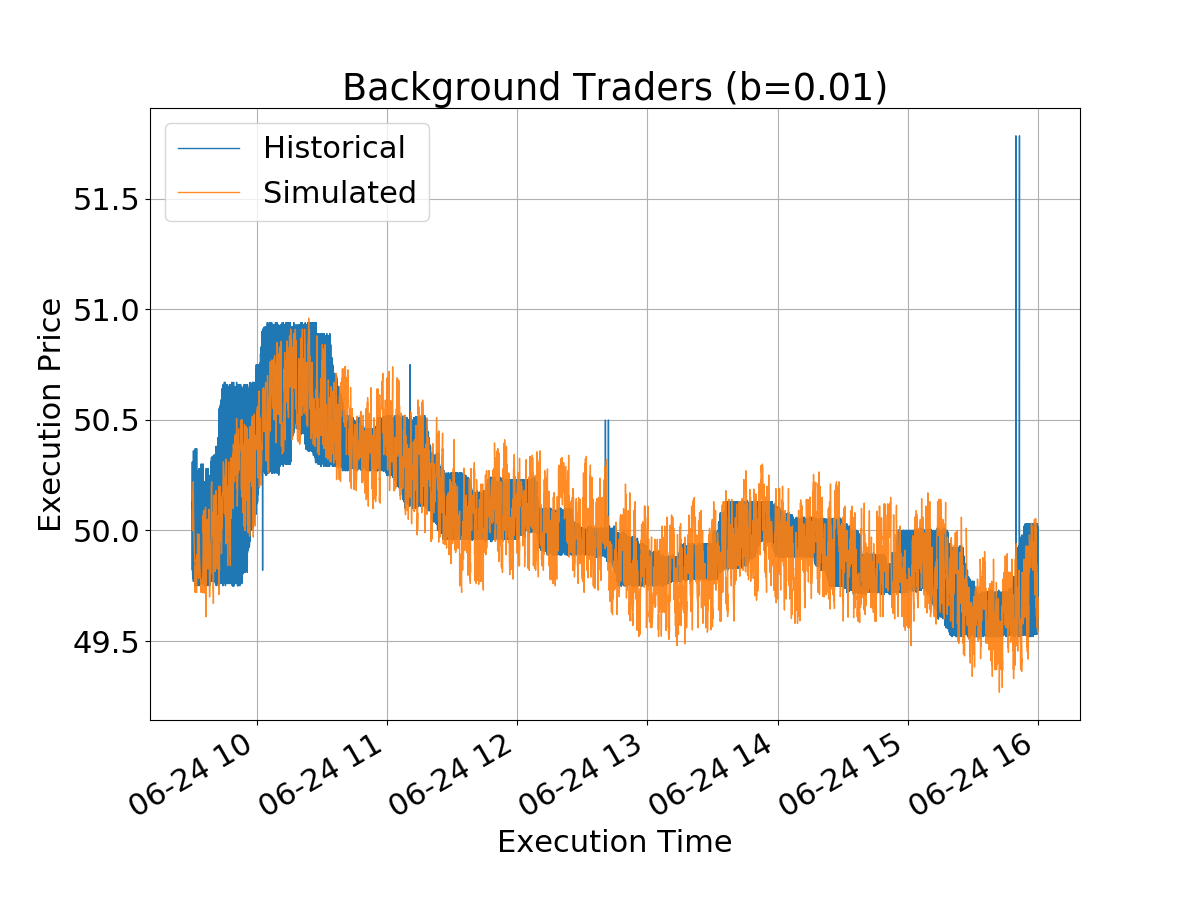}
    }
    \caption{ABIDES Simulated trades versus historical trades on two days.}
    \label{fig:background_abides}
\end{figure*}

ABIDES' architecture permits an in-depth exploration of market dynamics with various agents. Specifically, the exchange agent is designed to handle all transactions, processing buy and sell orders from other market participants. Upon receipt, the exchange agent matches and executes orders, updating the status of the order book. This allows agents to monitor the fluctuations in stock prices and observe the evolution of the order book. ABIDES also facilitates the configuration of an exogenous time series representing the ``fundamental" price for each stock, based on the mean-reverting Ornstein-Uhlenbeck process, reflecting the perception of the real world~\cite{uhlenbeck1930theory}. Figure~\ref{fig:background_abides} shows the synthetic market generated through ABIDES (orange lines) compared to actual intra-day price on two separate days in history. In such scenarios ABIDES uses 100 interacting background agents, calibrated according to common strategies. We can observe that the price history closely resembles the day in history, with similar statistical properties.

\paragraph{\textbf{Application: Orderbook market simulation}}
% Application 
Orderbook market simulation offers a novel alternative to the traditional backtesting, as it provides an interactive environment which allows us to study the market response (e.g., price impact~\cite{bouchaud2018trades}) to any experimental trading strategy.  Moreover, orderbook simulation can be used to generate thousand of synthetic markets, with more variability and without sensitive contents. 

% Approach
%A natural bottom-up approach to simulate financial markets is to model and simulate the real underlying system through a multi-agent simulation, described in Section~\ref{sec:sim_methods}. A state-of-art financial market simulator is ABIDES~\cite{byrd2019abides}. ABIDES is an Agent-Based Interactive Discrete Event Simulation supporting the simulation of tens of thousands of trading agents, which are defined with diverse strategies to mimic the real financial traders (see Figure \ref{fig:RDDL}). By interacting with an exchange agent, those agents simulate the real market evolution and generate synthetic intraday limit-order book data. ABIDES enables also to configure pairwise network latencies between each individual agent, as well as the exchange.

\begin{figure}[!ht]
    \centering
    \includegraphics[width=0.45\textwidth]{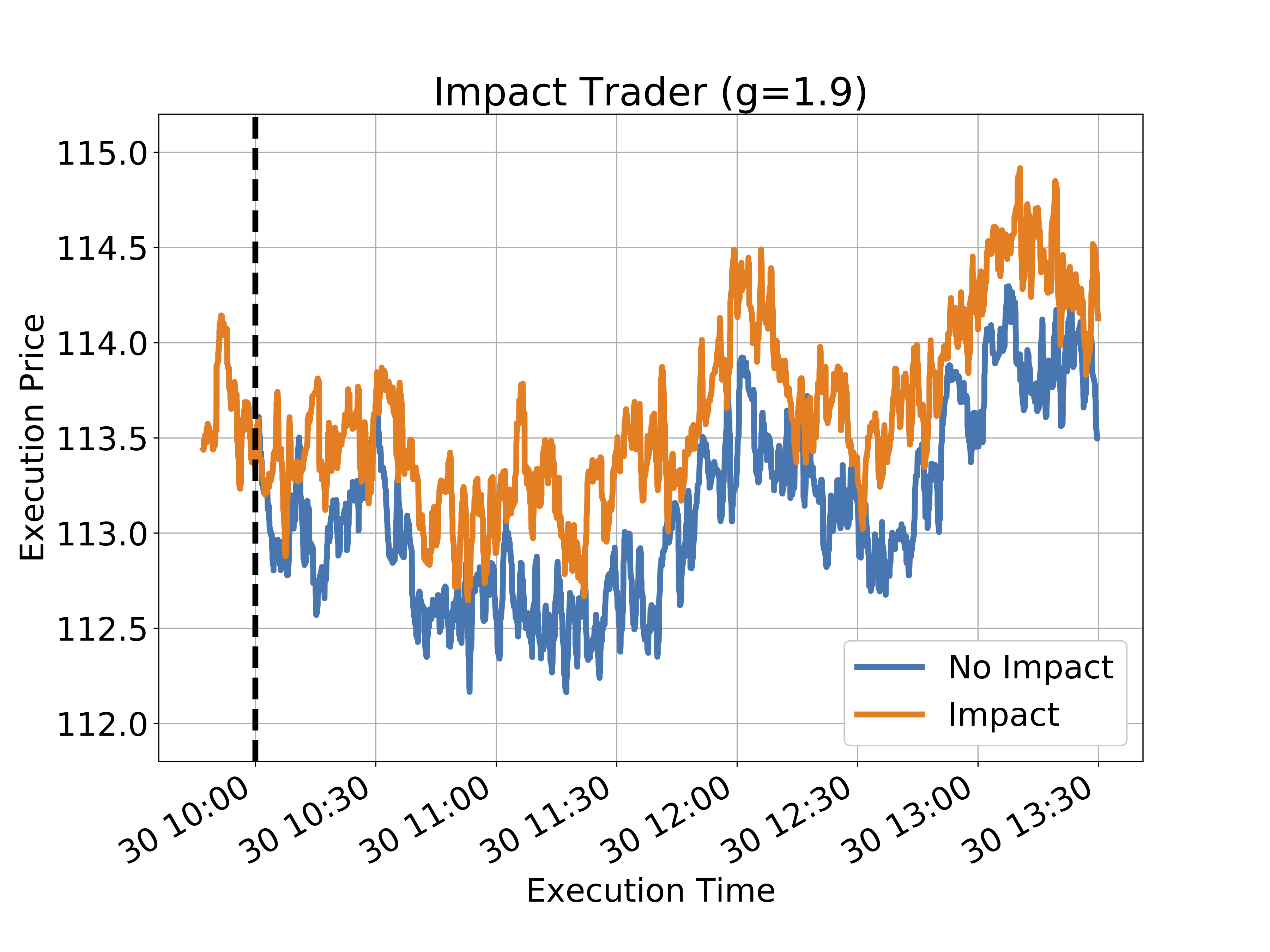}
    \includegraphics[width=0.42\textwidth]{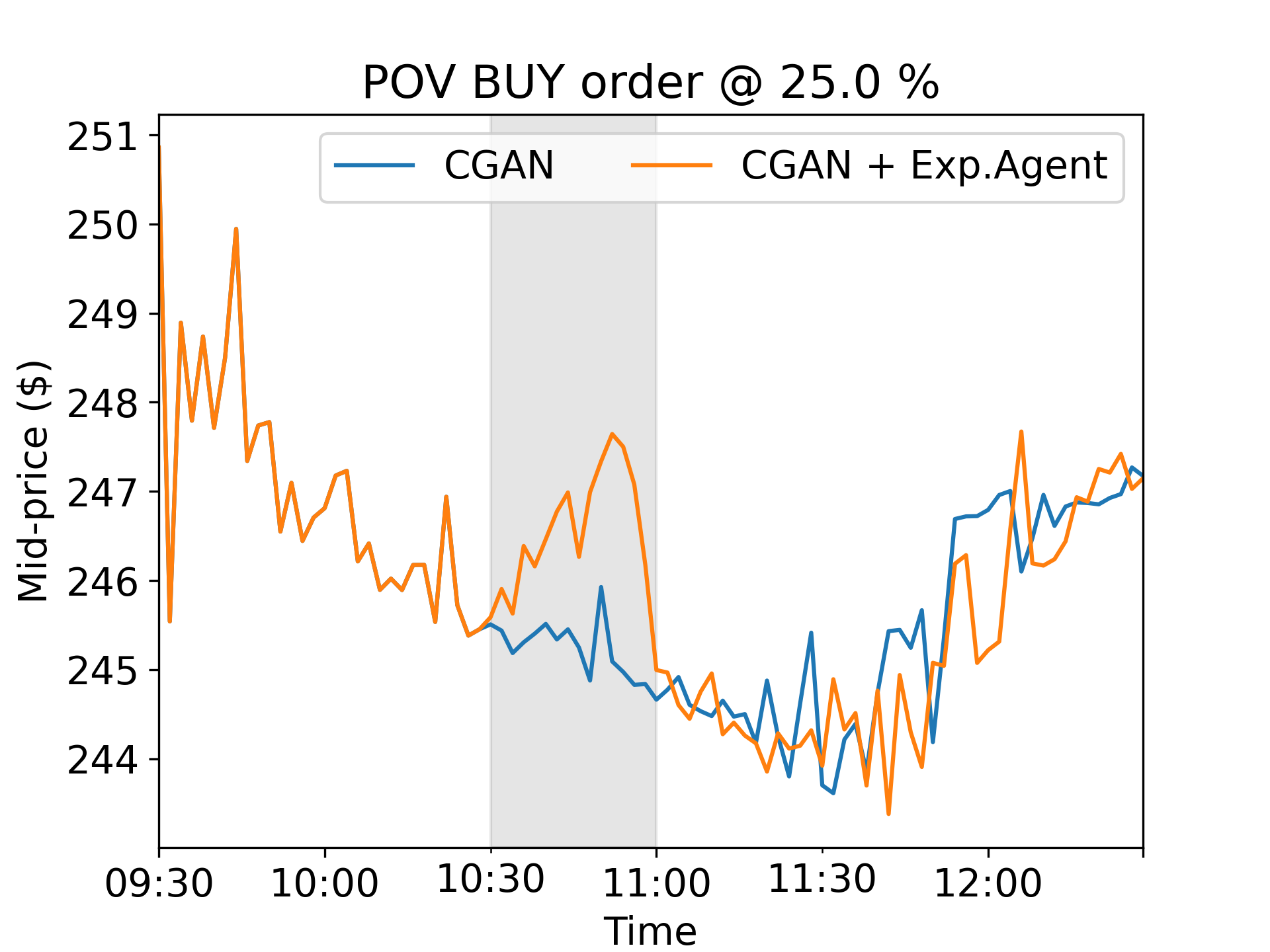}
    \caption{Example of price impact with an without an Experimental trading agent. For both the multi-agent simulator ABIDES (left) and the Conditional Generative Adversarial Network (right), the simulation changes and the price goes up after the experimental agent places multiple buy orders.}
    \label{fig:impact_simulation}
\end{figure}

%Approach 2 
Another recent approach uses conditional generative models for order book simulation~\cite{shi2023neural,coletta2022learning,prenzel2022dynamic,coletta2021towards,mizuta2016brief}. This approach has drawn recent attention thanks to its  high performance and ease of use. In fact, one major limitation of multi-agent simulators is their difficult calibration: to obtain realistic simulations, we need to identify  the correct number of agents and their strategies. 
%Most of existing multi-agent simulators approaches rely on pools of hand-crafted trading agents from financial literature (e.g., Zero-Intellingence agents~\cite{farmer2005predictive}), that reasonable mimic the complex real market dynamics. 
Instead, deep generative models can learn the different traders’ behaviors directly from real market data-sets~\cite{coletta2023k}. %and they are able to mimic all the agents and thus the whole market. 
In particular, recent work considers the whole market as a unique "world agent", which can be learnt from the anonymous historical data as a deep conditional generative model, and subsequently used to simulate realistic interactions of experimental trading agent with the market~\cite{coletta2021towards,coletta2022learning}. The synthetic generation can be conditioned upon the incoming orders of the exogenous investment strategy, and the generated time-series exhibit realistic price impact. 

% Results 
As results, these simulation environments enable further development of the investment strategies, providing an invaluable instrument to train, test and pursue "what if" studies. In Figure~\ref{fig:impact_simulation} we show an example of price impact that can be simulated using financial simulation. We consider exogenous trading strategies that place buy orders, and we show the price series generated using ABIDES (right chart) and the one generated using a deep generative model (left chart). This  "what if" market simulation analysis cannot be implemented in market replay or in a real financial system as the two situations (i.e., the exogenous orders arriving or not) are mutually exclusive.

\paragraph{\textbf{Application: Benchmarking Forecasting Algorithms to Distributional Shifts}} 

%Yousef
    Agent-based models are a common technique for simulating complex systems. For example, as previously mentioned, ABIDES can be used to simulate equities markets and provide intraday limit-order book data as a result of trading agents interacting with an exchange agent. One of the useful properties of simulation techniques such as ABIDES is that they can provide synthetic data to benchmark the performance of machine learning algorithms under different conditions that are not encountered in historical data. 

    For example, in~\cite{cao2022dslob}, ABIDES is utilized to simulate intraday limit-order book data under different market shock conditions. Using the simulated data, a test can easily be run to test the robustness of various deep learning approaches to temporal covariate shift on the task of midprice forecasting. The finding of the paper was that state-of-art techniques, such as DeepLOB~\cite{zhang2019deeplob}, were not robust to distributional and performed much better on unshocked market scenarios. Results for this experiment were conducted in~\cite{cao2022dslob}, but we repeat them here for convenience in Table~\ref{tab: dist_shift}.

    %\begin{wraptable}{l}{0.65\textwidth}
\begin{table*}[t]
%\vspace{1cm}  
% \setlength{\abovecaptionskip}{0.3cm}   
% \setlength{\belowcaptionskip}{0.0cm}
\small
\centering

\begin{tabular}{lccc}

\hline \hline
            & No Shock & Small Shock & Large Shock \\ \hline
 AdaRNN    & 1.02 $\pm$ 2.24e-4 & 1.00 $\pm$ 8.07e-5 & 0.99  $\pm$3.58e-5 \\ \hline
 Transformer & 0.87 $\pm$ 1.98e-3  & 1.02 $\pm$ 6.29e-3 & 1.08  $\pm$ 0.01 \\ \hline
 DeepLOB     & 0.66 $\pm$ 0.11 & 1.08 $\pm$ 0.07 &	2.25 $\pm$ 0.15  \\ \hline \hline

\end{tabular}
\caption{Forecasting model performance in terms of RMSE on the synthetic LOB dataset generated from ABIDES.} %under IID setting and OOD setting

\label{tab: dist_shift}
\end{table*}

We have discussed various aspects of synthetic time-series and their myriad applications to the financial domain. Our discussion on the various modalities concludes with the next section on unstructured data.
%\end{itemize}

%\subsubsection{generation}

%HyperTime, ensemble methods..

%\subsubsection{privacy}

\section{Unstructured data}
\label{sec:unstructured}
Many applications in a financial institution involve data arising from text, images, and documents. In particular, we focus on two applications from checks and client communications. 
%\vp{Add in models for the unstructured space and how it differs from typical tabular approaches.}

\subsection{Handwriting}
\label{sec:image}

%%%\vp{please include all the approaches we have explored such as scrabblegan, style equalitation, diffusion models etc
%%%Talk about the various approaches in the literature. We can give the application with font based but could also show the issues with standard approaches on internal dataset. 
%%%Open questions for the community? }

Optical Character Recognition (OCR) tasks transcribe images containing text (possibly handwritten) to machine-encoded text. OCR has multiple applications in automated data entry and information extraction, e.g., scanning of forms, checks, and documents. Commercial solutions are based on Convolutional Recurrent Neural Networks (CRNN)~\cite{hegghammer2022ocr, shi2016end}. These models are typically trained in a supervised fashion, with datasets consisting of \texttt{<image, ground\_truth\_text>} pairs. However, OCR datasets – particularly those of handwritten text – can be noisy with incorrect annotations and artefacts in the images. Additionally, multiple types of data within the firm are sensitive, and cannot be used to train models (e.g., PII considerations with addresses), causing downstream performance degradation. Synthetic Data offers a way of augmenting these datasets so as to possibly boost the performance of OCR models. 
%\vp{can we add a bit more detail on online and offline models? Also, references would be helpful for a quite a few statements in this section.}

\begin{figure}[!ht]
    \centering
    \includegraphics[width=\textwidth]{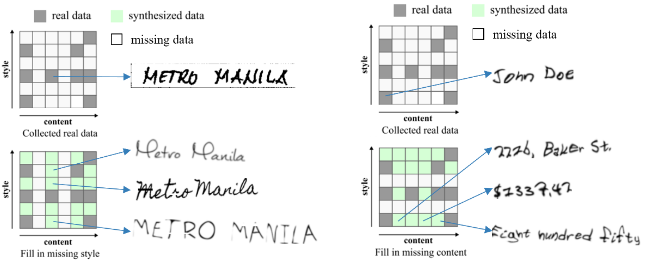}
    \caption{Content and Style Dimensions of Handwriting Data: augmenting by filling in missing styles (left) and augmenting by filling in missing content (right) (adapated from~\cite{chang2022data}. %\srijan{TODO: figure attribution?} \vp{find the relevant paper of oncel's group}
    }
    \label{fig:content_style}
\end{figure}

Handwriting data have two dimensions that are of particular interest: content, and style. Contents refers to the text contained in the image. Textual contents can be often be grouped into categories like numerals, words, alphanumeric, etc. Style refers to the manner in which the text is written, e.g., cursive or bold, along with author-specific intricacies. When generating synthetic handwriting data, we must decide which dimensions we wish to augment along – introduce new contents, capture novel handwriting styles, or both (see Figure~\ref{fig:content_style} for a schematic example).

OCR models can be hard to train because of challenges with datasets of handwritten text. These datasets can suffer from label noise (incorrect labels), and image noise (artefacts, cropping)%\vp{citations for these claims?}.
Additionally, missing data is the strongest pain point, as models cannot be trained on datasets containing PII (e.g., addresses of customers). This causes low performance in production.  We wish to investigate if augmentation with synthetic data is a feasible solution for addressing the aforementioned problems. Particularly, we focus on two settings:

\begin{itemize}
    \item \textit{Intra-set Augmentation}: Can we use synthetic data to improve performance with only style augmentation? No new contents is added to the training set;
    \item \textit{Inter-set Augmentation}: Can we use synthetic data to improve performance with both contents and style augmentation, particularly when the base dataset contents greatly differs from the test set (e.g., real base dataset $\in$ words, deployment data $\in$ numerals)? 
\end{itemize}

%\paragraph{\textbf{Augmentation Approaches}}
% \subsubsection{Learning based Approaches}

% \textbf{LSTM} \\
% \textbf{GAN}\\
% \textbf{Diffusion} \\

% \subsubsection{Font based Generation}
% \srijan{Elaborate on why font-based approach is needed}
Prior work has explored deep learning based approaches for generating synthetic handwritten text~\cite{fogel2020scrabblegan, chang2022style}. We will explore font-based generation as a viable alternative for improving classification performance in data scarce regimes.

\subsubsection{Font-based Augmentation}
We built a font-based text generator that uses computer fonts in conjunction with various image transformations to create synthetic handwriting data. The five image transformations we consider are rotation, elastic distortion, Gaussian blur, masking with white patches, and masking with black patches. The strength of these transformations can be varied, and they can be combined to modify a source image. 

The generator starts by selecting a font style from a set of fifteen options and produces an image using the input text. Next, it samples strength levels for the individual transformations and then applies them to combine random transformation. 

\textbf{Dataset details:} The base dataset consists of $\sim34,000$ images, which we split into a training set of $\sim26,500$ images, a validation set of $\sim5,000$ images, and an additional holdout set of $2,500$ images. Of the $34,000$ images, approximately $20,000$ are images of words (names, places etc.), with $\sim5,300$ unique labels. The remaining $\sim14,000$ are images containing numerals (dates, numeric amounts etc.) with $\sim3,500$ unique labels. 
%For a sample of both see Figure~\ref{fig:examples}.

%\begin{figure}[!h]
%    \centering
%    \includegraphics[width=\textwidth]{figures/handwriting/examples.PNG}
%    \caption{Examples of images in the base dataset, including images of words such as names and places (top) or numerals such as dates and numeric amount (bottom).}
%    \label{fig:examples}
%\end{figure}
%
\textbf{Experiment Procedure:} The experiment is conducted using the following steps:

\begin{enumerate}
    \item Create dataset with a preset number of real examples (= R) plus X\% new examples coming from augmentation. 
For instance, if R = $7000$ and augmentation level X = 50\%, the dataset would comprise 7,000 real examples, and 3,500 synthetic ones. 

\item Train OCR model using curated dataset. The OCR model selected is a Convolutional Neural network with 18 million parameters. 

\item Test on holdout test set, and measure Character Error Rate (CER)   
\end{enumerate}
The CER metric is based on the Levenshtein Edit Distance that computes the number of operations required to transform one string into another. It correlates with the percentage of characters in the ground-truth text that were incorrectly transcribed.
%The metric is computed as:
%\begin{equation}
%CER = \frac{S + D + I}%{N},
%\end{equation}
%where S is the number of substitutions, D the number of deletions, I the number of insertions and N the total number of characters when transforming one string into another.
%Just as a demonstrative example, consider the word in the figure below. The ground truth for that word is ``Salt Lake'', but assume a OCR model outputs ``Sah Larke''. We require 3 operations to transform the prediction to the ground truth (insert L after ``Sa'', substitute the H after ``Sa'' with T, delete the R after ``La''), so:
%
%
%\begin{minipage}{0.45\textwidth}
%\begin{figure}[H]
%\includegraphics{figures/handwriting/CER_example.PNG}
%\end{figure}
%\end{minipage}
%\begin{minipage}{0.45\textwidth}
%\begin{equation*}
%    CER = \frac{S + D + I}{N} = \frac{1 + 1 + 1}{9} = \frac{1}{3} = 0.\bar{3}
%\end{equation*}
%\end{minipage}
% 
Lower CER values are better, as they indicate lower error. Note that we can have CER $> 1.0$ in cases where the prediction is longer than the ground truth. 

\textbf{Results:} As previously stated, we investigate two scenarios:

\begin{enumerate}
    \item \textit{Intra-set Augmentation}: We would like to study if we can use synthetic data to improve performance with only style augmentation. In this case, no new contents is added to the training set. Instead, contents for the synthetic examples is sampled from the set of ground-truth labels of the training set. The first experiment was performed with the real dataset size $R = 7,000$ samples. This was to demonstrate the effects of the augmentation when the real dataset is of a limited size. We observed that augmenting with synthetic data improved model performance. As shown in Figure~\ref{fig:intra-results}, the highest level of improvement ($26.11\%$) was achieved with $100\%$ augmentation; further augmentation degraded performance from this level, although still a marked improvement from the $0\%$ baseline. The second experiment was performed with the real dataset size $R = 21,000$ samples. This represents a realistic setting in OCR training, and helps us gauge whether synthetic data can help boost performance in such scenarios. Similar to the previous experiment, Figure~\ref{fig:intra-results} shows that synthetic data augmentation improved model performance, with the most improvement ($18.92\%$) observed at $100\%$ augmentation. Further augmentation resulted in a plateau.

    \begin{figure}[!h]
    \centering
    \includegraphics[width=0.47\textwidth]{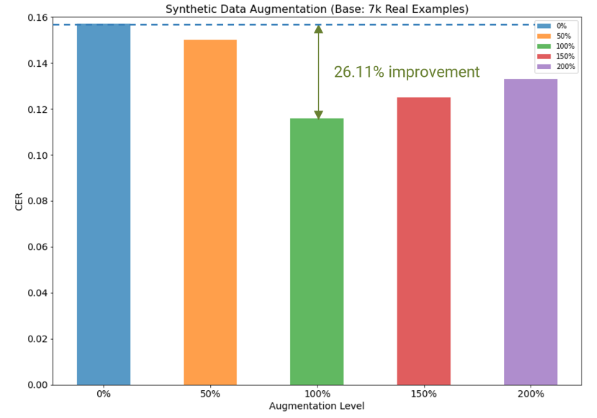}
    \hfill
    \includegraphics[width=0.47\textwidth]{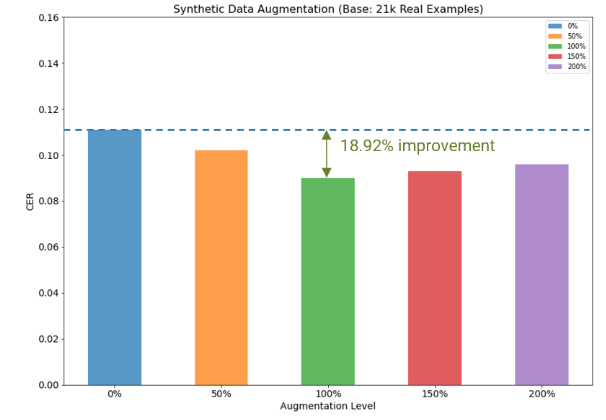}
    \caption{Results of synthetic data augmentation with real dataset size of $7,000$ samples. We observe a $26.11\%$ improvement with $100\%$ augmentation. Results of synthetic data augmentation with real dataset size of $21,000$ samples. We observe a $18.92\%$ improvement with $100\%$ augmentation.}
    \label{fig:intra-results}
    \end{figure}

   \item \textit{Inter-set Augmentation}: Now, we want to analyze if we can use synthetic data to improve performance with both contents and style augmentation, particularly when the training set contents greatly differs from and with potentially little to no overlap with the test set (e.g., real base dataset $\in$ words, deployment data $\in$ numerals).  This experiment is akin to test the OCR model in cases of dataset shift, i.e., when the distribution of the test set greatly differs from the training set. In this experiment, the real data comprises only words (names, places etc.), whereas the OCR model will be tested on a dataset comprising of numerals (dates, amounts etc.). Therefore, the synthetic data introduces samples with numeric contents to gauge if this helps the model perform inference on the test set. This mimics a real-world contents gap scenario, where the training set consists of words and numerals in independent samples (the dataset described previously), whereas the deployment dataset consists of addresses (alphanumeric data). It is not possible to train the model directly on addresses due to data privacy considerations related to PII (Personally Identifiable Information). As in the first experiment, we observed that augmenting with synthetic data improved model performance (Figure~\ref{fig:inter-results-1}). The highest level of improvement ($40.68\%$) was achieved with $100\%$ augmentation, with further augmentation degrading performance. The baseline in this experiment was set to be the OCR model trained with $1\%$ synthetic data augmentation. With no data augmentation, the OCR model implementation would indeed not compile on a test set with unrecognized and previously unseen characters.

    \begin{figure}[!h]
    \centering
    \includegraphics[width=0.85\textwidth]{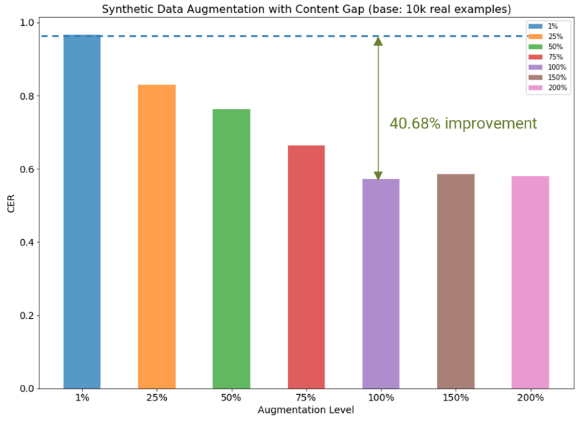}
    \caption{Results of synthetic data augmentation in a contents gap scenario with real dataset size of $10,000$ samples. We observe a $40.68\%$ improvement with $100\%$ augmentation.}
    \label{fig:inter-results-1}
    \end{figure}

\end{enumerate}
%\vp{any conclusions or takeaways? What are future steps on can take in this domain?}

\subsection{Document generation}
Financial documents such as prospectus, credit reports, purchase orders, sustainability disclosures, income statements, invoices etc.,  are presented in a range of visual styles and encompass a diverse set of semantics.
Machine interpretation of such documents demands numerous instances, typically with ground-truth examples included. However, curating these documents is challenging especially if they are proprietary or governed by copyright and licensing restrictions. Furthermore, annotation exercises at scale tend to be resource intensive. Synthetically generated documents provide a plausible alternative to explicit collection of documents and manual annotation with massive quantities of automatically annotated documents being cheaply produced on demand. 

\subsubsection{Layout generation}
\label{sec:document}
A financial document contains various elements such as headers, sections, tables, figures and fields that are organized in a logical fashion to ensure the document is visually appealing and easy to read. Comprehending this layout structure is critical to understanding the document and involves training a visual layout recognition system using a sizable number of documents labelled with their corresponding gold layout structure. Synthetic document generation can produce visually distinct layouts, introduce new layout categories and inherently capture the labels and extents of the layout elements during the construction process, leading to automatic inclusion of ground-truth labels for the various layouts.   

A simple approach for generating documents that appear dissimilar to the original is to perturb the various style attributes of an existing document~\cite{etter2019synthetic}. This technique though is only feasible for non-binary document formats like HTML, CSS, or LaTeX, and it is limited by a narrow range of variations. Deep generative models for layout generation using self-attention ~\cite{gupta2021layouttransformer}, adversarial training ~\cite{li2020layoutgan}, autoencoder structures~\cite{jyothi2019layoutvae} and semi-automatic annotation ~\cite{pisaneschi2023automatic} have been proposed. However, these methods tend to require initial annotations to warm up, cannot generate new primitives and suffer from image quality issues. Instead of deep neural models,~\cite{raman2022synthetic} proposes a Bayesian Network approach to generate realistic synthetic documents, which allows creating graphical units from scratch, does not require seed documents, and is interpretable. Their solution treats every primitive unit, style attribute and logical element of a document as a random variable and represents the dependencies between these random variables using conditional probability distributions. Figure ~\ref{fig:doclayout_1} shows an example synthetic form generated using this approach.

\begin{table*}[!t]
\centering
\begin{tabular}{cc}
\includegraphics[width=.49\linewidth]{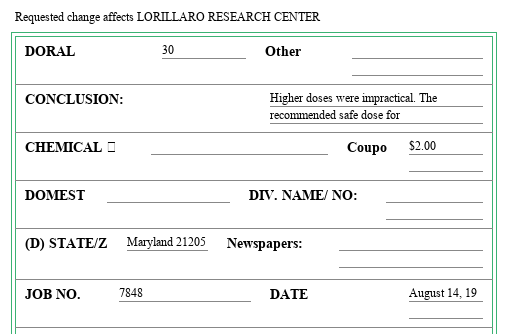} & \includegraphics[width=.49\linewidth]{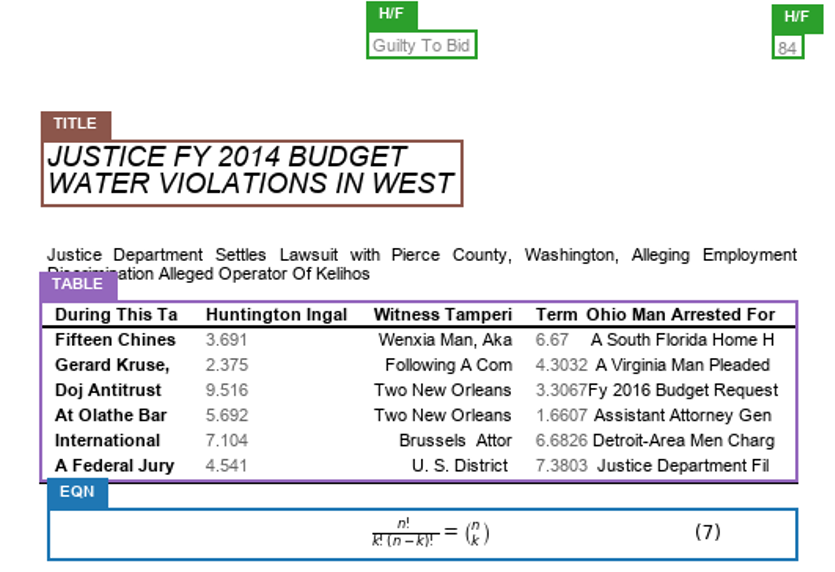}

\end{tabular}
\captionof{figure}{Machine generated synthetic documents~\cite{raman2022synthetic}. \emph{left}: A synthetic form. \emph{right}: A synthetic report annotated with layout categories such as headers, titles, or tables.}
\label{fig:doclayout_1}
\end{table*}

\subsubsection{Diagram generation}

Visual diagrams that depict enterprise ownerships, management hierarchies, supply chain networks and start-up activities are often embedded inside a financial document.  These diagrams illustrate complex ideas, relationships and incidents in a compact manner. An automatic question-answering system that explores these diagrams would offer considerable benefits to a business expert.  However, developing such a system for \emph{visually rich document understanding} demands the collection of enough diagrams, a challenging task given their proprietary nature.

Synthetic diagrams that faithfully reflect the properties of their real-world counterparts offer a viable option. While there have been previous efforts to curate plots~\cite{kahou2017figureqa}, infographics~\cite{mathew2022infographicvqa} , flowcharts~\cite{tannertflowchartqa} and scientific diagrams ~\cite{kembhavi2016diagram}, they do not pertain to the finance domain. ~\cite{babkin2023bizgraph} proposes an algorithmic approach for generating synthetic diagrams applicable to various businesses. Given the underlying graph structures of real-world diagrams, they treat a graph adjacency matrix as an image, and employ an image generative model to produce new samples of these graphs. The graphs are then rendered by applying different style attributes to produce new synthetic diagrams. The availability of the underlying graph structure also facilitates the population of template question and answers, which can be used to perform complex reasoning over the diagrams. Figure~\ref{fig:docdiag_1} provides a sample synthetic diagram of a company ownership structure.

\begin{table*}[!t]
\centering
\begin{tabular}{c}
\includegraphics[width=\linewidth]{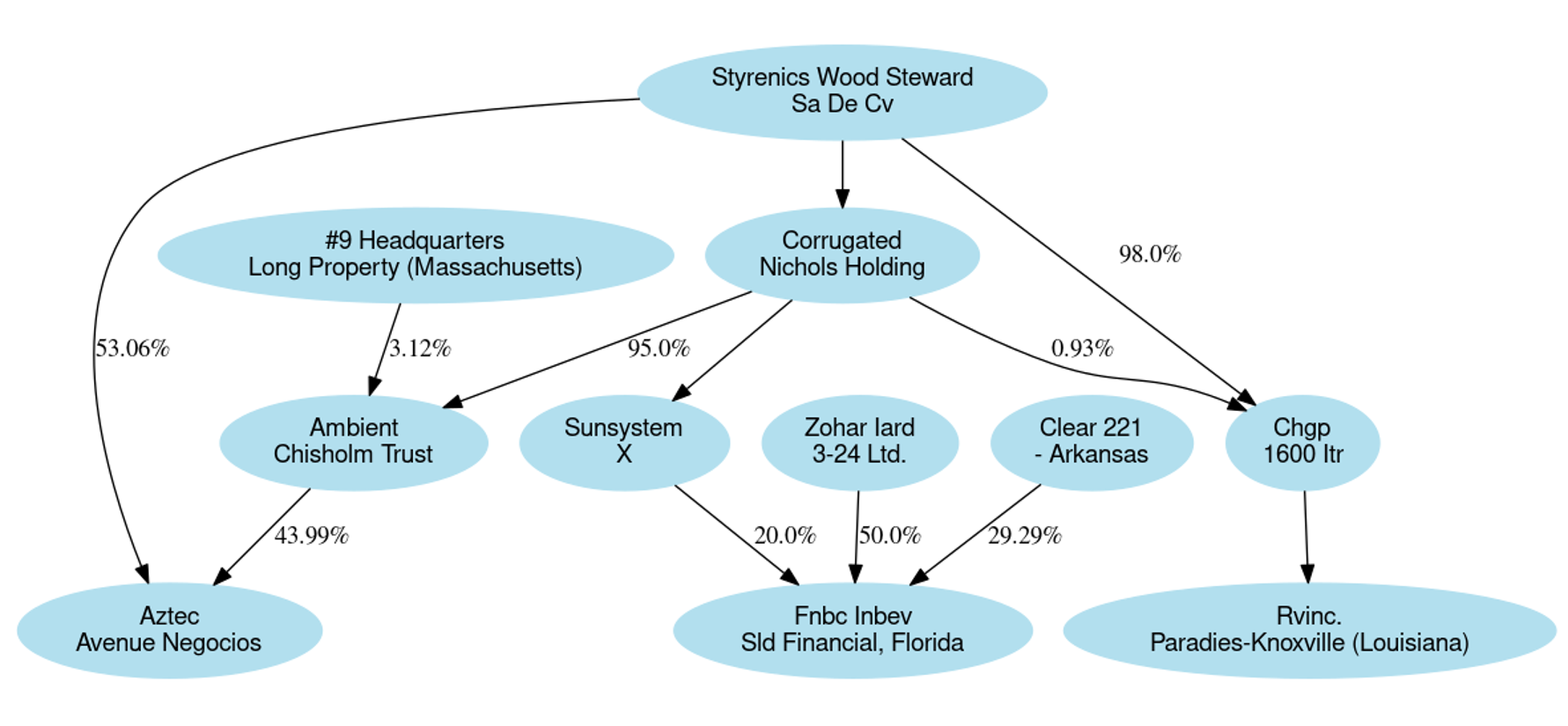} \\
Question: Who is the ultimate parent of Sunsystem X? \\
Answer: Styrenics Wood Steward Sa De Cv \\

\end{tabular}
\captionof{figure}{Synthetic enterprise ownership diagram, along with gold question and answer pair. The diagrams can be used for training a visually rich document understanding system~\cite{babkin2023bizgraph}.     }
\label{fig:docdiag_1}
\end{table*}

\subsubsection{Text generation}
Identifying the financial entities in a document or finance specific sentiment analysis~\cite{shah2022flue} and question answering~\cite{chen2021finqa} require the availability of annotated text. Since labelled data is scarce, it is common to augment manually annotated examples with synthetically generated text. Typical approaches for text augmentation include word substitutions, back translation, summarization and paraphrasing, where a given text is converted into its equivalent form using a language model~\cite{li2022data}. Recent approaches such as AugGPT~\cite{dai2023chataug} exploit large language models for conditional text generation and can achieve comparable or even better performance than task tuned models by using a small number of demonstrations. All the above methods largely aim to preserve the semantics of the original text while altering the lexical structure and can be applied to produce synthetic financial documents that are faithful to the original facts yet diverse in surface realizations. 

It is also useful to generate text that diverges in semantics from the original. For instance, let a synthetic document reflect a \emph{bearish} sentiment in contrast to the \emph{bullish} sentiment expressed in the original. Such documents can simulate different scenarios in the financial markets and help train robust models. While the counterfactual text generation techniques in ~\cite{madaan2021generate, dixit2022-core}  are not specific to finance, they provide a generic mechanism to develop documents that demonstrate an alternative outcome. Besides modifying the facts, the text generation process may also employ privacy enhancing techniques to create synthetic documents without any personally identifiable information~\cite{zhao2022survey}.   

%%%\subsection{Others}
%%%Chatgpt and NLP projects could go here and we can have additional sections instead of others.

\section{Conclusion and Open Challenges}

We have shown the application of synthetic data on a wide range of applications in the finance industry including fraud, customer acquisition, distributional shifts in markets, realistic time-series generation with constraints, and OCR models for checks. We expect with the increasing success of chatGPT~\cite{openai2023gpt4} in various domains that they will have a wide spread impact in financial applications. 

Evaluating the quality of generated data (e.g. do we have task agnostic accurate metrics for assessing the preservation of and diversity in semantics, perception, structure, statistical properties, anomalies etc.), 
transferability of the techniques across datasets and generalization to different tasks, explanation of what differs between a synthetic and original, Multimodal synthetic data (rather than data mode specific methods), availability of original representative data (e.g. companies with proprietary data hold advantage), and bias enhancement will increasingly play a role in the quality of synthetic data benchmarks. 
Distinguishing synthetic data from real is a very pertinent question and will be increasingly an important problem as we have seen with rise of Deepfakes~\cite{kietzmann2020deepfakes}. Watermarking~\cite{pmlr-v202-kirchenbauer23a} and other related cryptographic techniques such as digital signatures~\cite{katz2010digital} will increasingly play a role for solving these kind of problems. 
The use of synthetic data in financial applications is still in its infancy and we expect further open problems to arise with a wider adoption and deployments in real-world use cases.

\paragraph{\textbf{Disclaimer}:} This paper was prepared for informational purposes by the Artificial Intelligence Research group of JPMorgan Chase \& Co. and its affiliates (``JP Morgan''), and is not a product of the Research Department of JP Morgan. JP Morgan makes no representation and warranty whatsoever and disclaims all liability, for the completeness, accuracy or reliability of the information contained herein. This document is not intended as investment research or investment advice, or a recommendation, offer or solicitation for the purchase or sale of any security, financial instrument, financial product or service, or to be used in any way for evaluating the merits of participating in any transaction, and shall not constitute a solicitation under any jurisdiction or to any person, if such solicitation under such jurisdiction or to such person would be unlawful. 
\bibliographystyle{alpha}
\bibliography{sample,privacy}

\newcommand{\etalchar}[1]{$^{#1}$}
\begin{thebibliography}{GPAM{\etalchar{+}}14b}

\bibitem[ABG{\etalchar{+}}23]{arora2023faster}
Raman Arora, Raef Bassily, Tom{\'a}s Gonz{\'a}lez, Crist{\'o}bal~A Guzm{\'a}n,
  Michael Menart, and Enayat Ullah.
\newblock Faster rates of convergence to stationary points in differentially
  private optimization.
\newblock In {\em International Conference on Machine Learning}, pages
  1060--1092. PMLR, 2023.

\bibitem[ACB17]{WGAN}
Martin Arjovsky, Soumith Chintala, and L{\'e}on Bottou.
\newblock {W}asserstein generative adversarial networks.
\newblock In Doina Precup and Yee~Whye Teh, editors, {\em Proceedings of the
  34th International Conference on Machine Learning}, volume~70 of {\em
  Proceedings of Machine Learning Research}, pages 214--223. PMLR, 06--11 Aug
  2017.

\bibitem[ACG{\etalchar{+}}16]{Abadi_2016}
Martin Abadi, Andy Chu, Ian Goodfellow, H.~Brendan McMahan, Ilya Mironov, Kunal
  Talwar, and Li~Zhang.
\newblock Deep learning with differential privacy.
\newblock In {\em Proceedings of the 2016 {ACM} {SIGSAC} Conference on Computer
  and Communications Security}. {ACM}, oct 2016.

\bibitem[ACvdS20]{alaa2020generative}
Ahmed Alaa, Alex~James Chan, and Mihaela van~der Schaar.
\newblock Generative time-series modeling with fourier flows.
\newblock In {\em International Conference on Learning Representations}, 2020.

\bibitem[ADM{\etalchar{+}}20a]{assefa2020generating}
Samuel~A Assefa, Danial Dervovic, Mahmoud Mahfouz, Robert~E Tillman, Tucker
  Balch, Prashant Reddy, and Manuela Veloso.
\newblock Generating synthetic data in finance: opportunities, challenges and
  pitfalls.
\newblock In {\em Proceedings of the First ACM International Conference on AI
  in Finance}, pages 1--8, 2020.

\bibitem[ADM{\etalchar{+}}20b]{assefa2020generatingb}
Samuel~A Assefa, Danial Dervovic, Mahmoud Mahfouz, Robert~E Tillman, Prashant
  Reddy, Tucker Balch, and Manuela Veloso.
\newblock Generating synthetic data in finance: opportunities, challenges and
  pitfalls.
\newblock In {\em Proceedings of the First ACM International Conference on AI
  in Finance}, pages 1--8, 2020.

\bibitem[ADR17]{abhishek2017multi}
Vibhanshu Abhishek, Stylianos Despotakis, and R~Ravi.
\newblock Multi-channel attribution: The blind spot of online advertising.
\newblock {\em Available at SSRN 2959778}, 2017.

\bibitem[ADR{\etalchar{+}}19]{asghar2019differentially}
Hassan~Jameel Asghar, Ming Ding, Thierry Rakotoarivelo, Sirine Mrabet, and
  Mohamed~Ali Kaafar.
\newblock Differentially private release of high-dimensional datasets using the
  gaussian copula, 2019.

\bibitem[ADY{\etalchar{+}}18]{arava2018deep}
Sai~Kumar Arava, Chen Dong, Zhenyu Yan, Abhishek Pani, et~al.
\newblock Deep neural net with attention for multi-channel multi-touch
  attribution.
\newblock {\em arXiv preprint arXiv:1809.02230}, 2018.

\bibitem[AMAF19]{abdulkareem2019bayesian_data_human_mix_agentModels}
Shaheen~A Abdulkareem, Yaseen~T Mustafa, Ellen-Wien Augustijn, and Tatiana
  Filatova.
\newblock Bayesian networks for spatial learning: a workflow on using limited
  survey data for intelligent learning in spatial agent-based models.
\newblock {\em Geoinformatica}, 23:243--268, 2019.

\bibitem[ATLO17]{abar2017agentbasedsim_survey}
Sameera Abar, Georgios~K Theodoropoulos, Pierre Lemarinier, and Gregory~MP
  O’Hare.
\newblock Agent based modelling and simulation tools: A review of the
  state-of-art software.
\newblock {\em Computer Science Review}, 24:13--33, 2017.

\bibitem[AvBSvdS21]{alaa2021faithful}
Ahmed~M Alaa, Boris van Breugel, Evgeny Saveliev, and Mihaela van~der Schaar.
\newblock How faithful is your synthetic data? sample-level metrics for
  evaluating and auditing generative models.
\newblock {\em arXiv preprint arXiv:2102.08921}, 2021.

\bibitem[AVG{\etalchar{+}}22]{ardon2022phantom}
Leo Ardon, Jared Vann, Deepeka Garg, Tom Spooner, and Sumitra Ganesh.
\newblock Phantom--an rl-driven framework for agent-based modeling of complex
  economic systems and markets.
\newblock {\em arXiv preprint arXiv:2210.06012}, 2022.

\bibitem[BBDG18]{bouchaud2018trades}
Jean-Philippe Bouchaud, Julius Bonart, Jonathan Donier, and Martin Gould.
\newblock {\em Trades, quotes and prices: financial markets under the
  microscope}.
\newblock Cambridge University Press, 2018.

\bibitem[BD19]{wrongModels_useful_box1919essentially}
George~EP Box and Norman~R Draper.
\newblock Essentially, all models are wrong, but some are useful.
\newblock {\em Statistician}, 3(28):2013, 1919.

\bibitem[BDI{\etalchar{+}}23]{belgodere2023auditing}
Brian Belgodere, Pierre Dognin, Adam Ivankay, Igor Melnyk, Youssef Mroueh,
  Aleksandra Mojsilovic, Jiri Navartil, Apoorva Nitsure, Inkit Padhi, Mattia
  Rigotti, et~al.
\newblock Auditing and generating synthetic data with controllable trust
  trade-offs.
\newblock {\em arXiv preprint arXiv:2304.10819}, 2023.

\bibitem[BFELV23]{Bamford2023MADSMA}
Thomas Bamford, Elizabeth Fons, Yousef El-Laham, and Svitlana Vyetrenko.
\newblock Mads: Modulated auto-decoding siren for time series imputation.
\newblock {\em ArXiv}, abs/2307.00868, 2023.

\bibitem[BHB19]{byrd2019abides}
David Byrd, Maria Hybinette, and Tucker~Hybinette Balch.
\newblock Abides: Towards high-fidelity market simulation for {AI} research.
\newblock {\em arXiv preprint arXiv:1904.12066}, 2019.

\bibitem[BML{\etalchar{+}}19]{balch2019evaluate}
Tucker~Hybinette Balch, Mahmoud Mahfouz, Joshua Lockhart, Maria Hybinette, and
  David Byrd.
\newblock How to evaluate trading strategies: Single agent market replay or
  multiple agent interactive simulation?
\newblock {\em arXiv preprint arXiv:1906.12010}, 2019.

\bibitem[BP66]{baum1966statistical}
Leonard~E Baum and Ted Petrie.
\newblock Statistical inference for probabilistic functions of finite state
  markov chains.
\newblock {\em The annals of mathematical statistics}, 37(6):1554--1563, 1966.

\bibitem[BR11]{bauerle2011markov_financeUses}
Nicole B{\"a}uerle and Ulrich Rieder.
\newblock {\em Markov decision processes with applications to finance}.
\newblock Springer Science \& Business Media, 2011.

\bibitem[Bra14]{MonteCarloSim_finance}
Paolo Brandimarte.
\newblock {\em Handbook in Monte Carlo simulation: applications in financial
  engineering, risk management, and economics}.
\newblock John Wiley \& Sons, 2014.

\bibitem[BV20]{arxiv-finplan-simulator}
Daniel Borrajo and Manuela Veloso.
\newblock Domain-independent generation and classification of behavior traces.
\newblock {\em arXiv e-prints}, abs/2011.02918, 2020.

\bibitem[BVD17]{boucherie2017markov_mdpuses_2}
Richard~J Boucherie and Nico~M Van~Dijk.
\newblock {\em Markov decision processes in practice}, volume 248.
\newblock Springer, 2017.

\bibitem[BVS20]{icaif20}
Daniel Borrajo, Manuela Veloso, and Sameena Shah.
\newblock Simulating and classifying behavior in adversarial environments based
  on action-state traces: An application to money laundering.
\newblock In {\em Proceedings of the First ACM International Conference on AI
  in Finance}, New York (EEUU), 2020.
\newblock Also in: https://arxiv.org/abs/2011.01826.

\bibitem[BWM{\etalchar{+}}23]{babkin2023bizgraph}
Petr Babkin, William Watson, Zhiqiang Ma, Lucas Cecchi, Natraj Raman, Armineh
  Nourbakhsh, and Sameena Shah.
\newblock Bizgraphqa: A dataset for image-based inference over graph-structured
  diagrams from business domains.
\newblock In {\em Proceedings of the 46th International ACM SIGIR Conference on
  Research and Development in Information Retrieval}, 2023.

\bibitem[Byr19]{byrd2019explaining}
David Byrd.
\newblock Explaining agent-based financial market simulation.
\newblock {\em CoRR}, abs/1909.11650, 2019.

\bibitem[CBC{\etalchar{+}}22]{chang2022data}
Jen-Hao~Rick Chang, Martin Bresler, Youssouf Chherawala, Adrien Delaye, Thomas
  Deselaers, Ryan Dixon, and Oncel Tuzel.
\newblock Data incubation—synthesizing missing data for handwriting
  recognition.
\newblock In {\em ICASSP 2022-2022 IEEE International Conference on Acoustics,
  Speech and Signal Processing (ICASSP)}, pages 4188--4192. IEEE, 2022.

\bibitem[CBHK02]{chawla2002smote}
Nitesh~V Chawla, Kevin~W Bowyer, Lawrence~O Hall, and W~Philip Kegelmeyer.
\newblock Smote: synthetic minority over-sampling technique.
\newblock {\em Journal of artificial intelligence research}, 16:321--357, 2002.

\bibitem[CCS{\etalchar{+}}21]{chen2021finqa}
Zhiyu Chen, Wenhu Chen, Charese Smiley, Sameena Shah, Iana Borova, Dylan
  Langdon, Reema Moussa, Matt Beane, Ting-Hao Huang, Bryan Routledge, et~al.
\newblock Finqa: A dataset of numerical reasoning over financial data.
\newblock 2021.

\bibitem[CELT{\etalchar{+}}22]{cao2022dslob}
Defu Cao, Yousef El-Laham, Loc Trinh, Svitlana Vyetrenko, and Yan Liu.
\newblock Dslob: A synthetic limit order book dataset for benchmarking
  forecasting algorithms under distributional shift, 2022.

\bibitem[CG16]{chen2016xgboost}
Tianqi Chen and Carlos Guestrin.
\newblock Xgboost: A scalable tree boosting system.
\newblock In {\em Proceedings of the 22nd acm sigkdd international conference
  on knowledge discovery and data mining}, pages 785--794, 2016.

\bibitem[CGBV23]{coletta2023constrained}
Andrea Coletta, Sriram Gopalakrishan, Daniel Borrajo, and Svitlana Vyetrenko.
\newblock On the constrained time-series generation problem.
\newblock In {\em NeurIPS}, 2023.

\bibitem[Cho95]{chorafas1995financial_models_and_sim}
D~Chorafas.
\newblock {\em Financial models and simulation}.
\newblock Springer, 1995.

\bibitem[CI80]{cox1980point}
David~Roxbee Cox and Valerie Isham.
\newblock {\em Point processes}, volume~12.
\newblock CRC Press, 1980.

\bibitem[Cin13]{cinlar2013introduction_stochMarkovprocess_book}
Erhan Cinlar.
\newblock {\em Introduction to stochastic processes}.
\newblock Courier Corporation, 2013.

\bibitem[CJSV23]{coletta2023conditional}
Andrea Coletta, Joseph Jerome, Rahul Savani, and Svitlana Vyetrenko.
\newblock Conditional generators for limit order book environments:
  Explainability, challenges, and robustness.
\newblock {\em arXiv preprint arXiv:2306.12806}, 2023.

\bibitem[CK11]{chakraboty}
Tanmoy Chakraborty and Michael Kearns.
\newblock Market making and mean reversion.
\newblock pages 307--314, 06 2011.

\bibitem[CLM22]{chiang2022hawkes}
Wen-Hao Chiang, Xueying Liu, and George Mohler.
\newblock Hawkes process modeling of covid-19 with mobility leading indicators
  and spatial covariates.
\newblock {\em International journal of forecasting}, 38(2):505--520, 2022.

\bibitem[CLW21]{chen2021}
Yinbo Chen, Sifei Liu, and Xiaolong Wang.
\newblock Learning continuous image representation with local implicit image
  function.
\newblock In {\em Proceedings of the IEEE/CVF Conference on Computer Vision and
  Pattern Recognition}, pages 8628--8638, 2021.

\bibitem[CMVB22]{coletta2022learning}
Andrea Coletta, Aymeric Moulin, Svitlana Vyetrenko, and Tucker Balch.
\newblock Learning to simulate realistic limit order book markets from data as
  a world agent.
\newblock In {\em Proceedings of the Third ACM International Conference on AI
  in Finance}, pages 428--436, 2022.

\bibitem[Cox55]{cox1955some}
David~R Cox.
\newblock Some statistical methods connected with series of events.
\newblock {\em Journal of the Royal Statistical Society: Series B
  (Methodological)}, 17(2):129--157, 1955.

\bibitem[CPC{\etalchar{+}}21]{coletta2021towards}
Andrea Coletta, Matteo Prata, Michele Conti, Emanuele Mercanti, Novella
  Bartolini, Aymeric Moulin, Svitlana Vyetrenko, and Tucker Balch.
\newblock Towards realistic market simulations: a generative adversarial
  networks approach.
\newblock In {\em Proceedings of the Second ACM International Conference on AI
  in Finance}, pages 1--9, 2021.

\bibitem[CR10]{caiola2010random}
Gregory Caiola and Jerome~P Reiter.
\newblock Random forests for generating partially synthetic, categorical data.
\newblock {\em Trans. Data Priv.}, 3(1):27--42, 2010.

\bibitem[CRS{\etalchar{+}}19]{carmon2022unlabeled}
Yair Carmon, Aditi Raghunathan, Ludwig Schmidt, John~C Duchi, and Percy~S
  Liang.
\newblock Unlabeled data improves adversarial robustness.
\newblock In H.~Wallach, H.~Larochelle, A.~Beygelzimer, F.~d\textquotesingle
  Alch\'{e}-Buc, E.~Fox, and R.~Garnett, editors, {\em Advances in Neural
  Information Processing Systems}, volume~32. Curran Associates, Inc., 2019.

\bibitem[CSK{\etalchar{+}}22]{chang2022style}
Jen-Hao~Rick Chang, Ashish Shrivastava, Hema Koppula, Xiaoshuai Zhang, and
  Oncel Tuzel.
\newblock Style equalization: Unsupervised learning of controllable generative
  sequence models.
\newblock In {\em International Conference on Machine Learning}, pages
  2917--2937. PMLR, 2022.

\bibitem[CVB23]{coletta2023k}
Andrea Coletta, Svitlana Vyetrenko, and Tucker Balch.
\newblock K-shap: Policy clustering algorithm for anonymous state-action pairs.
\newblock {\em arXiv preprint arXiv:2302.11996}, 2023.

\bibitem[CWL{\etalchar{+}}18]{Cao_BRITS}
Wei Cao, Dong Wang, Jian Li, Hao Zhou, Lei Li, and Yitan Li.
\newblock Brits: Bidirectional recurrent imputation for time series.
\newblock In S.~Bengio, H.~Wallach, H.~Larochelle, K.~Grauman, N.~Cesa-Bianchi,
  and R.~Garnett, editors, {\em Advances in Neural Information Processing
  Systems}, volume~31. Curran Associates, Inc., 2018.

\bibitem[CZM{\etalchar{+}}19]{autoaugment}
Ekin~D. Cubuk, Barret Zoph, Dandelion Mane, Vijay Vasudevan, and Quoc~V. Le.
\newblock Autoaugment: Learning augmentation strategies from data.
\newblock In {\em Proceedings of the IEEE/CVF Conference on Computer Vision and
  Pattern Recognition (CVPR)}, June 2019.

\bibitem[CZSL20]{randaugment}
Ekin~D. Cubuk, Barret Zoph, Jonathon Shlens, and Quoc~V. Le.
\newblock Randaugment: Practical automated data augmentation with a reduced
  search space.
\newblock In {\em Proceedings of the IEEE/CVF Conference on Computer Vision and
  Pattern Recognition (CVPR) Workshops}, June 2020.

\bibitem[DFWB21]{desai2021timevae}
Abhyuday Desai, Cynthia Freeman, Zuhui Wang, and Ian Beaver.
\newblock Timevae: A variational auto-encoder for multivariate time series
  generation.
\newblock {\em arXiv preprint arXiv:2111.08095}, 2021.

\bibitem[DL24]{du2024towards}
Yuntao Du and Ninghui Li.
\newblock Towards principled assessment of tabular data synthesis algorithms.
\newblock {\em arXiv preprint arXiv:2402.06806}, 2024.

\bibitem[DLAG{\etalchar{+}}20]{di2020efficient}
Luca Di~Liello, Pierfrancesco Ardino, Jacopo Gobbi, Paolo Morettin, Stefano
  Teso, and Andrea Passerini.
\newblock Efficient generation of structured objects with constrained
  adversarial networks.
\newblock {\em Advances in neural information processing systems},
  33:14663--14674, 2020.

\bibitem[DLL{\etalchar{+}}23]{dai2023chataug}
Haixing Dai, Zhengliang Liu, Wenxiong Liao, Xiaoke Huang, Zihao Wu, Lin Zhao,
  Wei Liu, Ninghao Liu, Sheng Li, Dajiang Zhu, et~al.
\newblock Chataug: Leveraging chatgpt for text data augmentation.
\newblock {\em arXiv preprint arXiv:2302.13007}, 2023.

\bibitem[DM19]{du2019implicit}
Yilun Du and Igor Mordatch.
\newblock Implicit generation and modeling with energy based models.
\newblock {\em Advances in Neural Information Processing Systems}, 32, 2019.

\bibitem[Doe16]{doersch2016tutorial}
Carl Doersch.
\newblock Tutorial on variational autoencoders.
\newblock {\em arXiv preprint arXiv:1606.05908}, 2016.

\bibitem[DPHZ22]{dixit2022-core}
Tanay Dixit, Bhargavi Paranjape, Hannaneh Hajishirzi, and Luke Zettlemoyer.
\newblock {CORE}: A retrieve-then-edit framework for counterfactual data
  generation.
\newblock In {\em Findings of the Association for Computational Linguistics:
  EMNLP 2022}, pages 2964--2984, Abu Dhabi, United Arab Emirates, December
  2022. Association for Computational Linguistics.

\bibitem[DR{\etalchar{+}}14]{dwork2014algorithmic}
Cynthia Dwork, Aaron Roth, et~al.
\newblock The algorithmic foundations of differential privacy.
\newblock {\em Found. Trends Theor. Comput. Sci.}, 9(3-4):211--407, 2014.

\bibitem[DVJ{\etalchar{+}}03]{daley2003introduction}
Daryl~J Daley, David Vere-Jones, et~al.
\newblock {\em An introduction to the theory of point processes: volume I:
  elementary theory and methods}.
\newblock Springer, 2003.

\bibitem[DZ13]{dassios2013hawkes}
Angelos Dassios and Hongbiao Zhao.
\newblock {Exact simulation of Hawkes process with exponentially decaying
  intensity}.
\newblock {\em Electronic Communications in Probability}, 18(none):1 -- 13,
  2013.

\bibitem[DZGZ21]{deng2021improving}
Zhun Deng, Linjun Zhang, Amirata Ghorbani, and James Zou.
\newblock Improving adversarial robustness via unlabeled out-of-domain data.
\newblock In Arindam Banerjee and Kenji Fukumizu, editors, {\em Proceedings of
  The 24th International Conference on Artificial Intelligence and Statistics},
  volume 130 of {\em Proceedings of Machine Learning Research}, pages
  2845--2853. PMLR, 13--15 Apr 2021.

\bibitem[EHR17a]{esteban2017real}
Crist{\'o}bal Esteban, Stephanie~L Hyland, and Gunnar R{\"a}tsch.
\newblock Real-valued (medical) time series generation with recurrent
  conditional gans.
\newblock {\em arXiv preprint arXiv:1706.02633}, 2017.

\bibitem[EHR17b]{esteban2017realvalued}
Cristóbal Esteban, Stephanie~L. Hyland, and Gunnar Rätsch.
\newblock Real-valued (medical) time series generation with recurrent
  conditional gans, 2017.

\bibitem[EKM13]{embrechts2013modelling_rare_events_finance}
Paul Embrechts, Claudia Kl{\"u}ppelberg, and Thomas Mikosch.
\newblock {\em Modelling extremal events: for insurance and finance},
  volume~33.
\newblock Springer Science \& Business Media, 2013.

\bibitem[ELDFV23]{el2023deep}
Yousef El-Laham, Niccol{\`o} Dalmasso, Elizabeth Fons, and Svitlana Vyetrenko.
\newblock Deep gaussian mixture ensembles.
\newblock {\em arXiv preprint arXiv:2306.07235}, 2023.

\bibitem[ELV22]{el2022styletime}
Yousef El-Laham and Svitlana Vyetrenko.
\newblock Styletime: Style transfer for synthetic time series generation.
\newblock In {\em Proceedings of the Third ACM International Conference on AI
  in Finance}, pages 489--496, 2022.

\bibitem[ERCS19]{etter2019synthetic}
David Etter, Stephen Rawls, Cameron Carpenter, and Gregory Sell.
\newblock A synthetic recipe for ocr.
\newblock In {\em 2019 International Conference on Document Analysis and
  Recognition (ICDAR)}, pages 864--869. IEEE, 2019.

\bibitem[ET08]{eno2008generating}
Josh Eno and Craig~W Thompson.
\newblock Generating synthetic data to match data mining patterns.
\newblock {\em IEEE Internet Computing}, 12(3):78--82, 2008.

\bibitem[EVY{\etalchar{+}}23]{FinPlan23-goal}
Andrew Estornell, Stylianos~Loukas Vasileiou, William Yeoh, Daniel Borrajo, and
  Rui Silva.
\newblock Predicting customer goals in financial institution services: A
  data-driven {LSTM} approach.
\newblock In {\em ICAPS Planning for Financial Services Workshop}, 2023.

\bibitem[FAEC{\etalchar{+}}20]{fogel2020scrabblegan}
Sharon Fogel, Hadar Averbuch-Elor, Sarel Cohen, Shai Mazor, and Roee Litman.
\newblock Scrabblegan: Semi-supervised varying length handwritten text
  generation.
\newblock In {\em Proceedings of the IEEE/CVF conference on computer vision and
  pattern recognition}, pages 4324--4333, 2020.

\bibitem[FDjZ{\etalchar{+}}21]{fons2021adaptive}
Elizabeth Fons, Paula Dawson, Xiao jun Zeng, John Keane, and Alexandros
  Iosifidis.
\newblock Adaptive weighting scheme for automatic time-series data
  augmentation, 2021.

\bibitem[FDZ{\etalchar{+}}20]{fons2020evaluating}
Elizabeth Fons, Paula Dawson, Xiao-jun Zeng, John Keane, and Alexandros
  Iosifidis.
\newblock Evaluating data augmentation for financial time series
  classification.
\newblock {\em arXiv preprint arXiv:2010.15111}, 2020.

\bibitem[FDZ{\etalchar{+}}21]{fons2021transfer}
Elizabeth Fons, Paula Dawson, Xiao-jun Zeng, John Keane, and Alexandros
  Iosifidis.
\newblock Augmenting transferred representations for stock classification.
\newblock In {\em ICASSP 2021 - 2021 IEEE International Conference on
  Acoustics, Speech and Signal Processing (ICASSP)}, pages 3915--3919, 2021.

\bibitem[fed]{federalreserve}
https://www.federalreserve.gov/publications/2023-stress-test-scenarios.htm.

\bibitem[FHMR88]{fox1988knowledgeBasedSim}
Mark Fox, Nizwer Husain, Malcolm McRoberts, and YV~Reddy.
\newblock {\em Knowledge based simulation: an artificial intelligence approach
  to system modeling and automating the simulation life cycle}.
\newblock Carnegie Mellon University, the Robotics Institute, 1988.

\bibitem[FSEL{\etalchar{+}}22]{Fons2022HyperTimeIN}
Elizabeth Fons, Alejandro Sztrajman, Yousef El-Laham, Alexandros Iosifidis, and
  Svitlana Vyetrenko.
\newblock Hypertime: Implicit neural representation for time series.
\newblock {\em ArXiv}, abs/2208.05836, 2022.

\bibitem[FVLK{\etalchar{+}}17]{fournier2017survey}
Philippe Fournier-Viger, Jerry Chun-Wei Lin, Rage~Uday Kiran, Yun~Sing Koh, and
  Rincy Thomas.
\newblock A survey of sequential pattern mining.
\newblock {\em Data Science and Pattern Recognition}, 1(1):54--77, 2017.

\bibitem[GBR21]{gogoshin2021synthetic}
Grigoriy Gogoshin, Sergio Branciamore, and Andrei~S Rodin.
\newblock Synthetic data generation with probabilistic bayesian networks.
\newblock {\em Mathematical biosciences and engineering: MBE}, 18(6):8603,
  2021.

\bibitem[GBWT23]{anonymeter}
Matteo Giomi, Franziska Boenisch, Christoph Wehmeyer, and Borbála Tasnádi.
\newblock A unified framework for quantifying privacy risk in synthetic data,
  2023.

\bibitem[GDL{\etalchar{+}}22]{ghassemi2022online}
Mohsen Ghassemi, Niccol\`o Dalmasso, Simran Lamba, Vamsi Potluru, Tucker Balch,
  Sameena Shah, and Manuela Veloso.
\newblock Online learning for mixture of multivariate hawkes processes.
\newblock In {\em Proceedings of the Third ACM International Conference on AI
  in Finance}, ICAIF '22, page 506–513, New York, NY, USA, 2022. Association
  for Computing Machinery.

\bibitem[GG16]{gal2016dropout}
Yarin Gal and Zoubin Ghahramani.
\newblock Dropout as a bayesian approximation: Representing model uncertainty
  in deep learning.
\newblock In {\em international conference on machine learning}, pages
  1050--1059. PMLR, 2016.

\bibitem[Gha97]{ghahramani1997learning}
Zoubin Ghahramani.
\newblock Learning dynamic bayesian networks.
\newblock {\em International School on Neural Networks, Initiated by IIASS and
  EMFCSC}, pages 168--197, 1997.

\bibitem[GKD{\etalchar{+}}22]{ghassemi2022hawkesprivacy}
Mohsen Ghassemi, Eleonora Kreačić, Niccolò Dalmasso, Vamsi~K. Potluru,
  Tucker Balch, and Manuela Veloso.
\newblock Differentially private learning of hawkes processes.
\newblock 2022.

\bibitem[GLA{\etalchar{+}}21]{gupta2021layouttransformer}
Kamal Gupta, Justin Lazarow, Alessandro Achille, Larry~S Davis, Vijay
  Mahadevan, and Abhinav Shrivastava.
\newblock Layouttransformer: Layout generation and completion with
  self-attention.
\newblock In {\em Proceedings of the IEEE/CVF International Conference on
  Computer Vision}, pages 1004--1014, 2021.

\bibitem[GNT04]{planning-book}
Malik Ghallab, Dana Nau, and Paolo Traverso.
\newblock {\em Automated Planning. {T}heory \& Practice}.
\newblock Morgan Kaufmann, 2004.

\bibitem[Goo16]{goodfellow2016nips}
Ian Goodfellow.
\newblock Nips 2016 tutorial: Generative adversarial networks.
\newblock {\em arXiv preprint arXiv:1701.00160}, 2016.

\bibitem[GPAM{\etalchar{+}}14a]{Goodfellow_GAN}
Ian Goodfellow, Jean Pouget-Abadie, Mehdi Mirza, Bing Xu, David Warde-Farley,
  Sherjil Ozair, Aaron Courville, and Yoshua Bengio.
\newblock Generative adversarial nets.
\newblock In Z.~Ghahramani, M.~Welling, C.~Cortes, N.~Lawrence, and K.Q.
  Weinberger, editors, {\em Advances in Neural Information Processing Systems},
  volume~27. Curran Associates, Inc., 2014.

\bibitem[GPAM{\etalchar{+}}14b]{goodfellow2014generative}
Ian~J. Goodfellow, Jean Pouget-Abadie, Mehdi Mirza, Bing Xu, David
  Warde-Farley, Sherjil Ozair, Aaron Courville, and Yoshua Bengio.
\newblock Generative adversarial networks, 2014.

\bibitem[Gra11]{graves2011practical}
Alex Graves.
\newblock Practical variational inference for neural networks.
\newblock {\em Advances in neural information processing systems}, 24, 2011.

\bibitem[GSLO{\etalchar{+}}17]{guimaraes2017objective}
Gabriel~Lima Guimaraes, Benjamin Sanchez-Lengeling, Carlos Outeiral, Pedro
  Luis~Cunha Farias, and Al{\'a}n Aspuru-Guzik.
\newblock Objective-reinforced generative adversarial networks (organ) for
  sequence generation models.
\newblock {\em arXiv preprint arXiv:1705.10843}, 2017.

\bibitem[HAP21]{harder2021dpmerf}
Frederik Harder, Kamil Adamczewski, and Mijung Park.
\newblock Dp-merf: Differentially private mean embeddings with random features
  for practical privacy-preserving data generation, 2021.

\bibitem[Haw71]{hawkes1971spectra}
Alan~G Hawkes.
\newblock Spectra of some self-exciting and mutually exciting point processes.
\newblock {\em Biometrika}, 58(1):83--90, 1971.

\bibitem[HCS{\etalchar{+}}22]{houssiau2022framework}
Florimond Houssiau, Samuel~N Cohen, Lukasz Szpruch, Owen Daniel, Michaela~G
  Lawrence, Robin Mitra, Henry Wilde, and Callum Mole.
\newblock A framework for auditable synthetic data generation.
\newblock {\em arXiv preprint arXiv:2211.11540}, 2022.

\bibitem[Heg22]{hegghammer2022ocr}
Thomas Hegghammer.
\newblock Ocr with tesseract, amazon textract, and google document ai: a
  benchmarking experiment.
\newblock {\em Journal of Computational Social Science}, 5(1):861--882, 2022.

\bibitem[HJC{\etalchar{+}}22]{houssiau2022tapas}
Florimond Houssiau, James Jordon, Samuel~N Cohen, Owen Daniel, Andrew Elliott,
  James Geddes, Callum Mole, Camila Rangel-Smith, and Lukasz Szpruch.
\newblock Tapas: a toolbox for adversarial privacy auditing of synthetic data.
\newblock In {\em NeurIPS 2022 Workshop on Synthetic Data for Empowering ML
  Research}, 2022.

\bibitem[HLC{\etalchar{+}}19]{PBA2019}
Daniel Ho, Eric Liang, Xi~Chen, Ion Stoica, and Pieter Abbeel.
\newblock Population based augmentation: Efficient learning of augmentation
  policy schedules.
\newblock In Kamalika Chaudhuri and Ruslan Salakhutdinov, editors, {\em
  Proceedings of the 36th International Conference on Machine Learning},
  volume~97 of {\em Proceedings of Machine Learning Research}, pages
  2731--2741. PMLR, 09--15 Jun 2019.

\bibitem[HNSOP23]{hamad2023supervised}
Fadi Hamad, Shinpei Nakamura-Sakai, Saheed Obitayo, and Vamsi Potluru.
\newblock A supervised generative optimization approach for tabular data.
\newblock In {\em 4th ACM International Conference on AI in Finance}, pages
  10--18, 2023.

\bibitem[HS97]{hochreiter1997long}
Sepp Hochreiter and J{\"u}rgen Schmidhuber.
\newblock Long short-term memory.
\newblock {\em Neural computation}, 9(8):1735--1780, 1997.

\bibitem[HSW89]{hornik1989multilayer}
Kurt Hornik, Maxwell Stinchcombe, and Halbert White.
\newblock Multilayer feedforward networks are universal approximators.
\newblock {\em Neural networks}, 2(5):359--366, 1989.

\bibitem[I{\etalchar{+}}08]{iacus2008simulation_markov_process_finance}
Stefano~M Iacus et~al.
\newblock {\em Simulation and inference for stochastic differential equations:
  with R examples}, volume 486.
\newblock Springer, 2008.

\bibitem[IU21]{timeseries_augmentation}
Brian~Kenji Iwana and Seiichi Uchida.
\newblock An empirical survey of data augmentation for time series
  classification with neural networks.
\newblock {\em PLOS ONE}, 16(7):e0254841, Jul 2021.

\bibitem[IW79]{isham1979self}
Valerie Isham and Mark Westcott.
\newblock A self-correcting point process.
\newblock {\em Stochastic processes and their applications}, 8(3):335--347,
  1979.

\bibitem[JB19]{jia2019neural}
Junteng Jia and Austin~R Benson.
\newblock Neural jump stochastic differential equations.
\newblock {\em Advances in Neural Information Processing Systems}, 32, 2019.

\bibitem[JCL{\etalchar{+}}22]{Jarrett_HyperImpute}
Daniel Jarrett, Bogdan Cebere, Tennison Liu, Alicia Curth, and Mihaela van~der
  Schaar.
\newblock Hyperimpute: Generalized iterative imputation with automatic model
  selection.
\newblock 2022.

\bibitem[JDH{\etalchar{+}}19]{jyothi2019layoutvae}
Akash~Abdu Jyothi, Thibaut Durand, Jiawei He, Leonid Sigal, and Greg Mori.
\newblock Layoutvae: Stochastic scene layout generation from a label set.
\newblock In {\em Proceedings of the IEEE/CVF International Conference on
  Computer Vision}, pages 9895--9904, 2019.

\bibitem[JSH{\etalchar{+}}22]{jordon2022synthetic}
James Jordon, Lukasz Szpruch, Florimond Houssiau, Mirko Bottarelli, Giovanni
  Cherubin, Carsten Maple, Samuel~N. Cohen, and Adrian Weller.
\newblock Synthetic data -- what, why and how?, 2022.

\bibitem[Kal60]{kalman1960new}
Rudolph~Emil Kalman.
\newblock A new approach to linear filtering and prediction problems.
\newblock {\em Transactions of the ASME--Journal of Basic Engineering},
  82(Series D):35--45, 1960.

\bibitem[Kat10]{katz2010digital}
Jonathan Katz.
\newblock {\em Digital signatures}, volume~1.
\newblock Springer, 2010.

\bibitem[KBRB23]{kotelnikov2023tabddpm}
Akim Kotelnikov, Dmitry Baranchuk, Ivan Rubachev, and Artem Babenko.
\newblock Tabddpm: Modelling tabular data with diffusion models.
\newblock In {\em International Conference on Machine Learning}, pages
  17564--17579. PMLR, 2023.

\bibitem[KBRF18]{kakalejvcik2018multichannel}
Luk{\'a}{\v{s}} Kakalej{\v{c}}{\'\i}k, Jozef Bucko, PA~Resende, and Martina
  Ferencova.
\newblock Multichannel marketing attribution using markov chains.
\newblock {\em Journal of Applied Management and Investments}, 7(1):49--60,
  2018.

\bibitem[KGW{\etalchar{+}}23]{pmlr-v202-kirchenbauer23a}
John Kirchenbauer, Jonas Geiping, Yuxin Wen, Jonathan Katz, Ian Miers, and Tom
  Goldstein.
\newblock A watermark for large language models.
\newblock In Andreas Krause, Emma Brunskill, Kyunghyun Cho, Barbara Engelhardt,
  Sivan Sabato, and Jonathan Scarlett, editors, {\em Proceedings of the 40th
  International Conference on Machine Learning}, volume 202 of {\em Proceedings
  of Machine Learning Research}, pages 17061--17084. PMLR, 23--29 Jul 2023.

\bibitem[KLMK20]{kietzmann2020deepfakes}
Jan Kietzmann, Linda~W Lee, Ian~P McCarthy, and Tim~C Kietzmann.
\newblock Deepfakes: Trick or treat?
\newblock {\em Business Horizons}, 63(2):135--146, 2020.

\bibitem[KMA{\etalchar{+}}17]{kahou2017figureqa}
Samira~Ebrahimi Kahou, Vincent Michalski, Adam Atkinson, {\'A}kos
  K{\'a}d{\'a}r, Adam Trischler, and Yoshua Bengio.
\newblock Figureqa: An annotated figure dataset for visual reasoning.
\newblock {\em arXiv preprint arXiv:1710.07300}, 2017.

\bibitem[KNP{\etalchar{+}}23]{KDTree2023differentially}
Eleonora Krea\v{c}i\'{c}, Navid Nouri, Vamsi~K. Potluru, Tucker Balch, and
  Manuela Veloso.
\newblock Differentially private synthetic data using {KD}-trees.
\newblock In {\em The 39th Conference on Uncertainty in Artificial
  Intelligence}, 2023.

\bibitem[Kri16]{krishnamurthy2016_pomdp}
Vikram Krishnamurthy.
\newblock {\em Partially observed Markov decision processes}.
\newblock Cambridge university press, 2016.

\bibitem[KSK{\etalchar{+}}16]{kembhavi2016diagram}
Aniruddha Kembhavi, Mike Salvato, Eric Kolve, Minjoon Seo, Hannaneh Hajishirzi,
  and Ali Farhadi.
\newblock A diagram is worth a dozen images.
\newblock In {\em Computer Vision--ECCV 2016: 14th European Conference,
  Amsterdam, The Netherlands, October 11--14, 2016, Proceedings, Part IV 14},
  pages 235--251. Springer, 2016.

\bibitem[LCC{\etalchar{+}}17]{DBLP:conf/nips/LiCCYP17}
Chun{-}Liang Li, Wei{-}Cheng Chang, Yu~Cheng, Yiming Yang, and Barnab{\'{a}}s
  P{\'{o}}czos.
\newblock {MMD} {GAN:} towards deeper understanding of moment matching network.
\newblock In Isabelle Guyon, Ulrike von Luxburg, Samy Bengio, Hanna~M. Wallach,
  Rob Fergus, S.~V.~N. Vishwanathan, and Roman Garnett, editors, {\em Advances
  in Neural Information Processing Systems 30: Annual Conference on Neural
  Information Processing Systems 2017, December 4-9, 2017, Long Beach, CA,
  {USA}}, pages 2203--2213, 2017.

\bibitem[LeB06]{lebaron2006agent}
Blake LeBaron.
\newblock Agent-based computational finance.
\newblock {\em Handbook of computational economics}, 2:1187--1233, 2006.

\bibitem[LHC22]{li2022data}
Bohan Li, Yutai Hou, and Wanxiang Che.
\newblock Data augmentation approaches in natural language processing: A
  survey.
\newblock {\em AI Open}, 3:71--90, 2022.

\bibitem[LKK{\etalchar{+}}19]{fastautoaugment}
Sungbin Lim, Ildoo Kim, Taesup Kim, Chiheon Kim, and Sungwoong Kim.
\newblock Fast autoaugment.
\newblock In H.~Wallach, H.~Larochelle, A.~Beygelzimer, F.~d\textquotesingle
  Alch\'{e}-Buc, E.~Fox, and R.~Garnett, editors, {\em Advances in Neural
  Information Processing Systems}, volume~32, pages 6665--6675. Curran
  Associates, Inc., 2019.

\bibitem[LKPL23]{li2023graphmaker}
Mufei Li, Eleonora Krea{\v{c}}i{\'c}, Vamsi~K Potluru, and Pan Li.
\newblock Graphmaker: Can diffusion models generate large attributed graphs?
\newblock {\em arXiv preprint arXiv:2310.13833}, 2023.

\bibitem[LLS{\etalchar{+}}23]{longa2023graph}
Antonio Longa, Veronica Lachi, Gabriele Santin, Monica Bianchini, Bruno Lepri,
  Pietro Lio, franco scarselli, and Andrea Passerini.
\newblock Graph neural networks for temporal graphs: State of the art, open
  challenges, and opportunities.
\newblock {\em Transactions on Machine Learning Research}, 2023.

\bibitem[LM22]{llugiqi2022empirical}
Majlinda Llugiqi and Rudolf Mayer.
\newblock An empirical analysis of synthetic-data-based anomaly detection.
\newblock In {\em International Cross-Domain Conference for Machine Learning
  and Knowledge Extraction}, pages 306--327. Springer, 2022.

\bibitem[LPB17]{lakshminarayanan2017de}
Balaji Lakshminarayanan, Alexander Pritzel, and Charles Blundell.
\newblock Simple and scalable predictive uncertainty estimation using deep
  ensembles.
\newblock {\em Advances in neural information processing systems}, 30, 2017.

\bibitem[LRD22]{li2022fairgan}
Jie Li, Yongli Ren, and Ke~Deng.
\newblock Fairgan: Gans-based fairness-aware learning for recommendations with
  implicit feedback.
\newblock In {\em Proceedings of the ACM Web Conference 2022}, pages 297--307,
  2022.

\bibitem[LWLQ22]{lin2022survey}
Tianyang Lin, Yuxin Wang, Xiangyang Liu, and Xipeng Qiu.
\newblock A survey of transformers.
\newblock {\em AI Open}, 2022.

\bibitem[LWSF23]{lin2023summary}
Zinan Lin, Shuaiqi Wang, Vyas Sekar, and Giulia Fanti.
\newblock Summary statistic privacy in data sharing, 2023.

\bibitem[LXJ14]{Li2014DifferentiallyPS}
Haoran Li, Li~Xiong, and Xiaoqian Jiang.
\newblock Differentially private synthesization of multi-dimensional data using
  copula functions.
\newblock {\em Advances in database technology : proceedings. International
  Conference on Extending Database Technology}, 2014:475--486, 2014.

\bibitem[LXZ{\etalchar{+}}15]{Luo2015multi}
Dixin Luo, Hongteng Xu, Yi~Zhen, Xia Ning, Hongyuan Zha, Xiaokang Yang, and
  Wenjun Zhang.
\newblock Multi-task multi-dimensional hawkes processes for modeling event
  sequences.
\newblock In {\em Proceedings of the 24th International Conference on
  Artificial Intelligence}, IJCAI'15, page 3685–3691, 2015.

\bibitem[LYH{\etalchar{+}}20]{li2020layoutgan}
Jianan Li, Jimei Yang, Aaron Hertzmann, Jianming Zhang, and Tingfa Xu.
\newblock Layoutgan: Synthesizing graphic layouts with vector-wireframe
  adversarial networks.
\newblock {\em IEEE Transactions on Pattern Analysis and Machine Intelligence},
  43(7):2388--2399, 2020.

\bibitem[LZF20]{li2020sync}
Zheng Li, Yue Zhao, and Jialin Fu.
\newblock Sync: A copula based framework for generating synthetic data from
  aggregated sources.
\newblock In {\em 2020 International Conference on Data Mining Workshops
  (ICDMW)}, pages 571--578. IEEE, 2020.

\bibitem[MA23]{manousakas2023usefulness}
Dionysis Manousakas and Sergül Aydöre.
\newblock On the usefulness of synthetic tabular data generation, 2023.

\bibitem[Mac75]{macchi1975coincidence}
Odile Macchi.
\newblock The coincidence approach to stochastic point processes.
\newblock {\em Advances in Applied Probability}, 7(1):83--122, 1975.

\bibitem[MBT{\etalchar{+}}22]{mathew2022infographicvqa}
Minesh Mathew, Viraj Bagal, Rub{\`e}n Tito, Dimosthenis Karatzas, Ernest
  Valveny, and CV~Jawahar.
\newblock Infographicvqa.
\newblock In {\em Proceedings of the IEEE/CVF Winter Conference on Applications
  of Computer Vision}, pages 1697--1706, 2022.

\bibitem[MCV{\etalchar{+}}23]{KDD23-SD}
Jing Ma, Chen Chen, Anil Vullikanti, Ritwick Mishra, Gregory~R Madden, Daniel
  Borrajo, and Jundong Li.
\newblock A look into causal effects under entangled treatment in graphs:
  Investigating the impact of contact on {MRSA} infection.
\newblock In {\em KDD Conference. Applied Data Science Track}, 2023.

\bibitem[ME10]{mabroukeh2010taxonomy}
Nizar~R Mabroukeh and Christie~I Ezeife.
\newblock A taxonomy of sequential pattern mining algorithms.
\newblock {\em ACM Computing Surveys (CSUR)}, 43(1):1--41, 2010.

\bibitem[Miz16]{mizuta2016brief}
Takanobu Mizuta.
\newblock A brief review of recent artificial market simulation (agent-based
  model) studies for financial market regulations and/or rules.
\newblock {\em Available at SSRN 2710495}, 2016.

\bibitem[MMS09]{maller2009ornstein}
Ross~A Maller, Gernot M{\"u}ller, and Alex Szimayer.
\newblock Ornstein--uhlenbeck processes and extensions.
\newblock {\em Handbook of financial time series}, pages 421--437, 2009.

\bibitem[MMS{\etalchar{+}}21]{mehrabi2021survey}
Ninareh Mehrabi, Fred Morstatter, Nripsuta Saxena, Kristina Lerman, and Aram
  Galstyan.
\newblock A survey on bias and fairness in machine learning.
\newblock {\em ACM computing surveys (CSUR)}, 54(6):1--35, 2021.

\bibitem[MPPS21]{madaan2021generate}
Nishtha Madaan, Inkit Padhi, Naveen Panwar, and Diptikalyan Saha.
\newblock Generate your counterfactuals: Towards controlled counterfactual
  generation for text.
\newblock In {\em Proceedings of the AAAI Conference on Artificial
  Intelligence}, volume~35, pages 13516--13524, 2021.

\bibitem[MST{\etalchar{+}}20]{mildenhall2020nerf}
Ben Mildenhall, Pratul~P. Srinivasan, Matthew Tancik, Jonathan~T. Barron, Ravi
  Ramamoorthi, and Ren Ng.
\newblock Nerf: Representing scenes as neural radiance fields for view
  synthesis.
\newblock In {\em ECCV}, 2020.

\bibitem[Mur02]{murphy2002dynamic_DBN}
Kevin~Patrick Murphy.
\newblock {\em Dynamic bayesian networks: representation, inference and
  learning}.
\newblock University of California, Berkeley, 2002.

\bibitem[MV12]{masarotto2012gaussian}
Guido Masarotto and Cristiano Varin.
\newblock Gaussian copula marginal regression.
\newblock 2012.

\bibitem[New83]{newbold1983arima}
Paul Newbold.
\newblock Arima model building and the time series analysis approach to
  forecasting.
\newblock {\em Journal of forecasting}, 2(1):23--35, 1983.

\bibitem[NS07]{narayanan2007break}
Arvind Narayanan and Vitaly Shmatikov.
\newblock How to break anonymity of the netflix prize dataset, 2007.

\bibitem[ODZ{\etalchar{+}}16]{oord2016wavenet}
Aaron van~den Oord, Sander Dieleman, Heiga Zen, Karen Simonyan, Oriol Vinyals,
  Alex Graves, Nal Kalchbrenner, Andrew Senior, and Koray Kavukcuoglu.
\newblock Wavenet: A generative model for raw audio.
\newblock {\em arXiv preprint arXiv:1609.03499}, 2016.

\bibitem[OE18]{oussidi2018deep_gen_models_survey}
Achraf Oussidi and Azeddine Elhassouny.
\newblock Deep generative models: Survey.
\newblock In {\em 2018 International conference on intelligent systems and
  computer vision (ISCV)}, pages 1--8. IEEE, 2018.

\bibitem[Oga81]{ogata1981lewis}
Yosihiko Ogata.
\newblock On lewis' simulation method for point processes.
\newblock {\em IEEE transactions on information theory}, 27(1):23--31, 1981.

\bibitem[Oga98]{ogata1998space}
Yosihiko Ogata.
\newblock Space-time point-process models for earthquake occurrences.
\newblock {\em Annals of the Institute of Statistical Mathematics},
  50:379--402, 1998.

\bibitem[OMI22]{Iwana2022gating}
D.~Oba, S.~Matsuo, and B.~Iwana.
\newblock Dynamic data augmentation with gating networks for time series
  recognition.
\newblock In {\em 2022 26th International Conference on Pattern Recognition
  (ICPR)}, pages 3034--3040, Los Alamitos, CA, USA, aug 2022. IEEE Computer
  Society.

\bibitem[Ope23]{openai2023gpt4}
OpenAI.
\newblock Gpt-4 technical report, 2023.

\bibitem[PAE{\etalchar{+}}17]{papernot2017}
Nicolas Papernot, Martín Abadi, Ulfar Erlingsson, Ian Goodfellow, and Kunal
  Talwar.
\newblock Semi-supervised knowledge transfer for deep learning from private
  training data, 2017.

\bibitem[PCCK22]{prenzel2022dynamic}
Felix Prenzel, Rama Cont, Mihai Cucuringu, and Jonathan Kochems.
\newblock Dynamic calibration of order flow models with generative adversarial
  networks.
\newblock In {\em Proceedings of the Third ACM International Conference on AI
  in Finance}, pages 446--453, 2022.

\bibitem[PEK21]{paine2021quantum}
Annie~E Paine, Vincent~E Elfving, and Oleksandr Kyriienko.
\newblock Quantum quantile mechanics: solving stochastic differential equations
  for generating time-series.
\newblock {\em arXiv preprint arXiv:2108.03190}, 2021.

\bibitem[PGM23]{pisaneschi2023automatic}
Lorenzo Pisaneschi, Andrea Gemelli, and Simone Marinai.
\newblock Automatic generation of scientific papers for data augmentation in
  document layout analysis.
\newblock {\em Pattern Recognition Letters}, 167:38--44, 2023.

\bibitem[PH22]{passino2022mutually}
Francesco~Sanna Passino and Nicholas~A Heard.
\newblock Mutually exciting point process graphs for modeling dynamic networks.
\newblock {\em Journal of Computational and Graphical Statistics}, pages 1--15,
  2022.

\bibitem[PMG{\etalchar{+}}18]{Park_2018}
Noseong Park, Mahmoud Mohammadi, Kshitij Gorde, Sushil Jajodia, Hongkyu Park,
  and Youngmin Kim.
\newblock Data synthesis based on generative adversarial networks.
\newblock {\em Proceedings of the {VLDB} Endowment}, 11(10):1071--1083, jun
  2018.

\bibitem[PMG{\etalchar{+}}23]{finRDDL_patra}
Sunandita Patra, Mahmoud Mahfouz, Sriram Gopalakrishnan, Daniele Magazzeni, and
  Manuela Veloso.
\newblock Finrddl: Can ai planning be used for quantitative finance problems?
\newblock {\em FinPlan 2023}, 2023.

\bibitem[PSH17]{ping2017datasynthesizer}
Haoyue Ping, Julia Stoyanovich, and Bill Howe.
\newblock Datasynthesizer: Privacy-preserving synthetic datasets.
\newblock In {\em Proceedings of the 29th International Conference on
  Scientific and Statistical Database Management}, pages 1--5, 2017.

\bibitem[PSZ22]{patel2022model}
Neel Patel, Reza Shokri, and Yair Zick.
\newblock Model explanations with differential privacy.
\newblock In {\em Proceedings of the 2022 ACM Conference on Fairness,
  Accountability, and Transparency}, pages 1895--1904, 2022.

\bibitem[Put14a]{puterman2014markov_MDPbook}
Martin~L Puterman.
\newblock {\em Markov decision processes: discrete stochastic dynamic
  programming}.
\newblock John Wiley \& Sons, 2014.

\bibitem[Put14b]{MDP_book}
Martin~L Puterman.
\newblock {\em Markov decision processes: discrete stochastic dynamic
  programming}.
\newblock John Wiley \& Sons, 2014.

\bibitem[PWV16a]{patki2016synthetic}
Neha Patki, Roy Wedge, and Kalyan Veeramachaneni.
\newblock The synthetic data vault.
\newblock In {\em 2016 IEEE International Conference on Data Science and
  Advanced Analytics (DSAA)}, pages 399--410. IEEE, 2016.

\bibitem[PWV16b]{7796926}
Neha Patki, Roy Wedge, and Kalyan Veeramachaneni.
\newblock The synthetic data vault.
\newblock In {\em 2016 IEEE International Conference on Data Science and
  Advanced Analytics (DSAA)}, pages 399--410, 2016.

\bibitem[Rei18]{reinhart2018review}
Alex Reinhart.
\newblock A review of self-exciting spatio-temporal point processes and their
  applications.
\newblock {\em Statistical Science}, 33(3):299--318, 2018.

\bibitem[RG18]{reinhart2018self}
Alex Reinhart and Joel Greenhouse.
\newblock Self-exciting point processes with spatial covariates.
\newblock {\em Journal of the Royal Statistical Society. Series C (Applied
  Statistics)}, 67(5):1305--1329, 2018.

\bibitem[RHR{\etalchar{+}}22]{rosenblatt2022epistemic}
Lucas Rosenblatt, Anastasia Holovenko, Taras Rumezhak, Andrii Stadnik, Bernease
  Herman, Julia Stoyanovich, and Bill Howe.
\newblock Epistemic parity: Reproducibility as an evaluation metric for
  differential privacy.
\newblock {\em arXiv preprint arXiv:2208.12700}, 2022.

\bibitem[RJ86]{rabiner1986introduction}
Lawrence Rabiner and Biinghwang Juang.
\newblock An introduction to hidden markov models.
\newblock {\em ieee assp magazine}, 3(1):4--16, 1986.

\bibitem[RLMX17]{rizoiu2017hawkes}
Marian-Andrei Rizoiu, Young Lee, Swapnil Mishra, and Lexing Xie.
\newblock Hawkes processes for events in social media.
\newblock In {\em Frontiers of multimedia research}, pages 191--218. 2017.

\bibitem[RLP{\etalchar{+}}20]{rosenblatt2020differentially}
Lucas Rosenblatt, Xiaoyan Liu, Samira Pouyanfar, Eduardo de~Leon, Anuj Desai,
  and Joshua Allen.
\newblock Differentially private synthetic data: Applied evaluations and
  enhancements, 2020.

\bibitem[RPG{\etalchar{+}}21]{ramesh2021zero}
Aditya Ramesh, Mikhail Pavlov, Gabriel Goh, Scott Gray, Chelsea Voss, Alec
  Radford, Mark Chen, and Ilya Sutskever.
\newblock Zero-shot text-to-image generation.
\newblock In {\em International Conference on Machine Learning}, pages
  8821--8831. PMLR, 2021.

\bibitem[RSGZ{\etalchar{+}}21]{shaham2021}
Tamar Rott~Shaham, Michael Gharbi, Richard Zhang, Eli Shechtman, and Tomer
  Michaeli.
\newblock Spatially-adaptive pixelwise networks for fast image translation.
\newblock In {\em Computer Vision and Pattern Recognition (CVPR)}, 2021.

\bibitem[RSV22]{raman2022synthetic}
Natraj Raman, Sameena Shah, and Manuela Veloso.
\newblock Synthetic document generator for annotation-free layout recognition.
\newblock {\em Pattern Recognition}, 128:108660, 2022.

\bibitem[RT{\etalchar{+}}09]{rubino2009rare_event_sim_book}
Gerardo Rubino, Bruno Tuffin, et~al.
\newblock {\em Rare event simulation using Monte Carlo methods}, volume~73.
\newblock Wiley Online Library, 2009.

\bibitem[S{\etalchar{+}}10]{sanner2010relational}
Scott Sanner et~al.
\newblock Relational dynamic influence diagram language (rddl): Language
  description.
\newblock {\em Unpublished ms. Australian National University}, 32:27, 2010.

\bibitem[SBL{\etalchar{+}}18]{sajjadi2018assessing}
Mehdi~SM Sajjadi, Olivier Bachem, Mario Lucic, Olivier Bousquet, and Sylvain
  Gelly.
\newblock Assessing generative models via precision and recall.
\newblock {\em Advances in neural information processing systems}, 31, 2018.

\bibitem[SBY16]{shi2016end}
Baoguang Shi, Xiang Bai, and Cong Yao.
\newblock An end-to-end trainable neural network for image-based sequence
  recognition and its application to scene text recognition.
\newblock {\em IEEE transactions on pattern analysis and machine intelligence},
  39(11):2298--2304, 2016.

\bibitem[SC23]{shi2023neural}
Zijian Shi and John Cartlidge.
\newblock Neural stochastic agent-based limit order book simulation: A hybrid
  methodology.
\newblock {\em arXiv preprint arXiv:2303.00080}, 2023.

\bibitem[SCE{\etalchar{+}}22]{shah2022flue}
Raj Shah, Kunal Chawla, Dheeraj Eidnani, Agam Shah, Wendi Du, Sudheer Chava,
  Natraj Raman, Charese Smiley, Jiaao Chen, and Diyi Yang.
\newblock When flue meets flang: Benchmarks and large pretrained language model
  for financial domain.
\newblock In {\em Proceedings of the 2022 Conference on Empirical Methods in
  Natural Language Processing}, pages 2322--2335, 2022.

\bibitem[SDWMG15]{sohl2015deep}
Jascha Sohl-Dickstein, Eric Weiss, Niru Maheswaranathan, and Surya Ganguli.
\newblock Deep unsupervised learning using nonequilibrium thermodynamics.
\newblock In {\em International conference on machine learning}, pages
  2256--2265. PMLR, 2015.

\bibitem[SHCV19]{sattigeri2019fairness}
Prasanna Sattigeri, Samuel~C Hoffman, Vijil Chenthamarakshan, and Kush~R
  Varshney.
\newblock Fairness gan: Generating datasets with fairness properties using a
  generative adversarial network.
\newblock {\em IBM Journal of Research and Development}, 63(4/5):3--1, 2019.

\bibitem[SKB23]{shahrooei2023falsification_multifidelity_sim_safety_testing}
Zahra Shahrooei, Mykel~J Kochenderfer, and Ali Baheri.
\newblock Falsification of learning-based controllers through multi-fidelity
  bayesian optimization.
\newblock In {\em 2023 European Control Conference (ECC)}, pages 1--6. IEEE,
  2023.

\bibitem[SMB{\etalchar{+}}20]{sitzmann2020siren}
Vincent Sitzmann, Julien~N.P. Martel, Alexander~W. Bergman, David~B. Lindell,
  and Gordon Wetzstein.
\newblock Implicit neural representations with periodic activation functions.
\newblock In {\em Proc. NeurIPS}, 2020.

\bibitem[SMH{\etalchar{+}}22]{sehwag2022robust}
Vikash Sehwag, Saeed Mahloujifar, Tinashe Handina, Sihui Dai, Chong Xiang, Mung
  Chiang, and Prateek Mittal.
\newblock Robust learning meets generative models: Can proxy distributions
  improve adversarial robustness?
\newblock In {\em International Conference on Learning Representations}, 2022.

\bibitem[SSS16]{DBLP:journals/corr/ShokriSS16}
Reza Shokri, Marco Stronati, and Vitaly Shmatikov.
\newblock Membership inference attacks against machine learning models.
\newblock {\em CoRR}, abs/1610.05820, 2016.

\bibitem[SSW11]{SUN2011526}
Xiaoxun Sun, Lili Sun, and Hua Wang.
\newblock Extended k-anonymity models against sensitive attribute disclosure.
\newblock {\em Computer Communications}, 34(4):526--535, 2011.
\newblock Special issue: Building Secure Parallel and Distributed Networks and
  Systems.

\bibitem[SZZ{\etalchar{+}}23]{Sun_2023}
Hui Sun, Tianqing Zhu, Zhiqiu Zhang, Dawei Jin, Ping Xiong, and Wanlei Zhou.
\newblock Adversarial attacks against deep generative models on data: A survey.
\newblock {\em {IEEE} Transactions on Knowledge and Data Engineering},
  35(4):3367--3388, apr 2023.

\bibitem[TAC23]{teo2023fair}
Christopher~TH Teo, Milad Abdollahzadeh, and Ngai-Man Cheung.
\newblock Fair generative models via transfer learning.
\newblock In {\em Proceedings of the AAAI Conference on Artificial
  Intelligence}, volume~37, pages 2429--2437, 2023.

\bibitem[TBV23]{tillman2023privacy}
Robert~E Tillman, Tucker Balch, and Manuela Veloso.
\newblock Privacy-preserving energy-based generative models for marginal
  distribution protection.
\newblock {\em Transactions on Machine Learning Research}, 2023.

\bibitem[TFB{\etalchar{+}}22]{tannertflowchartqa}
Simon Tannert, Marcelo Feighelstein, Jasmina Bogojeska, Joseph Shtok, Assaf
  Arbelle4~Peter Staar, and Anika Schumann3 Jonas Kuhn1~Leonid Karlinsky.
\newblock Flowchartqa: The first large-scale benchmark for reasoning over
  flowcharts.
\newblock In {\em The 3rd Workshop on Document Intelligence. KDD}, 2022.

\bibitem[TGG{\etalchar{+}}22]{taitler2022pyrddlgym}
Ayal Taitler, Michael Gimelfarb, Sriram Gopalakrishnan, Martin Mladenov,
  Xiaotian Liu, and Scott Sanner.
\newblock pyrddlgym: From rddl to gym environments.
\newblock {\em arXiv preprint arXiv:2211.05939}, 2022.

\bibitem[TK21]{takeishi2021knowledge}
Naoya Takeishi and Yoshinobu Kawahara.
\newblock Knowledge-based regularization in generative modeling.
\newblock In {\em Proceedings of the Twenty-Ninth International Conference on
  International Joint Conferences on Artificial Intelligence}, pages
  2390--2396, 2021.

\bibitem[TMH{\etalchar{+}}22]{tao2022benchmarking}
Yuchao Tao, Ryan McKenna, Michael Hay, Ashwin Machanavajjhala, and Gerome
  Miklau.
\newblock Benchmarking differentially private synthetic data generation
  algorithms, 2022.

\bibitem[TWB{\etalchar{+}}21]{tantipongpipat2021differentially}
Uthaipon~Tao Tantipongpipat, Chris Waites, Digvijay Boob, Amaresh~Ankit Siva,
  and Rachel Cummings.
\newblock Differentially private synthetic mixed-type data generation for
  unsupervised learning.
\newblock {\em Intelligent Decision Technologies}, 15(4):779--807, 2021.

\bibitem[TWOH03]{teh2003energy}
Yee~Whye Teh, Max Welling, Simon Osindero, and Geoffrey~E Hinton.
\newblock Energy-based models for sparse overcomplete representations.
\newblock {\em Journal of Machine Learning Research}, 4(Dec):1235--1260, 2003.

\bibitem[UO30]{uhlenbeck1930theory}
George~E Uhlenbeck and Leonard~S Ornstein.
\newblock On the theory of the brownian motion.
\newblock {\em Physical Review}, 36(5):823, 1930.

\bibitem[vBKBvdS21]{van2021decaf}
Boris van Breugel, Trent Kyono, Jeroen Berrevoets, and Mihaela van~der Schaar.
\newblock Decaf: Generating fair synthetic data using causally-aware generative
  networks.
\newblock {\em Advances in Neural Information Processing Systems},
  34:22221--22233, 2021.

\bibitem[VBP{\etalchar{+}}19]{vyetrenko2019real}
Svitlana Vyetrenko, David Byrd, Nick Petosa, Mahmoud Mahfouz, Danial Dervovic,
  Manuela Veloso, and Tucker~Hybinette Balch.
\newblock Get real: Realism metrics for robust limit order book market
  simulations, 2019.

\bibitem[vdMH08]{JMLR:v9:vandermaaten08a}
Laurens van~der Maaten and Geoffrey Hinton.
\newblock Visualizing data using t-sne.
\newblock {\em Journal of Machine Learning Research}, 9(86):2579--2605, 2008.

\bibitem[VdOKE{\etalchar{+}}16]{van2016conditional}
Aaron Van~den Oord, Nal Kalchbrenner, Lasse Espeholt, Oriol Vinyals, Alex
  Graves, et~al.
\newblock Conditional image generation with pixelcnn decoders.
\newblock {\em Advances in neural information processing systems}, 29, 2016.

\bibitem[VL18]{FTC_CFPB_van2018technology}
Rory Van~Loo.
\newblock Technology regulation by default: Platforms, privacy, and the cfpb.
\newblock 2018.

\bibitem[VSP{\etalchar{+}}17]{vaswani2017attention}
Ashish Vaswani, Noam Shazeer, Niki Parmar, Jakob Uszkoreit, Llion Jones,
  Aidan~N Gomez, {\L}ukasz Kaiser, and Illia Polosukhin.
\newblock Attention is all you need.
\newblock {\em Advances in neural information processing systems}, 30, 2017.

\bibitem[VVdB17]{voigt2017eu}
Paul Voigt and Axel Von~dem Bussche.
\newblock The eu general data protection regulation (gdpr).
\newblock {\em A Practical Guide, 1st Ed., Cham: Springer International
  Publishing}, 10(3152676):10--5555, 2017.

\bibitem[WB{\etalchar{+}}95]{welch1995introduction}
Greg Welch, Gary Bishop, et~al.
\newblock An introduction to the kalman filter.
\newblock 1995.

\bibitem[Whi85]{white1985real_mdpUses}
Douglas~J White.
\newblock Real applications of markov decision processes.
\newblock {\em Interfaces}, 15(6):73--83, 1985.

\bibitem[WKKK20]{Wiese_2020}
Magnus Wiese, Robert Knobloch, Ralf Korn, and Peter Kretschmer.
\newblock Quant gans: deep generation of financial time series.
\newblock {\em Quantitative Finance}, page 1–22, Apr 2020.

\bibitem[WKW{\etalchar{+}}23]{wei2023inherent}
Rongzhe Wei, Eleonora Krea{\v{c}}i{\'c}, Haoyu Wang, Haoteng Yin, Eli Chien,
  Vamsi~K Potluru, and Pan Li.
\newblock On the inherent privacy properties of discrete denoising diffusion
  models.
\newblock {\em arXiv preprint arXiv:2310.15524}, 2023.

\bibitem[WSS{\etalchar{+}}20]{Wen2020TimeSD}
Qingsong Wen, Liang Sun, Xiaomin Song, Jing Gao, Xue Wang, and Huan Xu.
\newblock Time series data augmentation for deep learning: A survey.
\newblock In {\em International Joint Conference on Artificial Intelligence},
  2020.

\bibitem[WZZX21]{wei2021goodness}
Song Wei, Shixiang Zhu, Minghe Zhang, and Yao Xie.
\newblock Goodness-of-fit test for mismatched self-exciting processes.
\newblock In {\em International Conference on Artificial Intelligence and
  Statistics}, pages 1243--1251. PMLR, 2021.

\bibitem[XDP{\etalchar{+}}23]{xiong2023fwc}
Zikai Xiong, Niccolò Dalmasso, Vamsi~K. Potluru, Tucker Balch, and Manuela
  Veloso.
\newblock Fair wasserstein coresets, 2023.

\bibitem[XFY{\etalchar{+}}17]{xiao2017wasserstein}
Shuai Xiao, Mehrdad Farajtabar, Xiaojing Ye, Junchi Yan, Le~Song, and Hongyuan
  Zha.
\newblock Wasserstein learning of deep generative point process models.
\newblock {\em Advances in neural information processing systems}, 30, 2017.

\bibitem[XLW{\etalchar{+}}18]{xie2018differentially}
Liyang Xie, Kaixiang Lin, Shu Wang, Fei Wang, and Jiayu Zhou.
\newblock Differentially private generative adversarial network, 2018.

\bibitem[XSC22]{xing2022artificially}
Yue Xing, Qifan Song, and Guang Cheng.
\newblock Why do artificially generated data help adversarial robustness.
\newblock {\em Advances in Neural Information Processing Systems}, 35:954--966,
  2022.

\bibitem[XSCIV19]{xu2019modeling}
Lei Xu, Maria Skoularidou, Alfredo Cuesta-Infante, and Kalyan Veeramachaneni.
\newblock Modeling tabular data using conditional gan.
\newblock {\em Advances in neural information processing systems}, 32, 2019.

\bibitem[XWY{\etalchar{+}}19]{xu2019achieving}
Depeng Xu, Yongkai Wu, Shuhan Yuan, Lu~Zhang, and Xintao Wu.
\newblock Achieving causal fairness through generative adversarial networks.
\newblock In {\em Proceedings of the Twenty-Eighth International Joint
  Conference on Artificial Intelligence}, 2019.

\bibitem[XYZW18]{xu2018fairgan}
Depeng Xu, Shuhan Yuan, Lu~Zhang, and Xintao Wu.
\newblock Fairgan: Fairness-aware generative adversarial networks.
\newblock In {\em 2018 IEEE International Conference on Big Data (Big Data)},
  pages 570--575. IEEE, 2018.

\bibitem[XYZW19]{xu2019fairganplus}
Depeng Xu, Shuhan Yuan, Lu~Zhang, and Xintao Wu.
\newblock Fairgan+: Achieving fair data generation and classification through
  generative adversarial nets.
\newblock In {\em 2019 IEEE International Conference on Big Data (Big Data)},
  pages 1401--1406. IEEE, 2019.

\bibitem[XZF{\etalchar{+}}18]{xu2018semantic}
Jingyi Xu, Zilu Zhang, Tal Friedman, Yitao Liang, and Guy Broeck.
\newblock A semantic loss function for deep learning with symbolic knowledge.
\newblock In {\em International conference on machine learning}, pages
  5502--5511. PMLR, 2018.

\bibitem[YJvdS19a]{Yoon2019TimeseriesGA}
Jinsung Yoon, Daniel Jarrett, and Mihaela van~der Schaar.
\newblock Time-series generative adversarial networks.
\newblock In {\em NeurIPS}, 2019.

\bibitem[YJvdS19b]{yoon2018pategan}
Jinsung Yoon, James Jordon, and Mihaela van~der Schaar.
\newblock {PATE}-{GAN}: Generating synthetic data with differential privacy
  guarantees.
\newblock In {\em International Conference on Learning Representations}, 2019.

\bibitem[YSHZ19]{yu2019review}
Yong Yu, Xiaosheng Si, Changhua Hu, and Jianxun Zhang.
\newblock A review of recurrent neural networks: Lstm cells and network
  architectures.
\newblock {\em Neural computation}, 31(7):1235--1270, 2019.

\bibitem[YSS15]{yadagiri2015non}
Meghanath~Macha Yadagiri, Shiv~Kumar Saini, and Ritwik Sinha.
\newblock A non-parametric approach to the multi-channel attribution problem.
\newblock In {\em Web Information Systems Engineering--WISE 2015: 16th
  International Conference, Miami, FL, USA, November 1-3, 2015, Proceedings,
  Part I 16}, pages 338--352. Springer, 2015.

\bibitem[ZC22]{zhao2022survey}
Ying Zhao and Jinjun Chen.
\newblock A survey on differential privacy for unstructured data content.
\newblock {\em ACM Computing Surveys (CSUR)}, 54(10s):1--28, 2022.

\bibitem[ZCP{\etalchar{+}}17]{PrivBayes}
Jun Zhang, Graham Cormode, Cecilia Procopiuc, Divesh Srivastava, and Xiaokui
  Xiao.
\newblock Privbayes: Private data release via bayesian networks.
\newblock {\em ACM Transactions on Database Systems}, 42:1--41, 10 2017.

\bibitem[ZDG{\etalchar{+}}22]{zhao2022fast}
Renbo Zhao, Niccol{\`o} Dalmasso, Mohsen Ghassemi, Vamsi~K Potluru, Tucker
  Balch, and Manuela Veloso.
\newblock Fast learning of multidimensional hawkes processes via frank-wolfe.
\newblock {\em arXiv preprint arXiv:2212.06081}, 2022.

\bibitem[ZJL{\etalchar{+}}20]{Zuo2020transformer}
Simiao Zuo, Haoming Jiang, Zichong Li, Tuo Zhao, and Hongyuan Zha.
\newblock Transformer {H}awkes process.
\newblock In {\em Proceedings of the 37th International Conference on Machine
  Learning}, volume 119 of {\em Proceedings of Machine Learning Research},
  pages 11692--11702. PMLR, 13--18 Jul 2020.

\bibitem[ZLKY19]{zhang2019self}
Qiang Zhang, Aldo Lipani, Omer Kirnap, and Emine Yilmaz.
\newblock Self-attentive hawkes processes.
\newblock {\em arXiv preprint arXiv:1907.07561}, 2019.

\bibitem[ZLZZ22]{zuo2022differentially}
Simiao Zuo, Tianyi Liu, Tuo Zhao, and Hongyuan Zha.
\newblock Differentially private estimation of hawkes process.
\newblock {\em arXiv preprint arXiv:2209.07303}, 2022.

\bibitem[ZWL{\etalchar{+}}21]{PrivSyn}
Zhikun Zhang, Tianhao Wang, Ninghui Li, Jean Honorio, Michael Backes, Shibo He,
  Jiming Chen, and Yang Zhang.
\newblock {PrivSyn}: Differentially private data synthesis.
\newblock In {\em 30th USENIX Security Symposium (USENIX Security 21)}, pages
  929--946. USENIX Association, August 2021.

\bibitem[ZWZZ20]{adversarialautoaugment}
Xinyu Zhang, Qiang Wang, Jian Zhang, and Zhao Zhong.
\newblock Adversarial autoaugment.
\newblock In {\em International Conference on Learning Representations}, 2020.

\bibitem[ZZR19]{zhang2019deeplob}
Zihao Zhang, Stefan Zohren, and Stephen Roberts.
\newblock Deeplob: Deep convolutional neural networks for limit order books.
\newblock {\em IEEE Transactions on Signal Processing}, 67(11):3001--3012,
  2019.

\end{thebibliography}

\end{document}